\documentclass[10pt,twocolumn,letterpaper]{article}

\usepackage{cvpr}              %
\usepackage{graphicx}
\usepackage{amsmath}
\usepackage{amssymb}
\usepackage{booktabs}
\usepackage{xcolor}
\usepackage{makecell}
\usepackage{url}            %
\usepackage{booktabs}       %
\usepackage{amsfonts}       %
\usepackage{nicefrac}       %
\usepackage{microtype}      %
\usepackage{graphicx}
\usepackage{graphics}
\usepackage{pifont}
\usepackage{color}
\usepackage{multirow}
\usepackage{adjustbox}
\usepackage{subcaption}
\usepackage{gensymb}
\usepackage{bm}
\usepackage{algorithm} 
\usepackage{algpseudocode} 
\usepackage{wrapfig}
\usepackage[accsupp]{axessibility}

\newcommand\method{{NeurMiPs}}

\usepackage[pagebackref,breaklinks,colorlinks, bookmarks=false]{hyperref}

\usepackage[capitalize]{cleveref}
\crefname{section}{Sec.}{Secs.}
\Crefname{section}{Section}{Sections}
\Crefname{table}{Table}{Tables}
\crefname{table}{Tab.}{Tabs.}

\def\cvprPaperID{4834} %
\def\confName{CVPR}
\def\confYear{2022}

\makeatletter
\g@addto@macro\@maketitle{
\vspace{-3em}
\begin{figure}[H]
\setlength{\linewidth}{\textwidth}
\setlength{\hsize}{\textwidth}
\centering
\includegraphics[width=\linewidth]{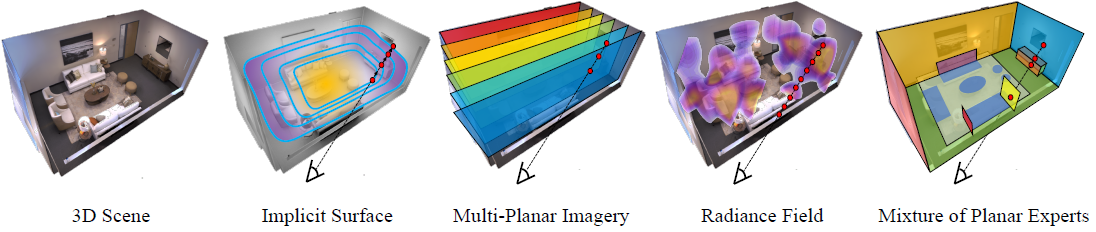}
\vspace{-7.5mm}
\caption{
\textbf{Comparison between different 3D representations for neural rendering.} \emph{Neural implicit surface} models accurate surface geometry but rendering requires expensive sequential sampling. 
  \emph{Multi-planar imagery} is efficient, but does not reflect true geometry
  and is not suitable for extrapolation. While NeRF is flexible, it is not sample-efficient and its learned density field might not reflect true scene geometry. Our proposed \emph{Mixture of Planar Experts} is efficient and is able to model complicated surface geometry and appearance.
}
\vspace{-1.5mm}
\label{fig:comparison}
\end{figure}
}
\makeatother

\begin{document}

\title{NeurMiPs: Neural Mixture of Planar Experts for View Synthesis}

\author{Zhi-Hao Lin$^{1,2}$ \quad Wei-Chiu Ma$^{3*}$ \quad Hao-Yu Hsu$^{2*}$ \quad Yu-Chiang Frank Wang$^{2}$ \quad Shenlong Wang$^{1}$\\
$^{1}$University of Illinois at Urbana-Champaign \quad $^{2}$National Taiwan University\\ $^{3}$Massachusetts Institute of Technology
}

\maketitle

\begin{abstract}
We present {Neural Mixtures of Planar Experts}~(NeurMiPs), a novel planar-based scene representation for modeling geometry and appearance. 
NeurMiPs leverages a collection of local planar experts in 3D space as the scene representation.  Each planar expert consists of the parameters of the local rectangular shape representing geometry and a neural radiance field modeling the color and opacity.  
We render novel views by calculating ray-plane intersections and composite output colors and densities at intersected points to the image. 
NeurMiPs blends the efficiency of explicit mesh rendering and flexibility of the neural radiance field. 
Experiments demonstrate superior performance and speed of our proposed method, compared to other 3D representations in novel view synthesis\footnote{Project page: \href{https://zhihao-lin.github.io/neurmips/}{https://zhihao-lin.github.io/neurmips/}.}.
\end{abstract}
\section{Introduction}
Imagine one day in the future where people can explore the world freely and immersively without
leaving their room. When they move forward, details pop up; and when they move sideways, the occluded regions re-appear. Whenever people take action, the world will respond
with corresponding visual scenes that look natural, as if people are visiting the place in person. While appealing,
bringing this vision to reality requires advancement in
multiple domains, one of which is real-time, high-quality,
and memory-efficient novel view synthesis. Specifically, given a set of posed images of the world, an ideal NVS system needs to re-render the scene from novel viewpoints photo-realistically. The system also needs to be fast and lightweight such that it can be deployed ubiquitously. %

Towards this grand goal, researchers have developed a plethora of methods to reproduce our visual world. One promising direction is to explicitly model the geometry of the scene (\eg multi-planar imagery~\cite{flynn2019deepview, zhou2018stereo, flynn2016deepstereo, Wizadwongsa2021NeX}, point clouds~\cite{aliev2020neural, ruckert2021adop, mallya2020world}, meshes~\cite{kopf2014first, riegler2020free, riegler2021stable}) and conduct image-based rendering (IBR) ~\cite{debevec1996modeling, debevec1998efficient, baker1998layered, szeliski19992d, kopf2014first}. By adapting visual features from other existing views, these approaches can render high-quality images efficiently. Unfortunately, they are often memory intensive and require good proxy geometry.
On the other hand, recent advances in neural radiance fields~\cite{mildenhall2020nerf, zhang2020nerf++, reiser2021kilonerf, yu2021plenoctrees, yu2021pixelnerf, zhang2021nerfactor} have allowed us to synthesize highly realistic images with low memory footprint. By encoding color and density functions as neural networks, they can handle complicated geometry and scene effects that are difficult for conventional methods, \eg, thin structures, specular reflections, and semi-transparent objects.
The flexibility of volume rendering, however, is a double-bladed sword. Without proper surface modeling, they cannot capture the scene geometry accurately, resulting in artifacts during view-extrapolation setup.

With these motivations in mind, we aim to find an alternative 3D scene representation that is compact, efficient, expressive, and generalizable. Specifically, we investigate planes, one of the simplest geometric primitives yet powerful for representing complicated scenes. 
Most surfaces in man-made environments are locally planar. Taking the scene in Fig.~\ref{fig:plane_fitting} as an example, one could use 500 planes to fit a $5\times 5$ m$^2$ scene with a maximum point-to-surface error of 8.66 mm. This result indicates that we might consider modeling our real-world surfaces through piece-wise local planar structures. 
We note that this concept is not new to many researchers in vision~\cite{martin2020gelato, liu2019planercnn, qian2020learning}. The planar structure also profoundly impacts the graphics community and is the most common representation for rendering. 

Unlike multi-planar imagery~\cite{flynn2019deepview, tucker2020single, Wizadwongsa2021NeX, zhou2018stereo}, which represents the scene through frontal-parallel planes, our approach allows each plane to have an arbitrary position, direction, and size. Consequently, our representation is more flexible to approximate the scene geometry. Unlike volume rendering, \method~explicitly models surface using planar geometry. Hence fast rendering can be done through the efficient ray-plane intersection and eliminate the computations in empty spaces. Fig.~\ref{fig:comparison} depicts our proposed 3D representation and a comparison to other representations for neural rendering. 

We validate our approach on several standard benchmarks for novel-view synthesis. The experiments demonstrate that our end-to-end method is significantly faster than the volume-based method with similar or better rendering quality and performs favorably against surface-based neural rendering methods with higher rendering quality and memory reduction. Furthermore, we evaluate our approach on a new challenging benchmark for view extrapolation, demonstrating superior performance compared to other state-of-the-art methods. In particular, \method~outperforms NeRF with over 1dB PSNR gain at significant novel testing views. The explicit planar surface representation of NeurMiPs could also readily be employed in modern graphics engines. 

\begin{figure}[t]
    \centering
    \vspace{-8mm}
    \includegraphics[width=.85\linewidth]{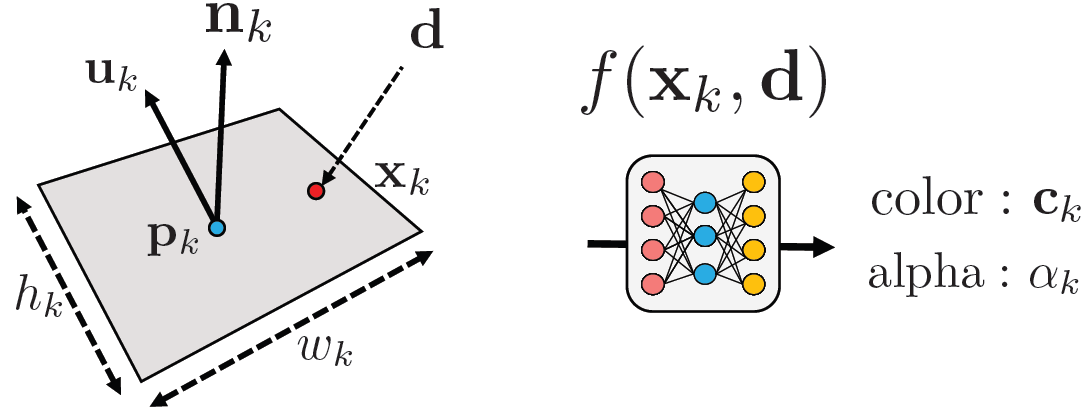}
       \vspace{-3mm}
 \caption{\textbf{Planar Expert Parameterization} Left: each plane consists of a 3D center, a plane normal, an up vector, and width and height; Right: the appearance is modeled through a neural radiance field function, which takes the 3D coordinate and ray direction as input and outputs color and opacity.}
      \vspace{-5mm}
    \label{fig:parameterization}
\end{figure}
\section{Related Work}
Our approach is closely related to classical work on image-based modeling and rendering, as well as recent learning-based efforts. We also draw inspiration from the prior art on planar scene representations. In this section, we briefly review previous work in these two major directions.

\subsection{Novel View Synthesis}
\paragraph{Explicit surface modeling: }
Explicit surface geometry is a critical component in pioneering works on view synthesis. For instance, mesh representations have been adopted as proxy geometry to guide the image-based warping from source views to a target pose \cite{debevec1996modeling, buehler2001unstructured, debevec1998efficient, kopf2014first}. Dense point clouds and surfels are alternative explicit representations of polygonal meshes~\cite{ruckert2021adop, aliev2020neural, yang2020surfelgan, mallya2020world}. Both are suitable for hardware acceleration, hence, they have superior efficiency and high rendering quality for synthetic data. It is, however, challenging to handle imperfect geometry and view-dependent effects. Recent works investigate learning approaches for explicit surface rendering~\cite{thies2019deferred, thies2020image,riegler2020free, riegler2021stable} or using 2D networks to retouch the images~\cite{mallya2020world, wiles2020synsin}. \method~is a form of explicit geometry. Thus we inherit its speed advantage. However, unlike most explicit geometry methods, we leverage a neural radiance field to improve rendering quality while remaining memory efficient.   

\paragraph{Multi-plane imagery: }
Another closely related structure is layered imagery, including multi-plane imagery (MPI) and layered depth imagery (LDI)~\cite{baker1998layered, flynn2016deepstereo, zhou2018stereo, srinivasan2019pushing, shade1998layered, mildenhall2019local, flynn2019deepview, Wizadwongsa2021NeX, hedman2018deep}. 
MPI represents the scene using a stack of frontal-parallel images. It allows fast rendering and can deliver photorealistic results with slight and frontal-parallel movement. 
However, its layered geometry structure brings artifacts 360$^\circ$ surrounding views or sagittal plane movement (\eg walking or flying). 
Recent works extend the view extrapolation ability~\cite{mildenhall2019local} by fusing multiple MPIs with additional memory cost. One of the closet MPI works to us is NeX~\cite{Wizadwongsa2021NeX}. Both utilize multi-planar geometry and neural radiance function. However, NeX uses fixed, frontal-parallel planes. In contrast, ours uses learnable, slanted planes, bringing more flexibility to handle complicated scenes and render extrapolated and surrounding views. 

\begin{figure}[t]
    \centering
    \vspace{-8mm}
    \begin{subfigure}{0.315\linewidth}
        \includegraphics[width=\linewidth]{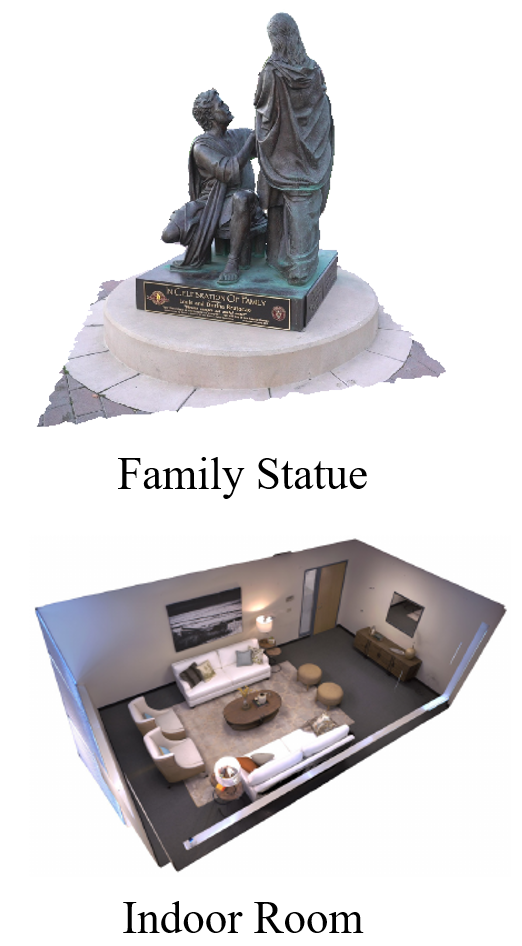}
        \caption{Target Scenes}
        \label{fig:room_statue}
    \end{subfigure}
    \begin{subfigure}{0.64\linewidth}
        \includegraphics[width=\linewidth]{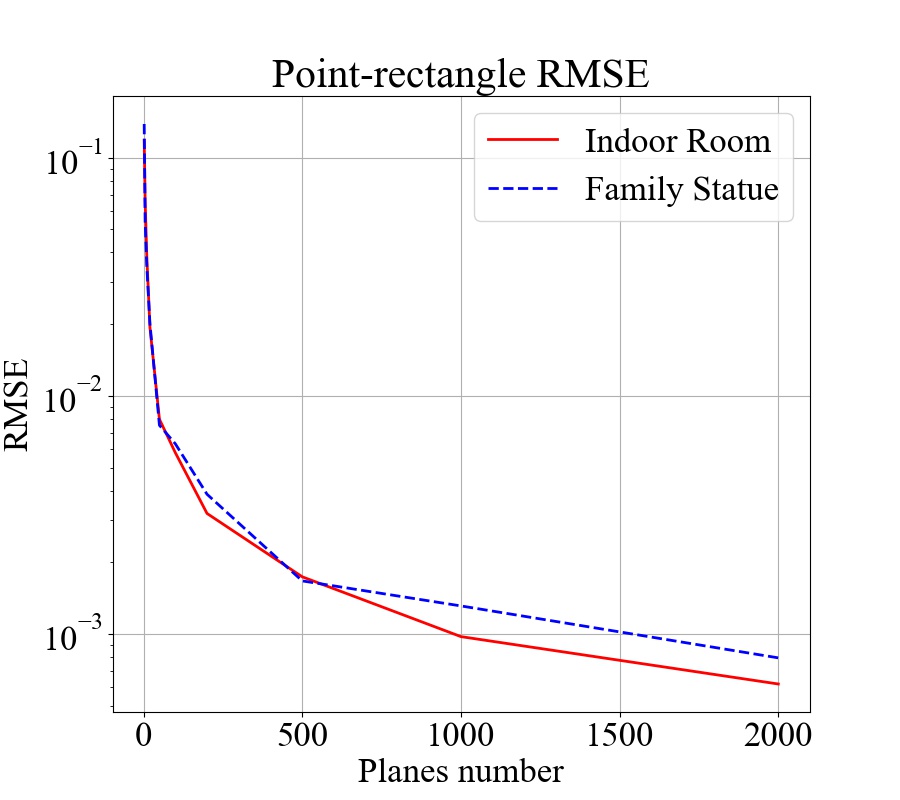}
        \caption{\# of Planes vs Point-to-Plane Error}
        \label{fig:point2rec_error}
    \end{subfigure}
        \vspace{-2mm}
    \caption{\textbf{Surface Fitting using a Mixture of Planar Sprites} Left: target scenes; Right: fitting results.}
    \vspace{-5mm}
    \label{fig:plane_fitting}
\end{figure}
\begin{figure*}[!t]
    \centering
    \vspace{-5mm}
    \includegraphics[width=.85\linewidth]{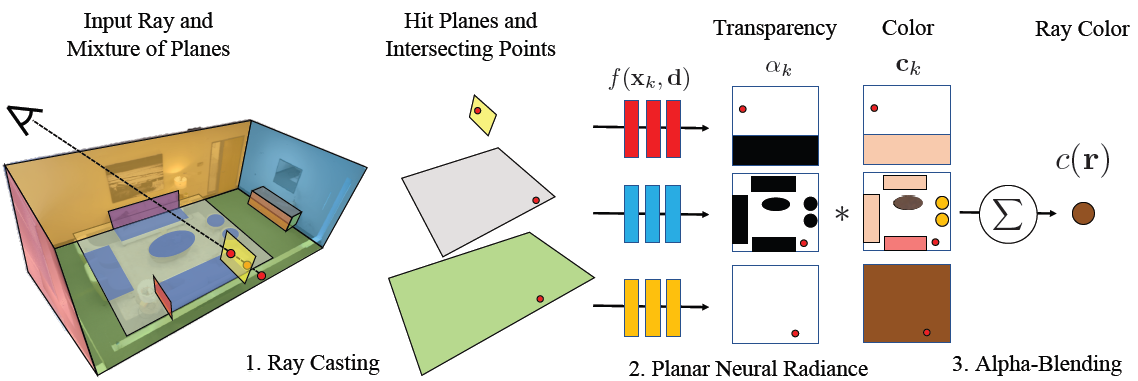}
    \vspace{-1mm}
\caption{\textbf{The Rendering Pipeline of ~\method.} We first cast rays and identify intersecting points and planes. Color and opacity can then be evaluated through each plane's neural radiance field. Finally, alpha-blending step is conducted to output the final ray color. 
    }
    \vspace{-3mm}
    \label{fig:3_method_overview}
\end{figure*}
\paragraph{Implicit surface: }
The limitations of explicit geometry could be alleviated through implicit surface modeling~\cite{park2019deepsdf}. Recent works have started to jointly model surface and appearance using neural representation~\cite{yariv2020multiview, yariv2021volume, xiang2021neutex, zhang2021ners}, achieving state-of-the-art reconstruction quality and decent view synthesis results. However, rendering the implicit function requires sequential ray marching steps, and additional steps are required to extract the surface. 

\paragraph{Neural volumetric rendering: }
Volumetric radiance fields date back to the late 90s~\cite{gortler1996lumigraph}. Recent works, such as NeRF~\cite{mildenhall2020nerf} and Neural Volumes~\cite{lombardi2019neural}, investigate deep learning for volume rendering, which combines the expressiveness of neural nets with the flexibility of volume rendering. 
A plethora of new approaches have been proposed during the past year to extend NeRF~\cite{mildenhall2020nerf, liu2020neural, barron2021mipnerf, martin2021nerf, garbin2021fastnerf, reiser2021kilonerf, hedman2021baking, wei2021nerfingmvs, zhang2020nerf++, zhang2021nerfactor}. Representative works can handle sparse input views~\cite{wang2021ibrnet, yu2021pixelnerf}, unbounded scenes~\cite{zhang2020nerf++}, overcome aliasing effect~\cite{barron2021mipnerf} and take as input unknown/noisy poses~\cite{lin2021barf}. 

The seminal NeRF \cite{mildenhall2020nerf} does not render in real-time. Several works attempt to accelerate NeRF using different strategies. For example, methods like~\cite{yu2021plenoctrees, reiser2021kilonerf, garbin2021fastnerf, rebain2021derf} choose to decompose the input scene into smaller regions and use smaller networks to model 3D geometry for each. Other approaches reduce the amount of samples per ray, through early ray termination~\cite{piala2021terminerf}, empty space skipping~\cite{yu2021plenoctrees, reiser2021kilonerf}, learnable sparse sampling ~\cite{neff2021donerf, arandjelovic2021nerf, sztrajman2021neural}, or closed-form sampling-free integration~\cite{lindell2021autoint}. Deferred rendering or baking techniques have also been used to accelerate NeRF~\cite{garbin2021fastnerf, hedman2021baking}. Our approach is a new instantiation of the above acceleration techniques through the planar representations. 
Perhaps closet to our work is MVP~\cite{lombardi2021mixture} who also take advantage of geometric primitives. There, however, exists a few key differences. First, while MVP exploits \emph{dense voxel grids} to capture the sophisticated texture of human head, we model the scene structures with \emph{planes}. Second, MVP explicitly generates RGB$\alpha$ for each voxel which is memory-consuming, whereas NeurMiPs models texture with neural nets.

\subsection{Planar Scene Representation}

We are not the first to realize the potential of multiple slanted planes for representing the scene geometry. The computer vision and graphics community has a long history of leveraging planar surfaces for modeling and rendering. Various forms of planar scene representations have been investigated~\cite{sawhney19943d, irani1996parallax, baker1998layered, szeliski19992d, kanade1980theory, fouhey2014unfolding, lee2009geometric, gupta2010blocks, hedau2010thinking}; representative works include polygon mesh~\cite{baumgart1972winged, groueix2018papier}, Marr sketch~\cite{marr}, Manhattan world~\cite{hu2021worldsheet, furukawa2009manhattan}, Binary space partitioning tree~\cite{chen2020bspnet}, 3D box layout~\cite{hedau2009recovering, hedau2010thinking}, origami theory~\cite{kanade1980theory}, slanted planes~\cite{baker1998layered, bleyer2011patchmatch, szeliski19992d}, \etc. A closely related line of research is layered sprites~\cite{szeliski19992d}, which shares a similar geometric representation to ours. 
The key difference between our work and theirs lies in the appearance representation and rendering. Layered sprites use image textures for each plane and rendering is done through homography warping. Our method in contrast exploits an expressive neural radiance field and rendering is done through ray casting, allowing us to better capture view-dependent effects and run faster in complicated scenes.

Numerous methods have been developed to reason about planar structures from images. For example, one can detect planes from images~\cite{liu2019planercnn, YuZLZG19}, recover meshes~\cite{gkioxari2019mesh}, reconstruct slanted planar surfaces from stereo~\cite{bleyer2011patchmatch, gallup2007real}, leverage planar structure for SLAM~\cite{salas2014dense}, estimate surface normal and boundary based on the local planar assumption~\cite{fouhey2014unfolding} and finally reconstruct planar spirits from multiple images~\cite{sawhney19943d, irani1996parallax}. The proposed \method~can be treated as a multi-view slanted plane reconstruction method by minimizing the photo-metric rendering loss. 
\begin{table*}[h]
\centering
\vspace{-5mm}
{
\setlength\tabcolsep{1.5pt} 
\scalebox{0.78}{
\begin{tabular}{cccccccc}
\includegraphics[width=0.15\textwidth]{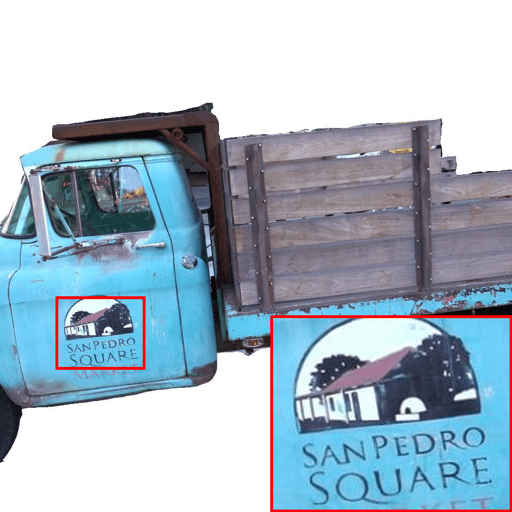} &
\includegraphics[width=0.15\textwidth]{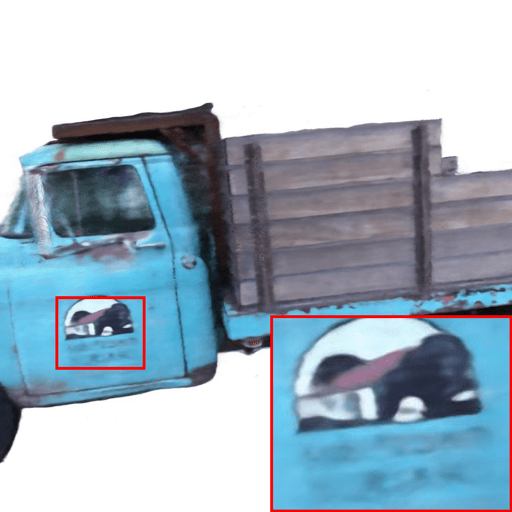} &
\includegraphics[width=0.15\textwidth]{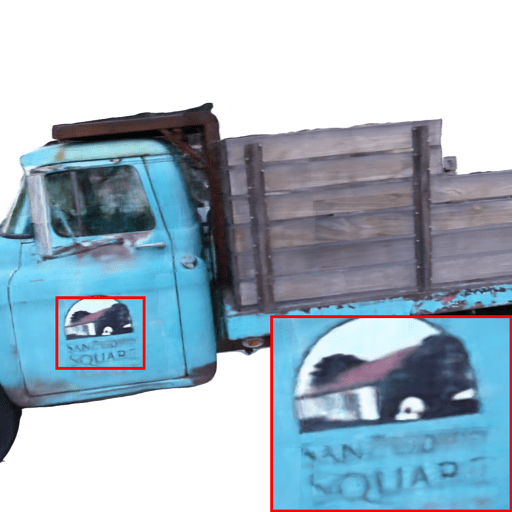} &
\includegraphics[width=0.15\textwidth]{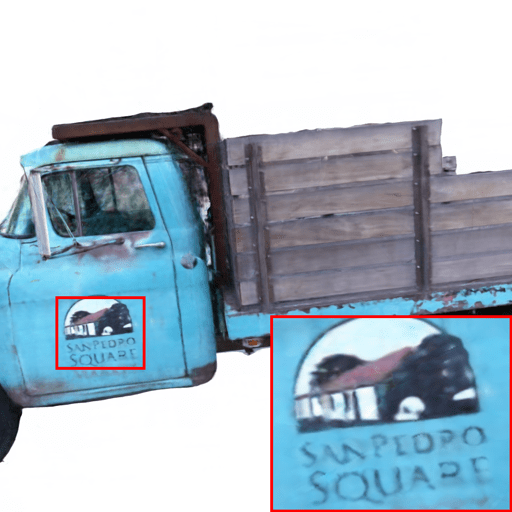} &
\includegraphics[width=0.15\textwidth]{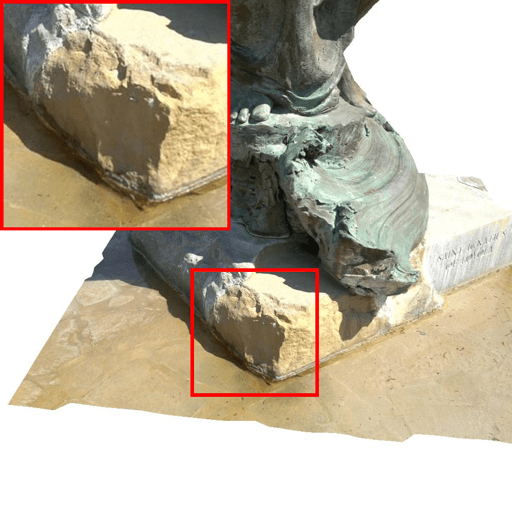} &
\includegraphics[width=0.15\textwidth]{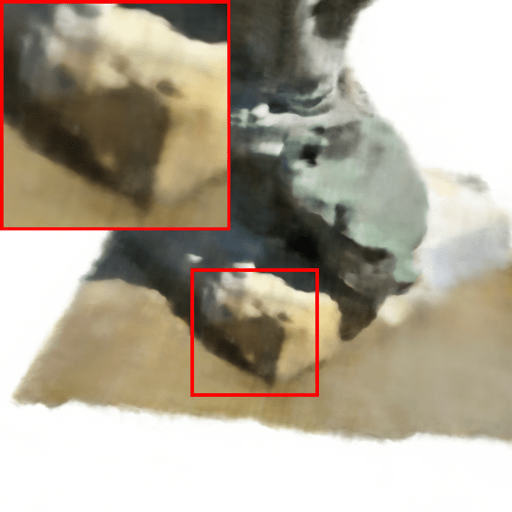} &
\includegraphics[width=0.15\textwidth]{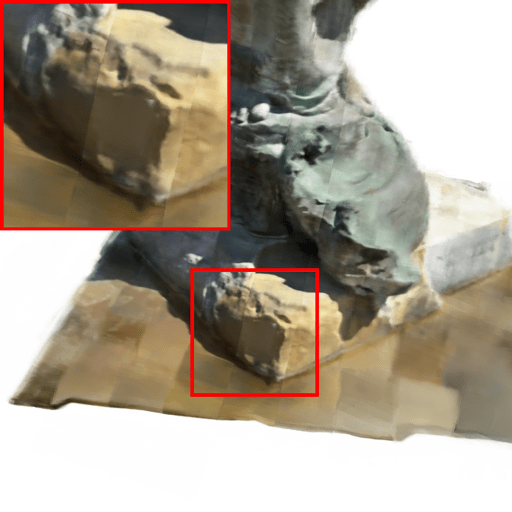} &
\includegraphics[width=0.15\textwidth]{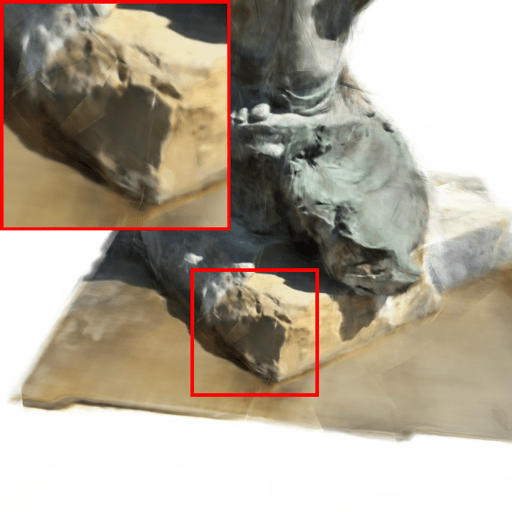}
\\
\includegraphics[width=0.15\textwidth]{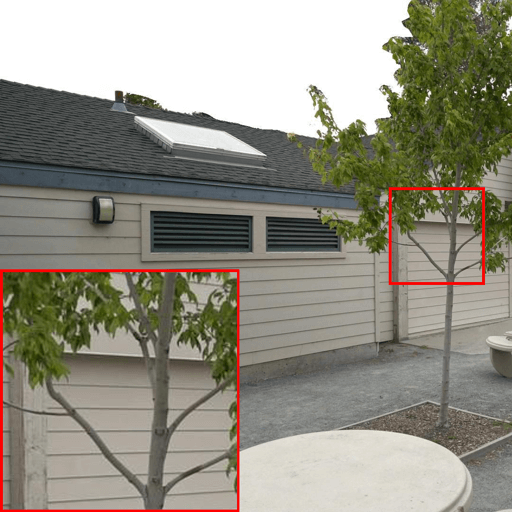} &
\includegraphics[width=0.15\textwidth]{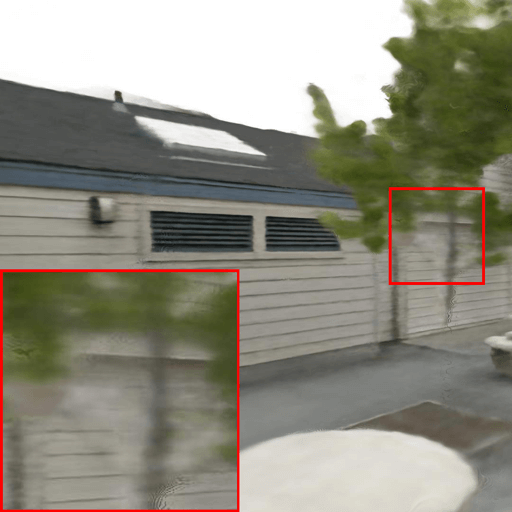} &
\includegraphics[width=0.15\textwidth]{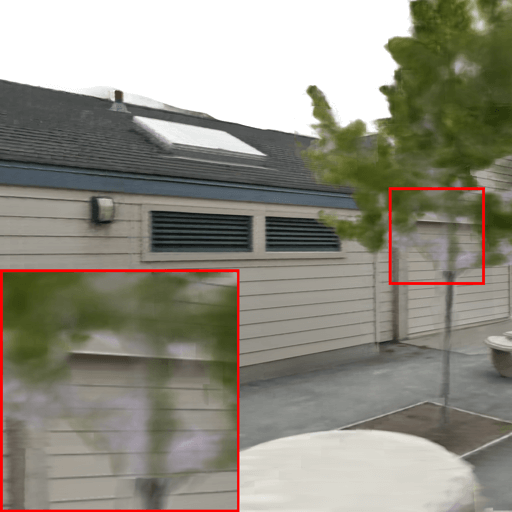} &
\includegraphics[width=0.15\textwidth]{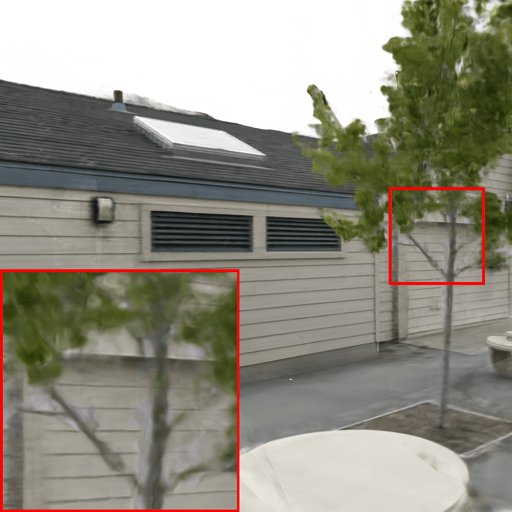} &
\includegraphics[width=0.15\textwidth]{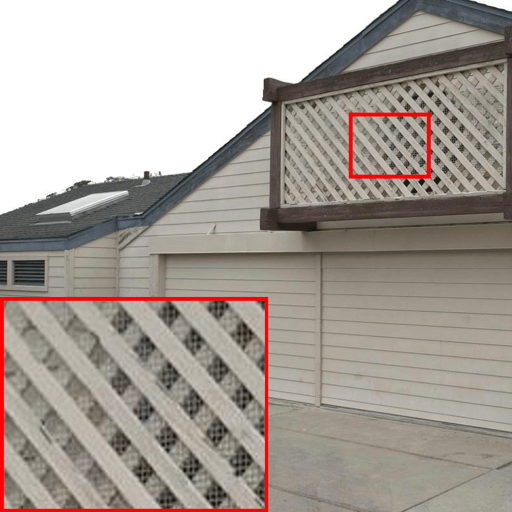} &
\includegraphics[width=0.15\textwidth]{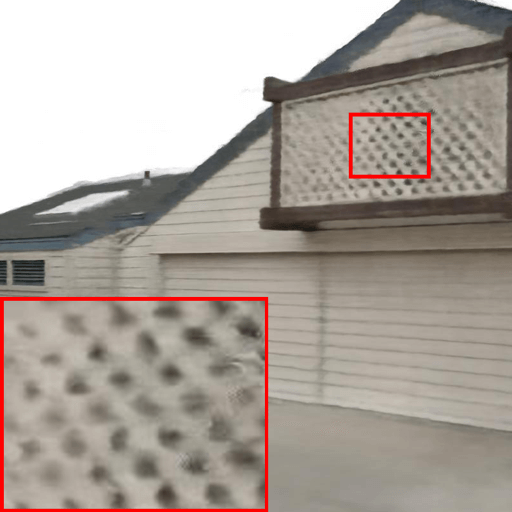} &
\includegraphics[width=0.15\textwidth]{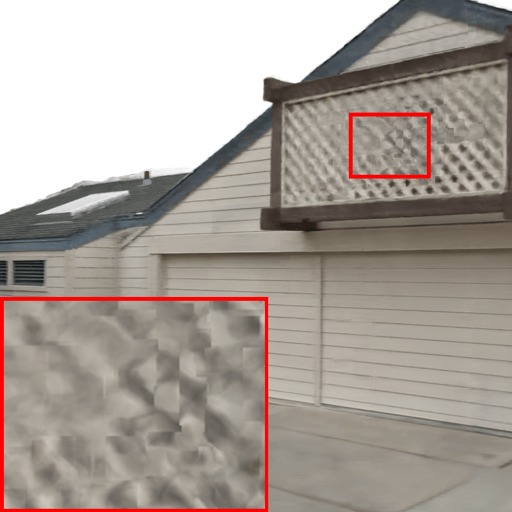} &
\includegraphics[width=0.15\textwidth]{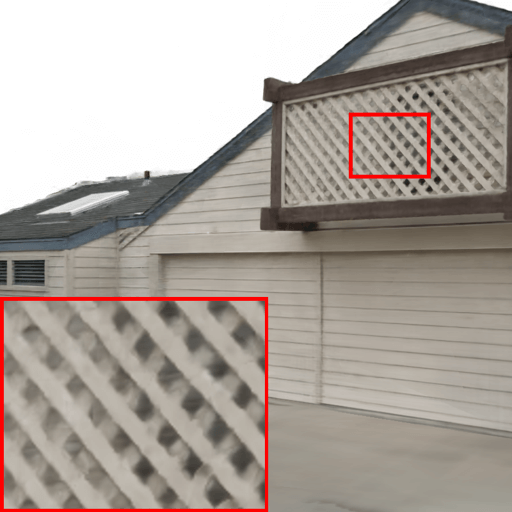}
\\
{\large GT} & {\large NeRF}&  {\large KiloNeRF}&  {\large Ours} & {\large GT} & {\large NeRF}& {\large KiloNeRF}& {\large Ours}
\end{tabular}
}
}
\vspace{-3mm}
\captionof{figure}{\textbf{Qualitative Results of Tanks \& Temples.} Zoom in for better visual comparisons. 
}
\label{tab:4_quality_tanks}
\vspace{-5mm}
\end{table*}

\section{Method}

In this work, we tackle the problem of novel view synthesis. Our goal is to improve the rendering efficiency as much as possible while improving the rendering quality at extreme novel views. Towards this goal, we propose a novel neural representation called the mixture of planer experts and design a neural rendering method using NeurMiPs. 

Specifically, we first represent the scene as a mixture of local planar surfaces. Every local surface is an oriented 2D rectangle in 3D.  We then use a neural radiance field function for each plane to encode its view-dependent appearance and transparency. Both the geometry and the radiance fields are learned end-to-end from the input images. During the rendering time, our method will first conduct a ray-rectangle intersection check. Each ray will only hit a small subset of surfaces. The color and transparency will be evaluated based on intersecting point's coordinate. Finally, the ray color will be calculated through alpha blending the colors of all intersecting points.

Fig.~\ref{fig:comparison} compares different 3D representations for view synthesis. Compared to neural surface rendering, our method is efficient in both memory and computation; the approach is significantly more sample-efficient than volume rendering with better extrapolation; compared to multi-plane imagery, our approach better reflects the geometry.

\subsection{Mixture of Planar Experts}
\label{sec:representation}
The mixture of planar experts representation consists of $K$ rectangular surface parameterized by $\{\mathbf{s}_k=(\mathbf{p}_k, \mathbf{n}_k, \mathbf{u}_k, w_k, h_k) \}$, where $\mathbf{p}_k$ is the rectangle center; $\mathbf{n}_k$ is a normalized 3D vector representing the plane's normal; $\mathbf{u}_k$ is the normalized up vector defines the y-axis direction of the plane coordinate in the world; $(w_k, h_k)$ is the plane's size. 
Inspired by the success of recent neural rendering~\cite{mildenhall2020nerf, Wizadwongsa2021NeX}, we represent each planar expert's view-dependent appearance and transparency as a 3D neural radiance field function 
\begin{equation}
\label{eq:radiance}
(\mathbf{c}_k, \alpha_k) = f_k(\mathbf{x}_{k}, \mathbf{d})
\end{equation}
{The function takes the 3D space coordinate $\mathbf{x}_k$ and normalized 3D ray direction $\mathbf{d} = (d_x, d_y, d_z) \in \mathbb{S}^2$ as input} and output the corresponding color and transparency. See Fig.~\ref{fig:parameterization} for an illustration of the planar representation.

\paragraph{Network Architecture}
{Each planar sprite only needs to model a local slice of the full radiance field.}
Hence, we use a significantly smaller multi-layer perceptron (MLPs) for each planar expert model. 
Each MLP model consists of three fully-connected hidden layers, with ReLU activation for each hidden layer and sigmoid activation for final output. The networks predict both color and alpha values. Following recent works~\cite{mildenhall2020nerf, tancik2020fourier}, the ray input is transformed into a higher dimensional space with high-frequency functions before passing to the network, which enhances the capability of capturing high-frequency textures.

\paragraph{Representing the Scene Geometry} A natural question arises: is it sufficient for representing the complicated world with a mixture of planes? To answer this question, we conduct a quick experiment to demonstrate its power. Specifically, we choose two complicated 3D scenes, an indoor environment from Replica dataset~\cite{straub2019replica} and an outdoor scene from Tanks and Temple~\cite{knapitsch2017tanks}. Both scenes consist of sufficiently complicated geometric structures like trees, poles, circular shape surfaces. 
We use a mixture of planar experts to fit the surface geometry, by minimizing the point to plane distance for points sampled from the scene's surface. 
Fig.~\ref{fig:plane_fitting} illustrates the local planar surface fitting performance as a curve of the number of rectangles vs. average point-to-plane distance. From the figure, we could see that with only {1000 planes, we could reach $10^{-3}$ RMSE point-to-plane error, with the whole scene normalized into a unit sphere}. These results suggest the multi-plane surface geometry is suitable to represent complicated scenes for neural rendering. 

\begin{table*}[h]
\vspace{-5mm}
\centering
{
\setlength\tabcolsep{1.5pt} 
\scalebox{0.65}{
\begin{tabular}{ccccc|ccccc} 
\includegraphics[width=0.145\textwidth]{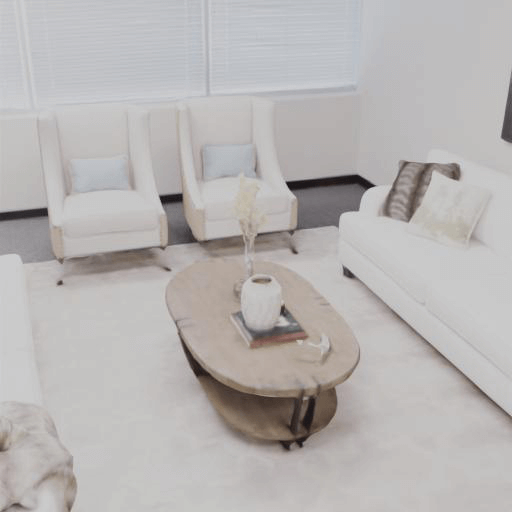}&
\includegraphics[width=0.145\textwidth]{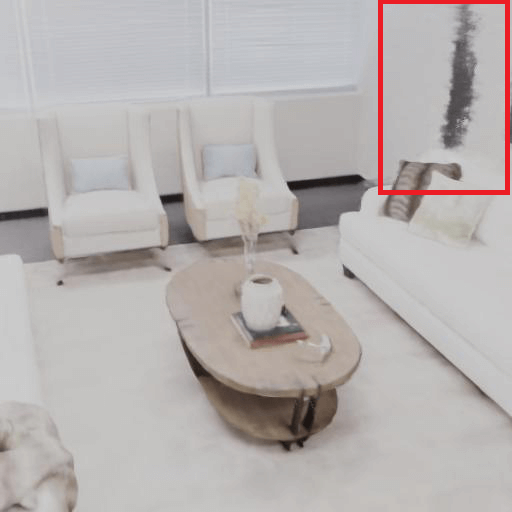}&
\includegraphics[width=0.145\textwidth]{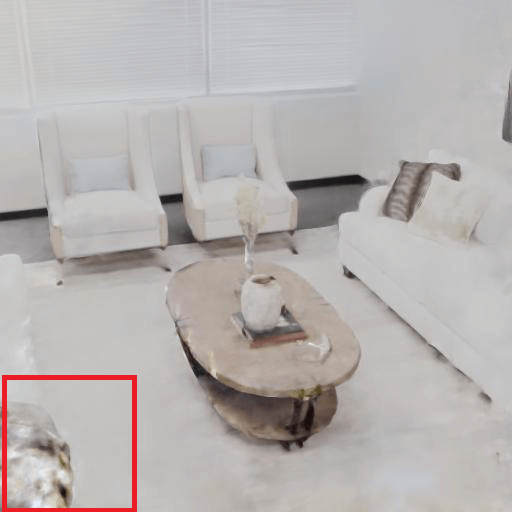}&
\includegraphics[width=0.145\textwidth]{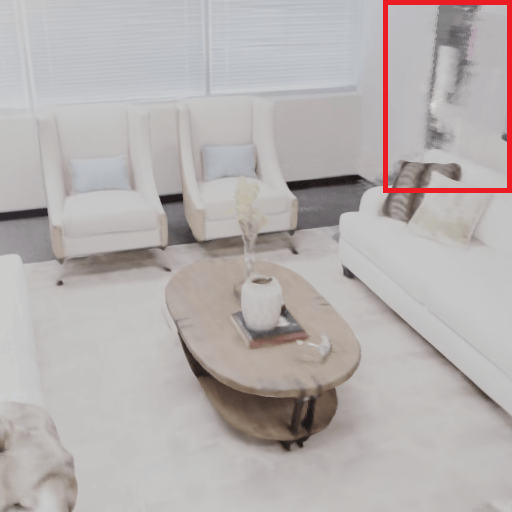}&
\includegraphics[width=0.145\textwidth]{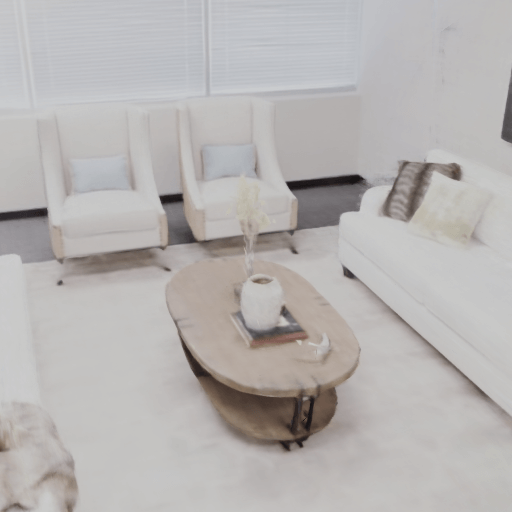}&
\includegraphics[width=0.145\textwidth]{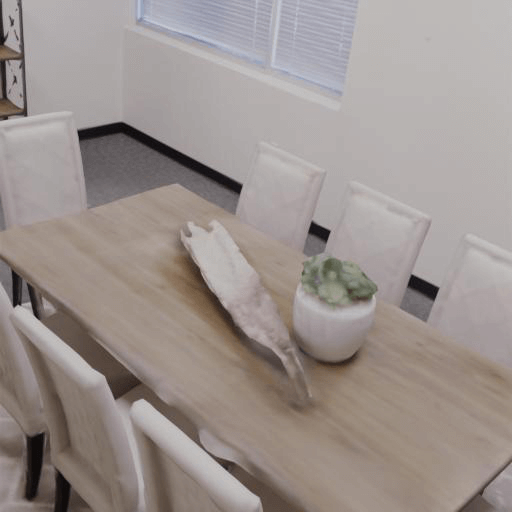}&
\includegraphics[width=0.145\textwidth]{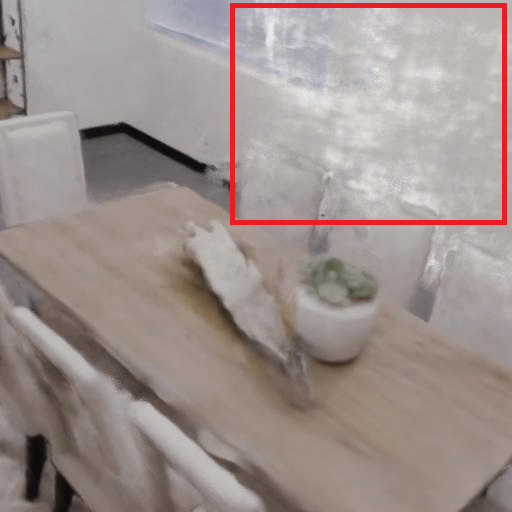}&
\includegraphics[width=0.145\textwidth]{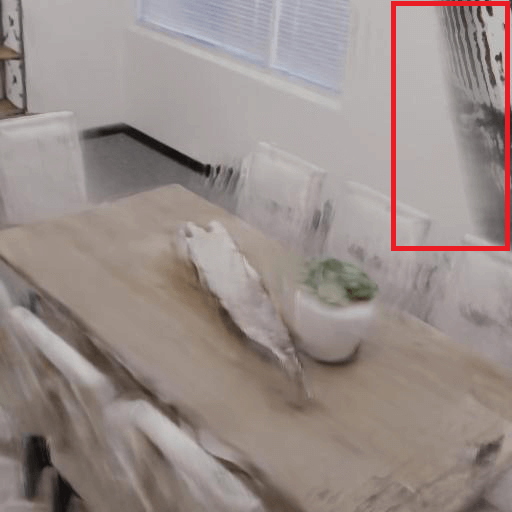}&
\includegraphics[width=0.145\textwidth]{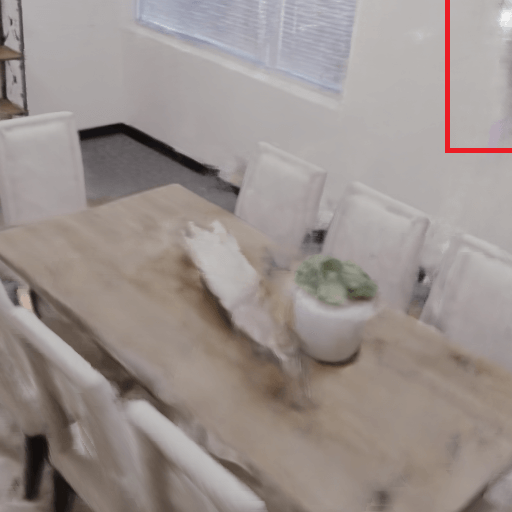}&
\includegraphics[width=0.145\textwidth]{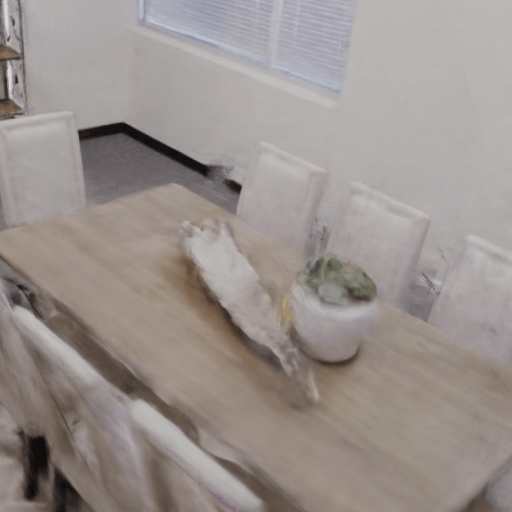}
\\
\includegraphics[width=0.145\textwidth]{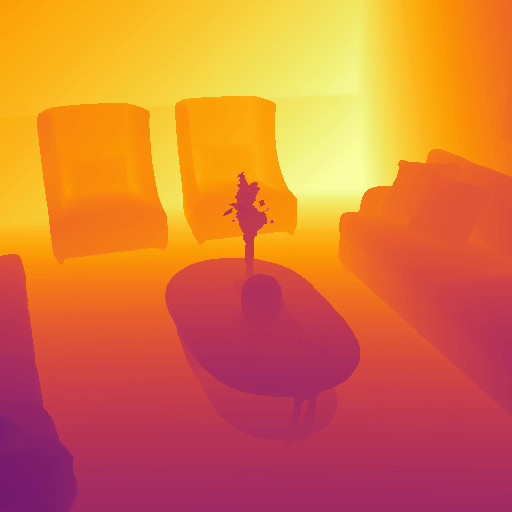}&
\includegraphics[width=0.145\textwidth]{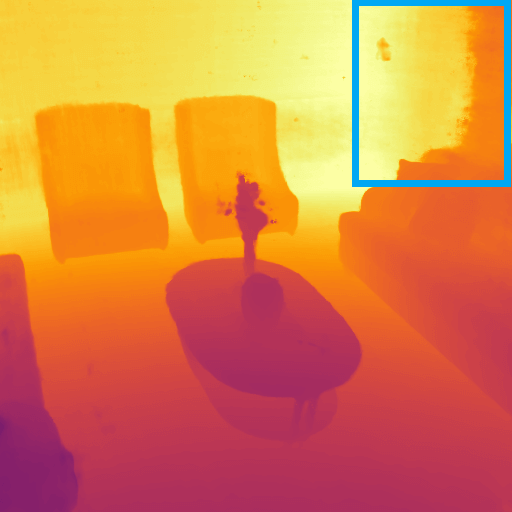}&
\includegraphics[width=0.145\textwidth]{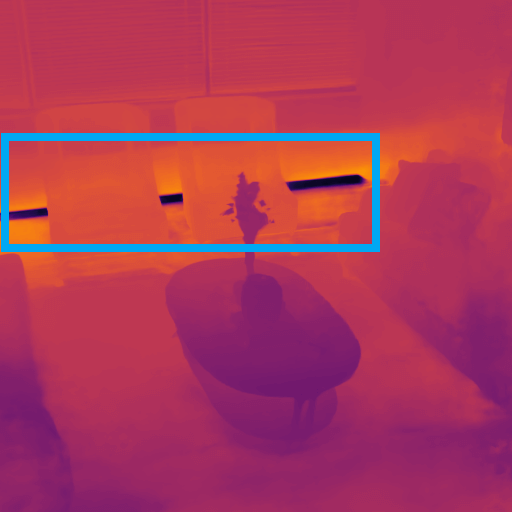}&
\includegraphics[width=0.145\textwidth]{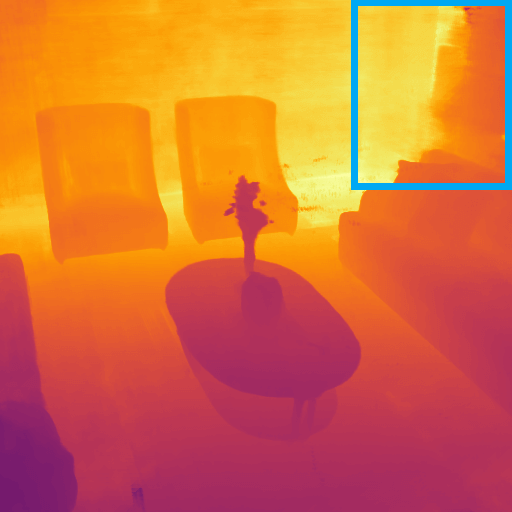}&
\includegraphics[width=0.145\textwidth]{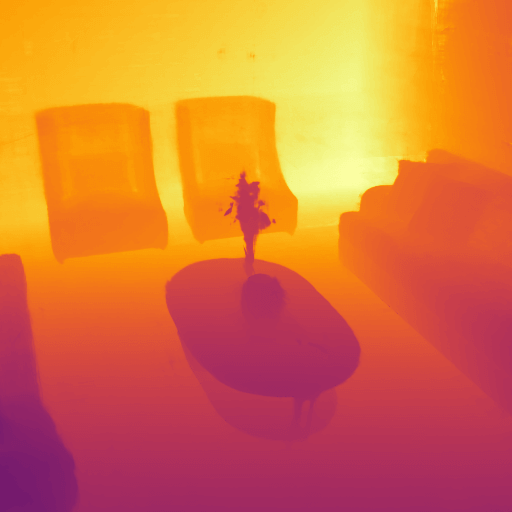}&
\includegraphics[width=0.145\textwidth]{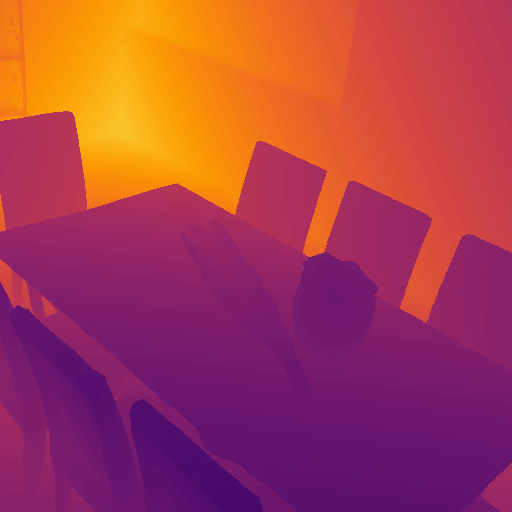}&
\includegraphics[width=0.145\textwidth]{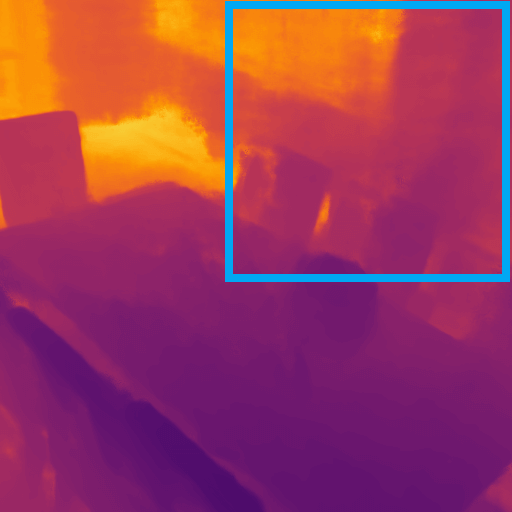}&
\includegraphics[width=0.145\textwidth]{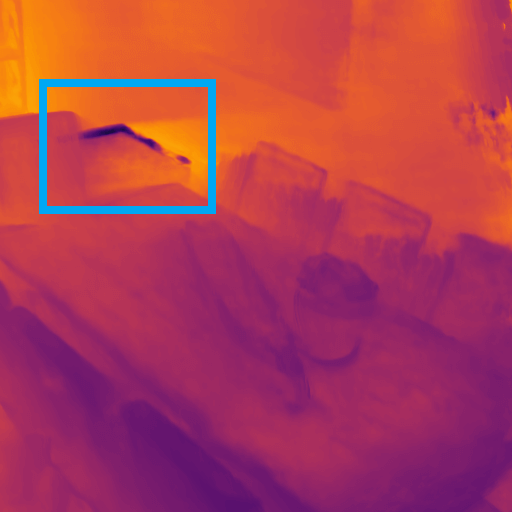}&
\includegraphics[width=0.145\textwidth]{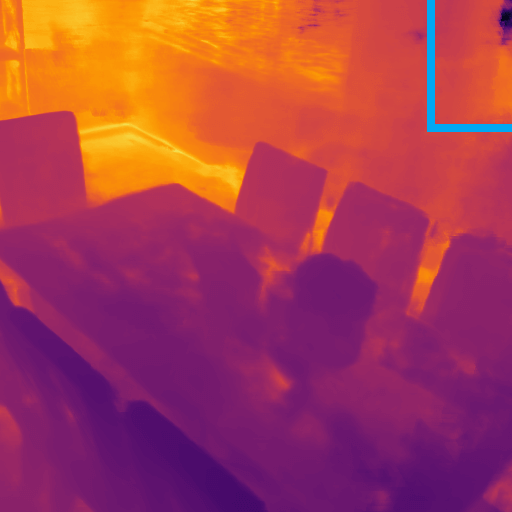}&
\includegraphics[width=0.145\textwidth]{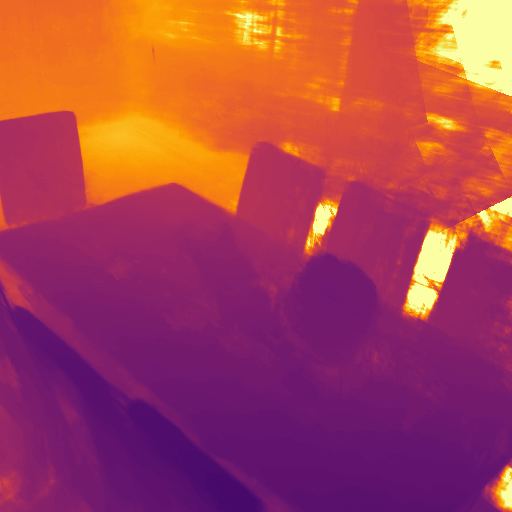}
\\
{\Large GT} & {\Large NeRF}&  {\Large NeX}& {\Large KiloNeRF}&  {\Large Ours} & {\Large GT} & {\Large NeRF}& {\Large NeX}& {\Large KiloNeRF}& {\Large Ours}
\end{tabular}
}
}
\vspace{-3mm}
\captionof{figure}{\textbf{Qualitative Results of the Replica Dataset.}
}
\vspace{-5mm}
\label{tab:4_quality_replica}
\end{table*}
\subsection{Rendering}

At the testing time, a pixel is rendered by shooting a ray from the eye and evaluating the radiance along the ray. Formally, the input ray consists of a original point and a normalized directional vector $\mathbf{r} = \{ \mathbf{o}, \mathbf{d} \}$ where $\mathbf{o}$ is the origin and $\mathbf{d}$ is the direction. 

\paragraph{Ray-Plane Intersection} 
Rendering \method~is very efficient thanks to its simple geometry. The first step is the ray-rectangle intersection: decide whether a local plane is intersecting the ray. This intersection check can be done analytically. Firstly, we will find the intersection between a given ray $\mathbf{r} = \{ \mathbf{o}, \mathbf{d} \}$ and an infinite size plane $\{ \mathbf{n}_k, \mathbf{p}_k\}$:
\begin{equation}
    \label{eq:ray-plane-intersection}
\mathbf{x}_{k} = \mathbf{o} + \frac{(\mathbf{p}_k - \mathbf{o})\cdot \mathbf{n}_k}{\mathbf{d} \cdot \mathbf{n}_k} \mathbf{d}
\end{equation}
We will only keep the intersected rectangles for the next phase. In practice, only a small fraction will be kept. 

\paragraph{Radiance Evaluation}
We then evaluate the ray's color and transparency at the intersecting planar experts. Specifically, given the input coordinate transform $\mathbf{x}_{p, k}$ and the direction $\mathbf{d}$ for each planar expert, we will output a transparency value $\alpha_k$ and a color $\mathbf{c}_k$ by evaluating its neural radiance function, according to Eq.~\ref{eq:radiance}. 

\paragraph{Alpha Composition}
Now, each ray has collected a set of of point samples $\{ \mathbf{x}_j, \mathbf{c}_j, \alpha_j \}$. We sort the point from closest to furthest to the eye $\mathbf{o}$, and conduct alpha composition to get the final estimation of the ray's color:
\begin{equation}\label{eq:alpha}
    c(\mathbf{r}) = \sum_j \prod_i^{j-1} (1-\alpha_i) \alpha_j \mathbf{c}_j
\end{equation}
Fig.~\ref{fig:3_method_overview} depicts the detailed procedure of our rendering process per each ray. We want to highlight three important properties of our method. First, thanks to our mixture-of-planar structure, the ray-geometry intersection is computed efficiently in closed-form, which is extremely efficient. Second, each ray will only hit a handful of planes. This results in a small number of samples we need to evaluate for each ray, significantly boosting the speed. Finally, each planar expert radiance function only needs to model a local surface. Hence, the required computation for the radiance MLPs is also significantly smaller than NeRF. 

\subsection{Training}
\label{sec:training-tech}

Training \method~requires jointly optimizing the plane geometry $\{(\mathbf{c}_k, \mathbf{n}_k, \mathbf{u}_k, w_k, h_k) \}$ and the radiance $f_k(\cdot, \cdot)$. Training from scratch results in artifacts and low-fidelity geometric structures. In practice we observed better results can be acquired through two training techniques: 1) planar initialization through geometric loss minimization; 2) distillation from a large teacher radiance model.
\paragraph{{Plane Initialization}}

We initialize the plane geometric parameters $\{ \mathbf{s}_k \}$
using the coarse 3D point cloud $\{\mathbf{x}_i\}$ estimated by structure-from-motion~\cite{schoenberger2016sfm}. Specifically, we optimize the following point-to-rectangle distance function:
\begin{equation}
\label{eq:geometry}
 \mathcal{L}_{g} = \sum_i \min_k d(\mathbf{x}_i, \mathbf{s}_k ) + \lambda \sum_k (w_k h_k)^2,
\end{equation}
where $\min_k d(\mathbf{x}, \mathbf{s_k})$ is the distance from point $x$ to the closest rectangular surface $s_k$. An area regularization $(w_k h_k)^2$ is adopted to forbid the rectangle to be arbitrarily large.

\paragraph{Distillation}
After the planar geometry initialization, we then jointly optimize radiance and geometry. Inspired by the success of NeRF distillation~\cite{reiser2021kilonerf, yu2021plenoctrees, su2021nerf}, we first train a large-capacity, ordinary NeRF as the teacher model to distill knowledge from. It follows standard NeRF neural network architecture with two noticeable differences. First, the point sampling is conducted through \method's ray-planar intersection. Second we also jointly optimize the planar experts' geometric parameters. The joint geometry and photo-metric loss are used for training the planar-guided NeRF: 
\[
\mathcal{L}_{total} = \mathcal{L}_g + \mathcal{L}_c,
\]
where $\mathcal{L}_g$ is the point-to-rectangle geometry loss defined in Eq.~\ref{eq:geometry} and $\mathcal{L}_c$ is the L2-photometric loss
\begin{equation}
    \mathcal{L}_c = \sum_\mathbf{r} \| c(\mathbf{r}) - c_\mathrm{gt}(\mathbf{r}) \|_2^2.
\end{equation}
For each planar expert, we then distill knowledge from the teacher model by minimizing the difference between the teacher's output and each student's output. Specifically, we draw random points uniformly from the rectangle and random view directions from the half unit sphere for each batch. Student network's parameters are updated by minimizing the L2 loss of both alpha values and the colors. 

\paragraph{Fine-Tuning} After over-fitting to the teacher network, we will fix plane parameters and fine-tune our student radiance field models to further improve the rendering quality. Specifically, we minimize the L2-photometric loss $\mathcal{L}_c$ between the rendered pixel color and the ground-truth color.

\subsection{Implementation}
\label{setc:implementaiton}
Despite being efficient by design, \method~benefits from several techniques during implementation to further boost the rendering speed which makes it real-time.  

\paragraph{Alpha Baking}
Another acceleration technique is texture pre-baking. 
Inspired by prior work on pre-caching NeRF~\cite{garbin2021fastnerf, Wizadwongsa2021NeX, yu2021plenoctrees}, we propose to pre-render alpha values and bake them as alpha textures for each rectangular plane. {To be more specific, the view-independent alpha is baked into each plane $i$ as texture maps $A_i$ of size $w \times h$. During inference, when a ray hits a surface, we could retrieve its corresponding alpha value by bilinear interpolating from the baked alpha texture $A_i$.} Note that applying alpha baking directly during inference might bring a minor rendering performance drop. {Therefore, we fine-tune the RGB branch of planar experts for better rendering quality.}

\paragraph{Early Ray Termination} 
Evaluating radiance for all intersecting rectangles is unnecessary, in particular when the transmittance value $\prod_i (1 - \alpha_i)$ is close to zero (\eg ray hitting a non-transparent plane) since it may only have a minor impact. In practice, we exploit early ray termination to avoid additional network evaluation, thereby enhancing rendering efficiency considerably.
We also perform re-normalization if the sum of the alpha values is smaller than 1. We empirically observe that it will improve performance.

{\paragraph{Custom CUDA kernel} To further accelerate model inference, we implement a custom CUDA kernel for ray-plane intersection, model inference, and alpha composition. We fuse the network evaluation of each expert into a single CUDA kernel so that all experts can render in parallel. As we will show in Sec. \ref{sec:analysis} and supp. materials, this improves rendering efficiency significantly.}

\begin{table}[t]
\centering
\small

\scalebox{0.9}
{
\begin{tabular}{ l|c|c|c} 
 \hline
 Model & PSNR$\uparrow$ & SSIM $\uparrow$ & LPIPS $\downarrow$ \\
 \hline 
 NeRF~\cite{mildenhall2020nerf}           & 28.32 & 0.904 & 0.168 \\
 SRN~\cite{sitzmann2019scene}             & 24.10 & 0.847 & 0.251 \\
 Neural Volumes~\cite{lombardi2019neural} & 23.70 & 0.834 & 0.260 \\
 NSVF~\cite{liu2020neural}                & 28.40 & 0.900 & 0.153 \\
 PlenOctrees~\cite{yu2021plenoctrees}      & 27.99 & \textbf{0.917} & 0.131 \\ 
 KiloNeRF~\cite{reiser2021kilonerf}       & 28.41 & 0.910 & 0.090 \\
 \hline
 Ours            & \textbf{28.46} & 0.908 & \textbf{0.089} \\
 \hline
\end{tabular}
}
\vspace{-3mm}
\caption{\textbf{Quantitative comparison on Tanks~\&~Temples.}}
\label{tab:4_quantity_tanks}
\vspace{-1mm}

\end{table}
\section{Experiments}
\subsection{Experimental Setup}
\label{sec:exp-setup}
\paragraph{Datasets:}
We evaluate \method~on two challenging datasets: {Tanks \& Temples} \cite{knapitsch2017tanks} and {Replica} \cite{straub2019replica}. 
Tanks \& Temples consists of five bounded \emph{real-world} scenes~\cite{liu2020neural}. Each scene contains $152\sim384$ high-resolution images ($1920 \times 1080$) captured from surrounding 360$^\circ$ viewpoints.

Replica is a \emph{synthetic} dataset featuring a diverse set of indoor scenes. Each scene is equipped with high-quality geometry and photorealistic textures, allowing one to render high-fidelity images from arbitrary camera poses. 
The flexibility also allows us to generate challenging novel view synthesis scenarios that are not available in existing benchmarks (\eg, extreme view extrapolation).
In this work, we randomly select seven scenes and render 50 training images and 100 test images for each of them. 
The camera poses are sampled randomly within a pre-defined range. We adopt a wider range for the test split so that the test images can cover broader views and may include unseen areas. The setup allows to evaluate the extrapolation capability of existing approaches.
We adopt BlenderProc~\cite{denninger2019blenderproc} as our physical-based rendering engine. The image size is set to $512 \times 512$.

\paragraph{Metrics:} Following~\cite{reiser2021kilonerf, mildenhall2020nerf}, we adopt peak signal-to-noise ratio (PSNR), structural similarity index (SSIM)~\cite{wang2004image}, and perceptual metric (LPIPS)~\cite{zhang2018unreasonable} for quantitative evaluation.

\paragraph{Baselines:}
We compare our approach against state-of-the-art neural radiance field (\ie, NeRF~\cite{mildenhall2020nerf}), MPI-based method (\ie, NeX~\cite{Wizadwongsa2021NeX}), and hybrid, real-time methods (\ie, NVSF~\cite{liu2020neural}, KiloNeRF~\cite{reiser2021kilonerf}, PlenOctrees~\cite{yu2021plenoctrees}). 
We refer the readers to supp. materials for more details.

\paragraph{Implementation details:}
For each scene, we initialize our plane geometry with the sparse point clouds from COLMAP~\cite{schoenberger2016sfm, schoenberger2016mvs}. 
Plane centers $\{p_k\}$ are selected by farthest point sampling, and plane orientations $\{n_k, u_k\}$ are initialized as normals estimated from local point sets around $p_k$. 
We set the number of planar experts to be 500 for Replica and 1000 for Tanks and Temple.  
We first train the teacher model for 6K epochs, then we distill the planar experts for 1.5K epochs. Finally, we fine-tune the experts for 2.5K epochs.
We use Adam~\cite{kingma2014adam} optimizer with a learning rate of $5\times10^{-4}$ across all experiments.
\begin{table}[t]
\centering
\small

{
\scalebox{0.99}{
\begin{tabular}{ l|c|c|c} 
 \hline
 Model & PSNR$\uparrow$ & SSIM$\uparrow$ & LPIPS$\downarrow$\\
 \hline 
 NeX~\cite{Wizadwongsa2021NeX}                  & 24.76 & 0.832 & 0.152 \\
 NeRF~\cite{mildenhall2020nerf}                 & 30.12 & 0.901 & 0.097 \\ 

 PlenOctrees*~\cite{yu2021plenoctrees}    & 27.72 & 0.872 & 0.174 \\
 KiloNeRF*~\cite{reiser2021kilonerf}  & 29.37 & \textbf{0.904} & 0.097 \\
 
 \hline
 Ours           & \textbf{30.80} & 0.900 & \textbf{0.088} \\
 \hline
\end{tabular}
}

}
\vspace{-2mm}
\caption{{\textbf{Quantitative comparison on Replica.} All models are evaluated on a single TITAN RTX. Please see text for more details.}}
\vspace{-3mm}
\label{tab:4_quantity_replica}
\end{table}
\begin{table*}[t!]
    \centering
    \begin{minipage}{0.57\linewidth}
        \setlength\tabcolsep{4pt}
        \def\arraystretch{1.05}
        \resizebox{\linewidth}{!}{
            \begin{tabular}{l|ccc|ccc|ccc}
            \hline 
                & \multicolumn{3}{c}{Easy} & \multicolumn{3}{c}{Medium}  & \multicolumn{3}{c}{Hard} \\
                & PSNR $\uparrow$    & SSIM $\uparrow$   & LPIPS $\downarrow$   & PSNR $\uparrow$  & SSIM $\uparrow$ & LPIPS $\downarrow$   & PSNR $\uparrow$  & SSIM $\uparrow$      & LPIPS $\downarrow$   \\
            \hline 
            NeX~\cite{Wizadwongsa2021NeX}      & 25.88 & 0.844 & 0.136 & 24.56 & 0.828 & 0.156 & 23.86 & 0.826 & 0.164\\ 
            NeRF~\cite{mildenhall2020nerf}     & 31.41 & \textbf{0.915} & 0.081 & 29.98 & \textbf{0.901} & 0.097 & 29.02 & 0.887 & 0.113 \\
            PlenOctree~\cite{yu2021plenoctrees}& 28.05 & 0.877 & 0.171 & 27.613 & 0.872 & 0.174 & 27.51 & 0.866 & 0.179\\
            KiloNeRF~\cite{reiser2021kilonerf} & 28.00 & 0.908 & 0.085 & 28.06 & 0.893 & 0.105 & 27.50 & 0.886 & 0.117 \\
            \hline
            Ours   & \textbf{31.98}  & 0.914  & \textbf{0.074} & \textbf{30.31} & 0.894 & \textbf{0.092} & \textbf{30.05} & \textbf{0.892} & \textbf{0.096} \\ 
            \hline 
            \end{tabular}

        }
        \captionof{table}{\textbf{Performance vs viewpoint difference.}
        }
        \label{tab:4_quantity_replica_difficulty}   
    \end{minipage}
    \hspace{8px}
    \begin{minipage}{0.4\linewidth}
        \setlength\tabcolsep{4pt}
        \def\arraystretch{1.05}
        \resizebox{\linewidth}{!}{
            
            \begin{tabular}{l|c|c|c|c}
                \hline 
                Model & Median $\downarrow$ & \makecell{Inlier (\%)\\ $<0.05$} & \makecell{Inlier (\%)\\ $<0.10$} & \makecell{Inlier (\%)\\ $<0.50$}\\
                \hline
                NeX~\cite{Wizadwongsa2021NeX}      & 0.767 & 3.3 & 6.6  & 33.3\\
                NeRF~\cite{mildenhall2020nerf}     & 0.137 & 19.6 & 38.3  & \textbf{87.8}\\
                KiloNeRF~\cite{reiser2021kilonerf} & 0.125 & 32.9 & 45.9  & 77.2\\
                \hline
                Ours                               & \textbf{0.066} & \textbf{43.7} & \textbf{59.0} & 84.2\\
                 \hline
            \end{tabular}            

        }
        \captionof{table}{\textbf{Peformance of estimated depth.}
        }
        \label{tab:4_depth_metrics}        
    \end{minipage}
    \vspace{-4pt}
\end{table*}

\subsection{Tanks \& Temples}
As shown in Tab.~\ref{tab:4_quantity_tanks}, our approach is comparable to or better than prior art on all three metrics. 
Specifically, while NeRF generates blurry textures for large-scale scenes and KiloNeRF produces blocking artifacts, \method~is able to capture detailed textures on planar surfaces with sharp boundaries. Furthermore, our local planar structure can handle non-planar and thin objects well through a combination of planar experts and alpha composition. See the tree branches and leaves in the \emph{Barn} example in Fig. \ref{tab:4_quality_tanks}.

\subsection{Replica}
We further evaluate our approach on Replica. Since Replica consists of extrapolated views (as mentioned in Sec. \ref{sec:exp-setup}) that have not been observed during training, previous voxel-based implicit methods such as PlenOctrees \cite{yu2021plenoctrees}, KiloNeRF \cite{reiser2021kilonerf} suffer drastically. Those methods prune out redundant voxels during training. Therefore, they cannot estimate the appearances of unseen regions. To (partially) alleviate this issue, we reduce the pruning threshold so that voxels are preserved even if they have lower volume density, at the cost of larger memory footprint/slower inference speed. In contrast, \method~represents the scene geometry with multiple planes, and can generalize better in view extrapolation. As shown in Tab. \ref{tab:4_quantity_replica} and Tab. \ref{tab:4_replica_speed}, our approach reaches the best quality-efficiency trade-off and is comparable to or better than prior art on all three metrics. We refer the readers to the supp. material for implementation details of the baselines and more comprehensive comparison. 
We also show some qualitative results in Fig.~\ref{tab:4_quality_replica}. NeRF produces fog-like artifacts in free space, while NeX~\cite{Wizadwongsa2021NeX} has significant ``stack-of-cards'' effects. \method, in contrast, has significantly better visual quality, even at extrapolated novel views.

\subsection{Analysis}
\label{sec:analysis}
\paragraph{Performance vs viewpoint difference:} To gain insights into when \method~performs the best, we divide the test split of Replica into three categories --- easy, medium, and hard --- based on the proximity to the nearest training views, and evaluate our model. As shown in Tab.~\ref{tab:4_quantity_replica_difficulty}, \method~is comparable to or better than competing methods across all settings. In particular, \method~improves the performance the most when the viewpoint difference is large (\ie, {1.03} dB PSNR gain and {0.017} LPIPS score reduction).

\paragraph{Depth estimation: } To verify how well \method~models the scene geometricy, we follow previous work \cite{mildenhall2020nerf} to generate depth map at each viewpoint and compare with those from the baselines. Specifically, we estimate the expected depth value $d(\mathbf{r})$ along each ray $\mathbf{r}$ by alpha composition:
\begin{equation}\label{eq:depth}
    d(\mathbf{r}) = \sum_j \prod_i^{j-1} (1-\alpha_i) \alpha_j t_j
\end{equation}
where $t_j$ is the depth of sampled points $j$. As shown in Tab.~\ref{tab:4_depth_metrics}, our planar experts are flexible and are able to approximate scene geometry well in most cases. However, since the size of the plane is finite, the planes may not cover all regions in extreme viewpoints. The scenes that are modeled by the background thus induce higher depth error.
Note that this can be resolved by adopting multiple-layer boxes. We leave this for future study.

\paragraph{Speed-memory trade-off: } \method~models the scene with a mixture of planar experts. The compact representation of the planes and the associated tiny MLPs not only induces a significantly lower memory footprint compared to voxel-based approaches, the planar parameterization also allows us to exploit ray-plane intersection to sample query points efficiently for low-cost radiance evaluation. Together with our customized CUDA kernel, we can achieve {19.16} frames per second on Replica. Comparing to the baselines (see Tab. \ref{tab:4_replica_speed}), \method~achieves the best speed-memory trade-off.

\paragraph{Training strategy: } To validate the contribution of each training technique (Sec. \ref{sec:training-tech}), we evaluate our model with different combinations. As shown in Tab.~\ref{table:ablation_training}, initializing plane geometry with sparse point clouds significantly improves the performance. We conjecture this is because good initialization allows the model to alleviate the shape-radiance ambiguity~\cite{zhang2020nerf++} and converge to the correct geometry. With the help of distillation, one can further reduce the artifact and improve the results. We hypothesize this is because the guidance of teacher model prevents our model from getting stuck at local minima. Both observations concur with the findings of previous works~\cite{reiser2021kilonerf, wei2021nerfingmvs}. We also note that one needs to conduct SfM to obtain the camera poses in practice, hence the sparse point cloud from SfM is essentially ``free''. 

\begin{table}[t]
\centering
\vspace{-1mm}
\resizebox{0.96\columnwidth}{!}{
\begin{tabular}{ c|c|c|c|c } 
 \hline
  SfM geometry & Distillation & PSNR & SSIM & LPIPS\\
 \hline 
             &            & 25.069 & 0.818 & 0.158 \\
  \checkmark &            & 30.810 & 0.909 & 0.082 \\
  \checkmark & \checkmark & 33.659 & 0.941 & 0.051 \\
 \hline 
 
\end{tabular}
}
\vspace{-2mm}
\caption{\textbf{Ablation study.} Scene: Replica kitchen. ``SfM geometry'' refers to planes initialization with point cloud extracted by COLMAP~\cite{schoenberger2016sfm, schoenberger2016mvs}.}

\label{table:ablation_training}
\end{table}
\paragraph{Performance w.r.t. number of planar experts: } Since we aim to model the scene appearance and geometry with planar experts, one natural question to ask is: how well does the approach scale with the number of planar experts.
As shown in Tab.~\ref{table:ablation_planenum}, more planes in general leads to better results. This is reasonable as we can fit the scene much better. However, it may also increase the model size and reduce the efficiency due to more ray-plane intersections.

\begin{figure}[t]
    \vspace{-1mm}
    \centering
        \includegraphics[width=0.3\linewidth]{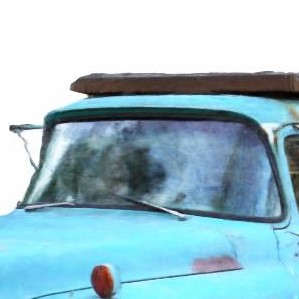}
        \includegraphics[width=0.3\linewidth]{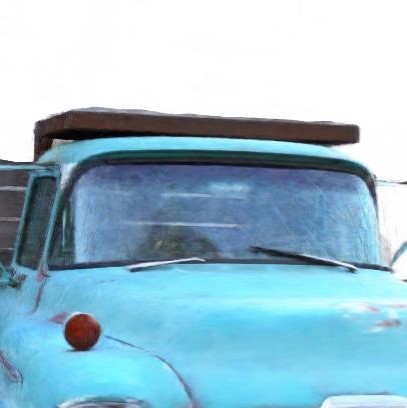}
        \includegraphics[width=0.3\linewidth]{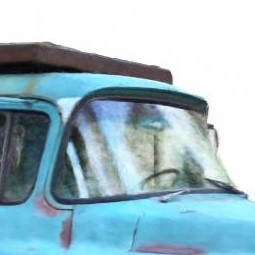}
    \caption{\textbf{Specular effects.} Dataset: Tanks\&Temples.}
    \vspace{-2mm}
    \label{fig:4_specular}
\end{figure}

\begin{table}[t]
\centering
\vspace{-2mm}
\resizebox{0.96\columnwidth}{!}{
\begin{tabular}{ c|c|c|c|c|c|c } 
 \hline
 \#~Planes & 25 & 50 & 100 & 200 & 500 & 1000\\
 \#~Params(M) & 0.15 & 0.31 & 0.62 & 1.24 & 3.11 & 6.21\\
 \hline 
 PSNR  & 26.41 & 27.69 & 29.19 & 30.10 & 30.87 & 30.64\\
 SSIM  & 0.828 & 0.851 & 0.877 & 0.888 & 0.900 & 0.902\\
 LPIPS & 0.168 & 0.140 & 0.112 & 0.098 & 0.088 & 0.085\\
 \hline 
 
\end{tabular}
}
\vspace{-2mm}
\caption{\textbf{Effect of plane number.} Dataset: Replica. Note that \#planes=500 achieves the best complexity-quality trade-off.}
\vspace{-0mm}
\label{table:ablation_planenum}
\end{table}
\paragraph{Specular effect: }
{Similar to NeRF, our planar experts model view-dependent effects by taking viewing direction $d$ into account (see Eq.~\ref{eq:radiance}). We further alpha-composite radiances from all intersecting planes ($\sim$10 per ray) to compensate for the specular effect (similar to MPI). An example of specular windows is shown in Fig.~\ref{fig:4_specular}.}

\begin{table}[t]
    \centering
    \resizebox{\linewidth}{!}
    {
    \setlength\tabcolsep{1.5pt} 
    \begin{tabular}{ccc}
    
    \includegraphics[width=0.3\linewidth]{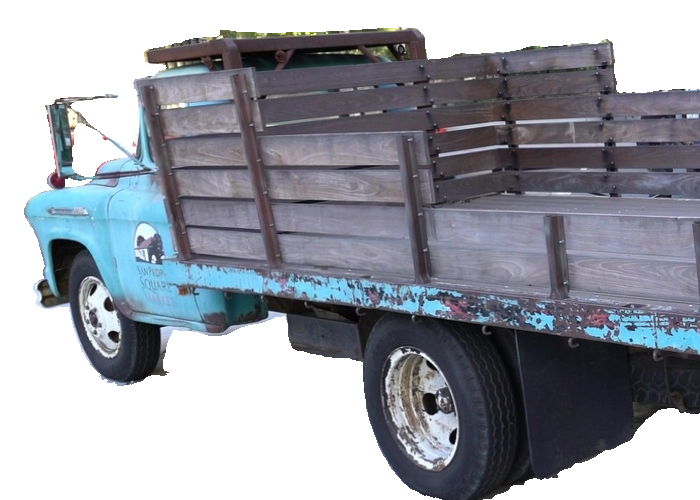} &
    \includegraphics[width=0.3\linewidth]{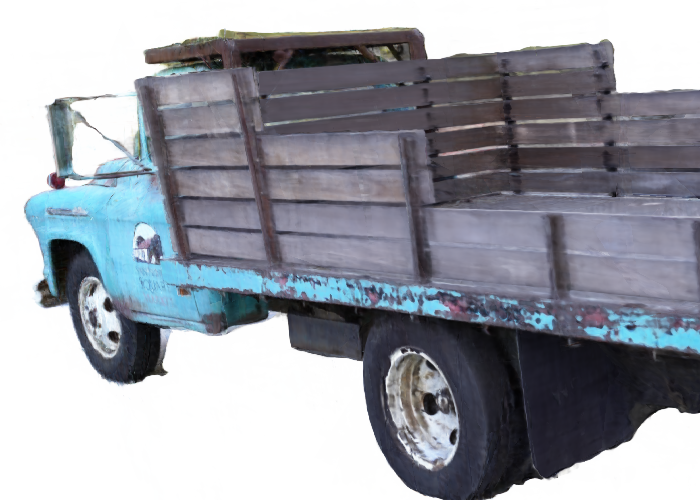} &
    \includegraphics[width=0.3\linewidth]{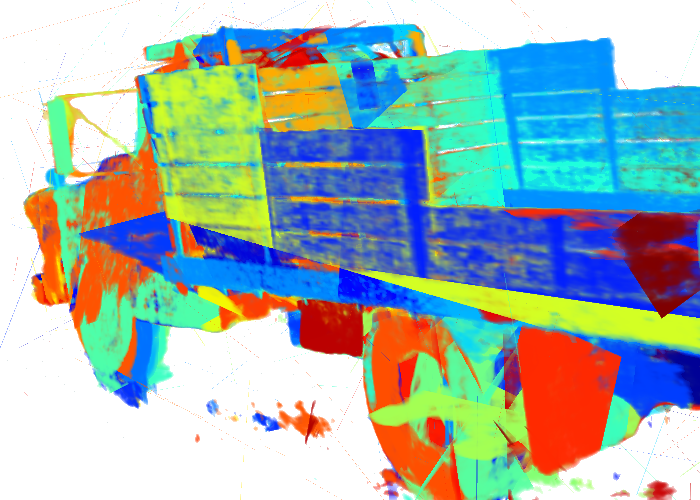} \\
    
    {Ground-Truth} &
    {Ours Rendering} &
    {Geometry} \\
    
    \end{tabular}
    }
    \vspace{-3mm}
   \captionof{figure}{\textbf{Visualizing the geometry of learned planes.} Each color is a plane learned by \method.
   }
   \vspace{0mm}
    \label{fig:4_geometry_visualization}
\end{table}
\begin{table}[t]
\centering
 \resizebox{0.96\linewidth}{!}{
{
\setlength\tabcolsep{1.5pt} 
\begin{tabular}{ ccccc } 
\includegraphics[width=0.19\linewidth]{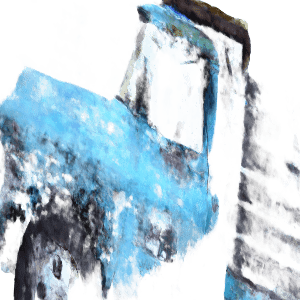}&
\includegraphics[width=0.19\linewidth]{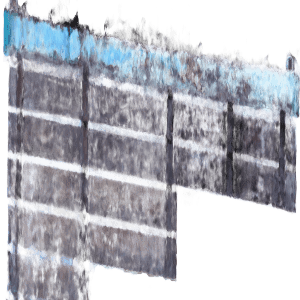}& 
\includegraphics[width=0.19\linewidth]{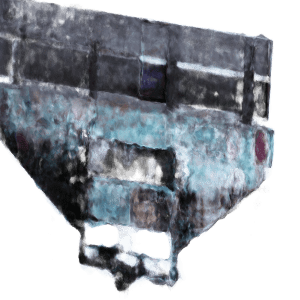}& 
\includegraphics[width=0.19\linewidth]{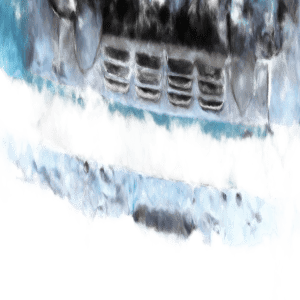}&
\includegraphics[width=0.19\linewidth]{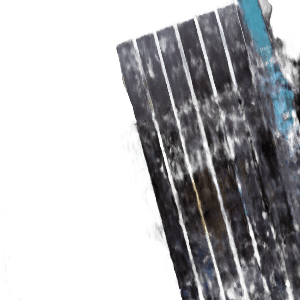}
\\
\includegraphics[width=0.19\linewidth]{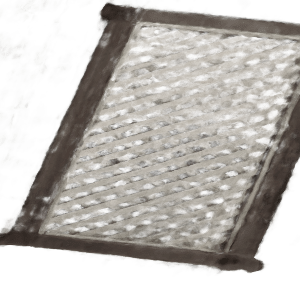} &
\includegraphics[width=0.19\linewidth]{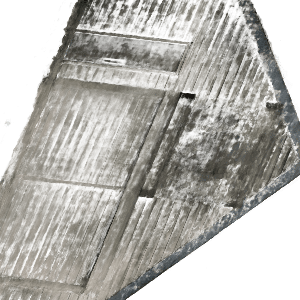}& 
\includegraphics[width=0.19\linewidth]{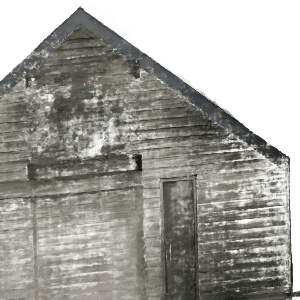}& 
\includegraphics[width=0.19\linewidth]{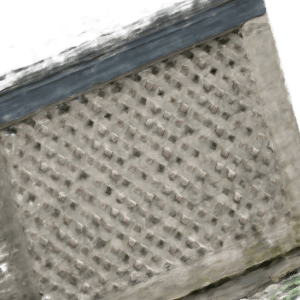}&
\includegraphics[width=0.19\linewidth]{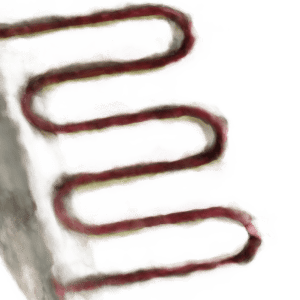}
\\
\includegraphics[width=0.19\linewidth]{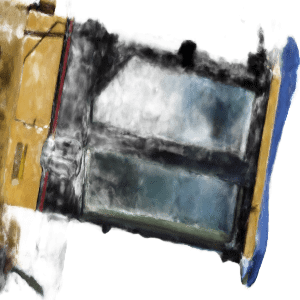} &
\includegraphics[width=0.19\linewidth]{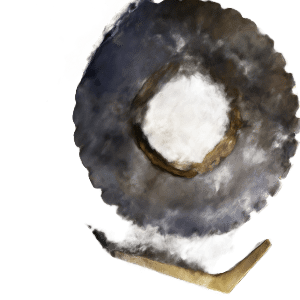}& 
\includegraphics[width=0.19\linewidth]{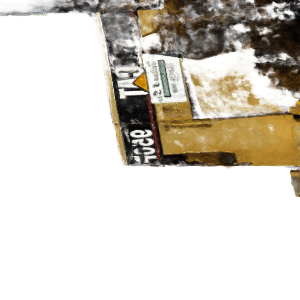}& 
\includegraphics[width=0.19\linewidth]{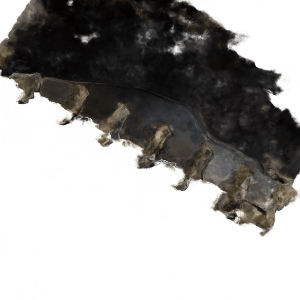}&
\includegraphics[width=0.19\linewidth]{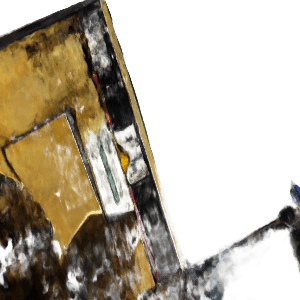}
\\
\end{tabular}
}
}
\vspace{-3mm}
\captionof{figure}{\textbf{Selected planar experts on Tanks \& Temple}. We prebake the alpha and color values into a 2D texture for each planar expert, capturing diverse local surfaces with diverse appearance and geometry, e.g., the bike rack and the wheel.}
\label{fig:4_vis_texture}
\vspace{-1mm}
\end{table}
\paragraph{Planar experts visualization: }
To better understand what is learned in the \method, we visualize its planar spirits geometry, plane index, and textures. Specifically, we show the rendered surface colored by alpha-composed planar surface indices in Fig.~\ref{fig:4_geometry_visualization}. \method~learns to capture these structures with few large planes (denoted in the same color), while approximating the non-planar regions (\eg tires, front of car) with more planes. Fig.~\ref{fig:4_vis_texture} depicts a collection of learned textures maps baked from the radiance field of several planar experts. We see that comprehensive per-plane textures have been learned with sufficient interpretability. 

\paragraph{Combining with graphics engine: }
One appealing property of our approach is its compatibility with polygon-mesh-based rendering engines. In fact, our representation could be considered as a polygon mesh with $K$ rectangular faces. We can thus pre-bake ray colors into high-resolution, view-dependent textures for each plane and save the textured mesh representation of the scene. We can then write our view-dependent shader in OpenGL and render the scene using the standard rasterization engine. Note that depth sorting for each planar surface and back-to-front rendering is necessary to ensure the correct alpha blending procedure in Eq.~\ref{eq:alpha}.  Texture baking significantly accelerates the rendering speed at the cost of additional memory consumption and small rendering quality drop, due to the discretization of the continuous radiance field. The final accelerated rendering achieves 976 frames per second for the 1000-plane Truck scene at 1920x1080 resolution on a single RTX 3090 desktop.

\paragraph{Limitations: }
Our method has several key limitations. First of all, \method~relies heavily on SfM point clouds for plane initialization (see Tab. \ref{table:ablation_training}). If the sparse point clouds are noisy or unavailable, our performance will degrade. Furthermore, our model currently cannot handle unbounded scenes. One possible solution is to incorporate techniques from NeRF++ \cite{zhang2020nerf++} to model the background texture with non-euclidean coordinates. We leave this for future study.

\begin{table}[t]
\centering
\small
\scalebox{0.85}{
\begin{tabular}{l|c|c|c|c|c} 
 \hline
 & NeX & NeRF & PlenOctree* & KiloNeRF* & Ours \\
 \hline
 \#~Params (M) & 21.28 & 1.19 & 1457.2 & 6.21 & 3.11 \\
 FPS & 0.142 & 0.106 & 78.04 & 4.19 & 19.16\\
 \hline
\end{tabular}
}
\vspace{-3mm}
\caption{{\textbf{Model size and inference speed on Replica.}}}
\vspace{-3mm}
\label{tab:4_replica_speed}
\end{table}

\section{Conclusion}
In this paper, we proposed NeurMiPs, a novel 3D representation for novel view synthesis. NeurMiPs represents the 3D scene with a mixture of learnable planar experts. Each plane consists of a rectangular shape and a neural radiance field. Our approach alleviates the frontal-parallel limitation of MPI-based methods while remaining efficient thanks to ray-casting-based rendering. We demonstrated that our approach could be integrated with classic rendering pipelines. We believe NeurMiPs will open new possibilities for 3D modeling and rendering. 
\paragraph{Acknowledgement} We thank National Center for High-performance Computing (NCHC) for providing computational and storage resources. We also thank NVIDIA for hardware donation.


\clearpage
\appendix
\section*{Supplementary Material}
\section{Overview}
In the supplementary material, we first visualize the training and testing views of dataset Replica in Section~\ref{sect:replica_camera}. In addition, we then provide the runtime breakdown and implementation details in Section~\ref{sect:runtime},~\ref{sect:baselines}. Finally, we provide more qualitative and quantitative results in Section~\ref{sect:qualitative} and ~\ref{sect:quantitative}. The supplementary video ``NeurMips-intro.mp4'' briefly introduces our method and compares qualitative results with other works. Complete comparison videos of each scene are also provided under the folder ``videos''.
\begin{table*}[h]
\centering
\setlength\tabcolsep{1.5pt} 

{
\begin{tabular}{ ccccccc } 
\includegraphics[width=0.13\textwidth]{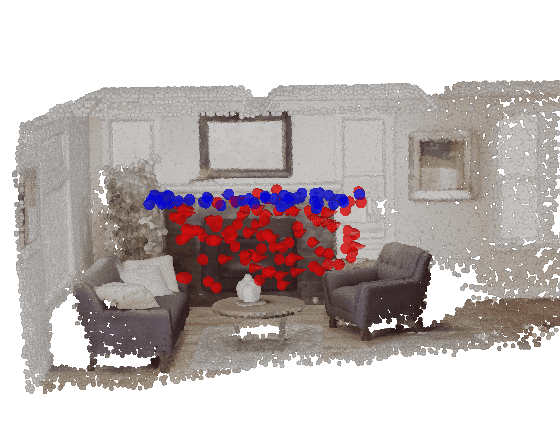} &
\includegraphics[width=0.13\textwidth]{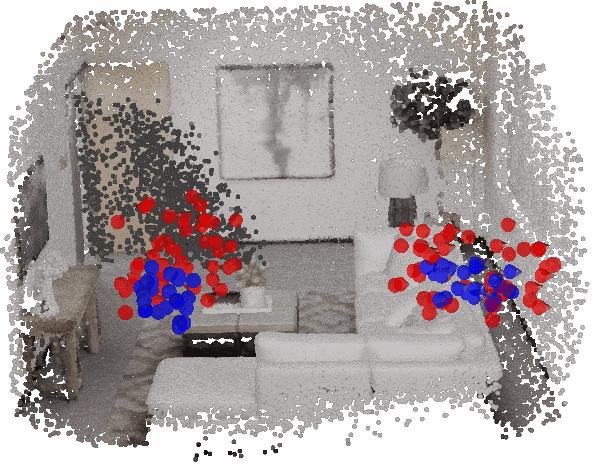} &
\includegraphics[width=0.13\textwidth]{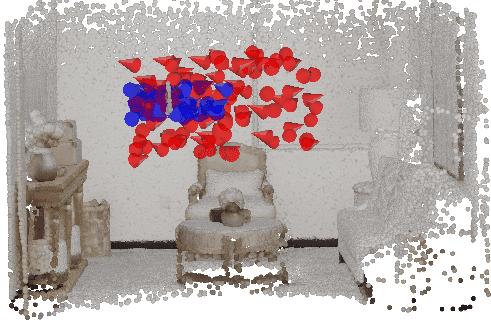} &
\includegraphics[width=0.13\textwidth]{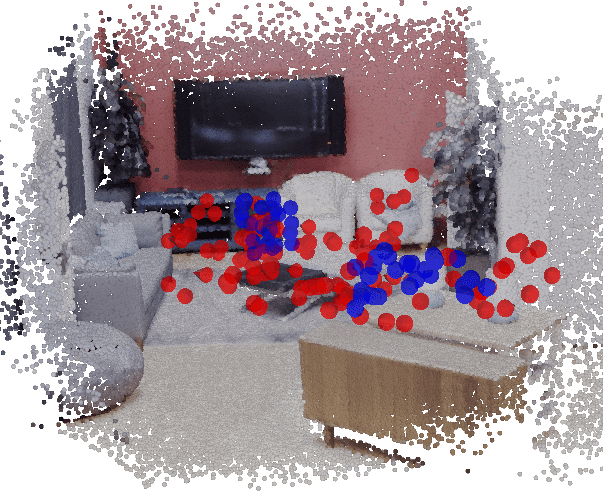} &
\includegraphics[width=0.13\textwidth]{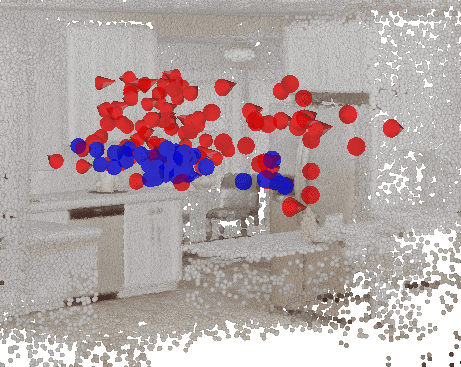} &
\includegraphics[width=0.13\textwidth]{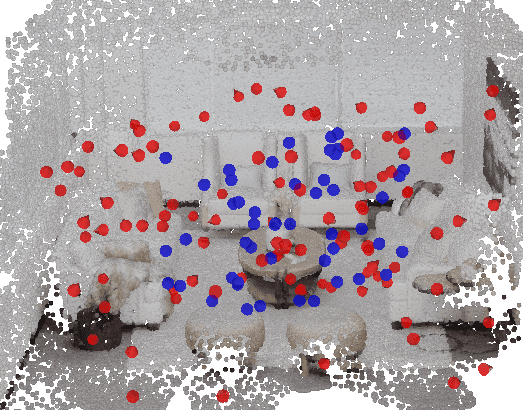}& 
\includegraphics[width=0.13\textwidth]{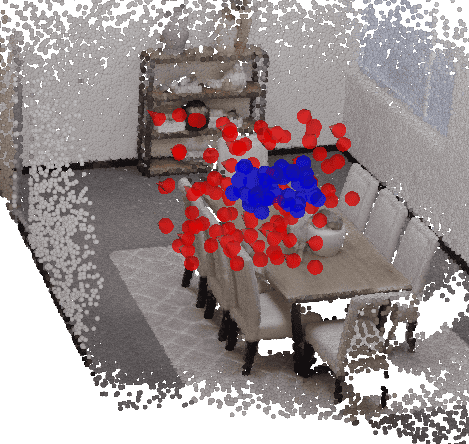}\\

\includegraphics[width=0.13\textwidth]{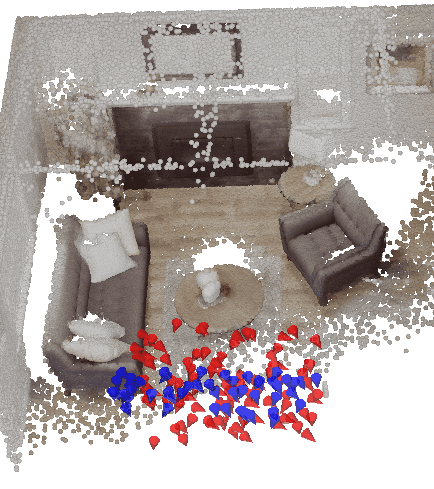} &
\includegraphics[width=0.13\textwidth]{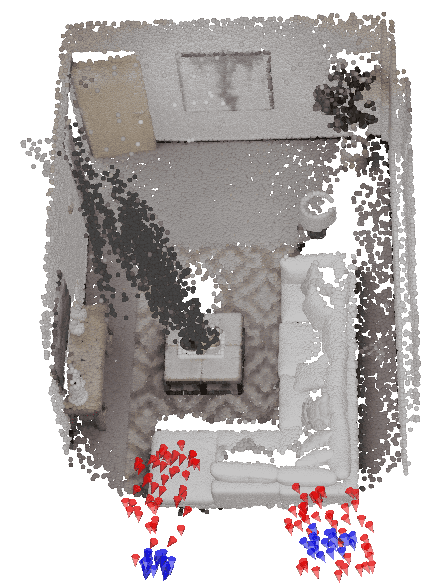} &
\includegraphics[width=0.13\textwidth]{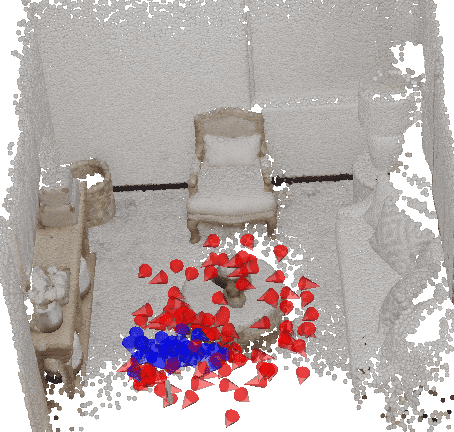} &
\includegraphics[width=0.13\textwidth]{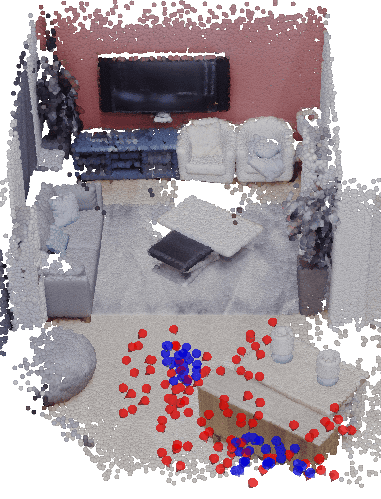} &
\includegraphics[width=0.13\textwidth]{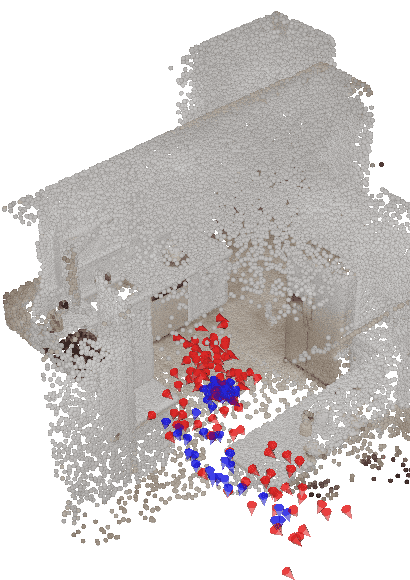} &
\includegraphics[width=0.13\textwidth]{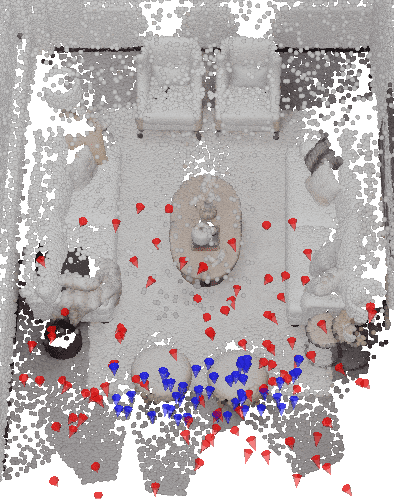}& 
\includegraphics[width=0.13\textwidth]{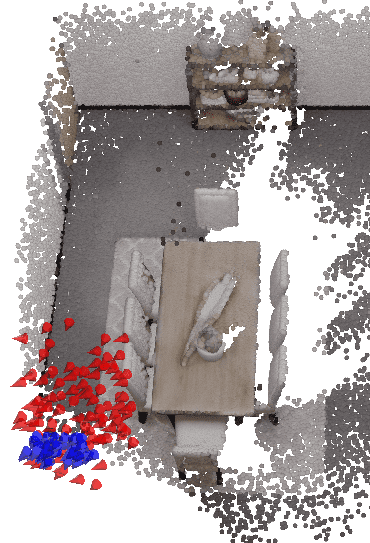}\\

\includegraphics[width=0.13\textwidth]{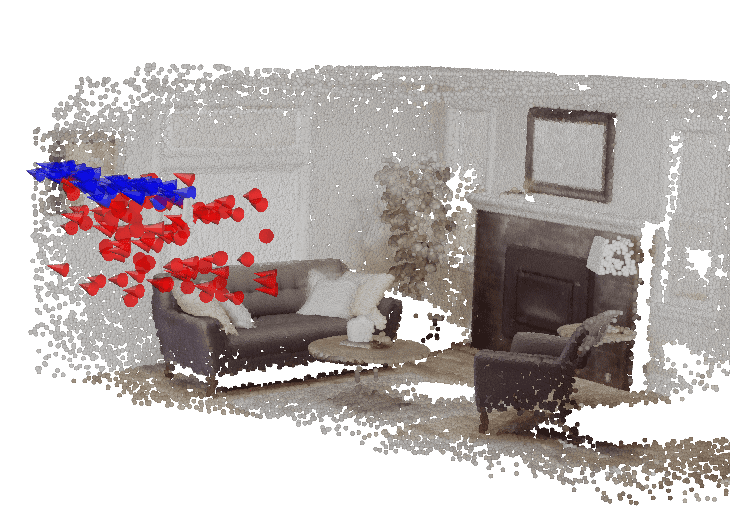} &
\includegraphics[width=0.13\textwidth]{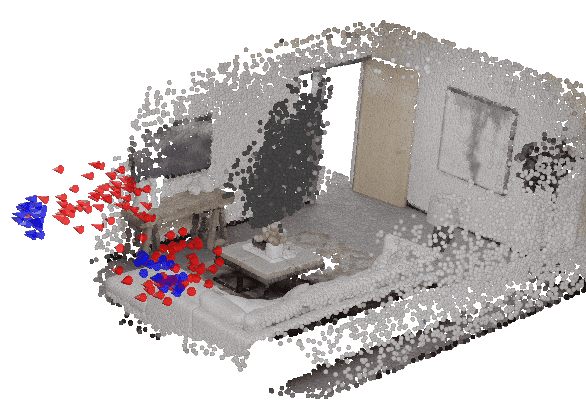} &
\includegraphics[width=0.13\textwidth]{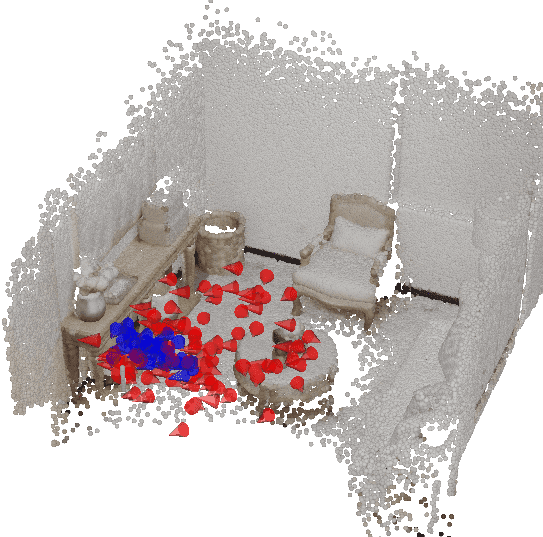} &
\includegraphics[width=0.13\textwidth]{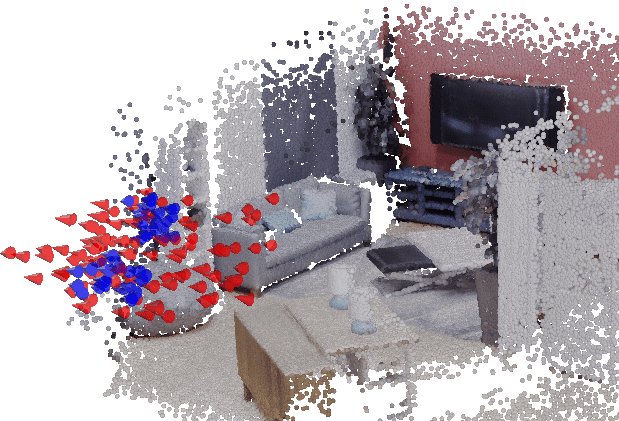} &
\includegraphics[width=0.13\textwidth]{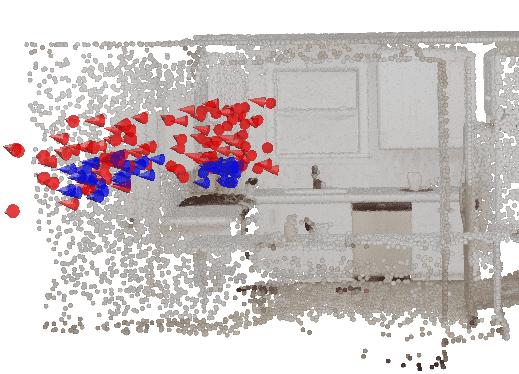}  &
\includegraphics[width=0.13\textwidth]{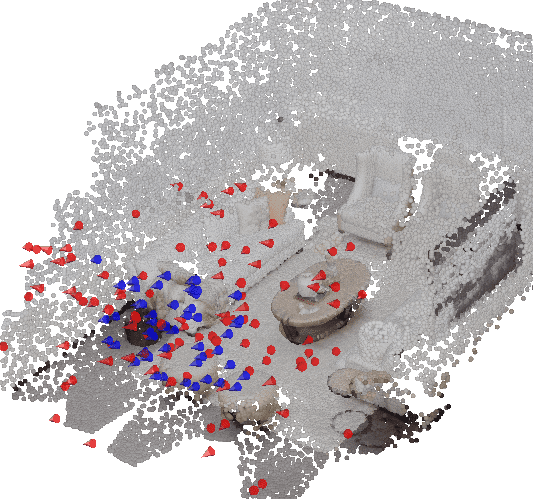}& 
\includegraphics[width=0.13\textwidth]{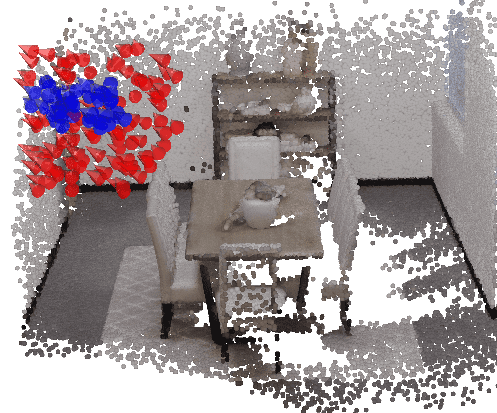}\\

Apartment 0 & Apartment 1 & Apartment 2 & FRL 0 & Kitchen & Room 0 & Room 2\\

\end{tabular}
}
\captionof{figure}{\textbf{Camera visualization of Replica.} Blue: Training views, Red: Testing views.}
\label{tab:replica_cameras}
\end{table*}
\section{Camera visualization of Replica}\label{sect:replica_camera}

We visualize the 3D scene, as well as the training and testing camera poses in Fig.~\ref{tab:replica_cameras}. Training camera poses are shown in blue, while testing poses are red. The RGBD images are unprojected to world coordinates with camera poses, representing the appearance and geometry of the 3D scene. Note that the testing views cover a much broader range than training views across all scenes, making this benchmark more challenging in terms of viewpoint variations. Specifically, we can utilize this dataset to evaluate different novel view synthesis models' performance regarding view interpolation, extrapolation, zoom in\&out.
\section{CUDA optimization details}\label{sect:runtime}

\begin{table}[h]
\centering
\resizebox{\linewidth}{!}
{
\begin{tabular}{ |c|c|c|c|c| } 

\hline
Intersection & Pre-processing & Network inference & Integration & Total time \\
\hline
0.0079 & 0.0252 & 0.0177 & 0.0030 & 0.0538 \\
\hline 

\end{tabular}
}
\caption{\textbf{Runtime breakdown (in seconds).} The runtime estimation is evaluated on Replica.}
\label{tab:runtime}
\end{table}

Table~\ref{tab:runtime} reports the runtime breakdown of our complete rendering process averaging across all testing images in the Replica dataset. Specifically, we divide the entire rendering procedure into four components: (1) ``Intersection": ray-plane intersection for all viewing rays; (2) ``Pre-processing": operations including depth sorting, sampling from pre-baked alpha planes, and filtering sampled alpha values; (3) ``Network inference": parallel inference of multiple planar experts; (4) ``Integration": alpha-composition for all rays and background color replacement for rays with low alpha weight; All components are leveraged by customized CUDA kernels to significantly accelerate the rendering pipeline.

\subsection{Network inference}

In a vanilla MLP, for each linear layer, a matrix-matrix multiplication is processed by multiplying a $B \times I$ input matrix with a $I \times O$ weight matrix, where $B$ refers to batch size, $I$ is the size of input features and $O$ is the size of output features of the associated layer. In our method, the original NeRF model distills into multiple planar experts, with each expert representing a planar surface with a tiny MLP. Therefore, it is reasonable that batch size $B_i$ for each network $i$ is not identical. Directly calling (torch.matmul) in a loop for each network in PyTorch might be a straightforward solution. However, it executes CUDA kernels for each matrix multiplication sequentially, does not fully utilize GPU's SMs, and therefore achieves bad performance. Although batched matrix multiplication (torch.bmm) seems like a better choice since it allows matrix multiplications for all networks in parallel with only a single CUDA kernel launch. However, this routine supported by PyTorch requires all batch sizes $B_i$ to be the same. Hence, we develop a routine that fused the entire network evaluation (including Fourier features calculation, linear layers, and activation functions) into a single CUDA kernel. In our CUDA kernel, each CUDA block is responsible for a particular network and each thread is responsible for handling a single network input. Due to the small size of the individual networks (\~25KB), parameters for a network can be fully loaded into on-chip RAM of an SM once per frame. All intermediate results are stored in per-thread registers to avoid latency of memory transfer.

\subsection{Alpha prebaking, sampling \& filtering}
Network evaluation of all possible ray-plane intersected points is inefficient, therefore, we aim to filter out those points with low alpha values. First, each plane is baked in PyTorch by evaluating alpha values of uniformly grid-sampled points ($200 \times 200$) on the plane beforehand. Then, for each ray-plane intersected point, we get an estimated alpha value by bilinear interpolation from neighboring alpha values on the plane. Finally, points with estimated alpha weight $\displaystyle\prod_i^{j-1} (1-\alpha_i) \alpha_j$ lower than a fixed threshold $=0.001$ are discarded, which do not contribute much to alpha composition results. With this strategy, nearly 60\% of points are filtered out on the Replica dataset, 70\% on the Tanks \& Temples dataset.

\section{Baseline implementation details}\label{sect:baselines}
In this section, we provide more implementation details of each baseline methods for dataset Replica.

\paragraph{NeRF~\cite{mildenhall2020nerf}}
We used the \href{https://github.com/facebookresearch/pytorch3d/tree/main/projects/nerf}{PyTorch3D implementation} of the original paper for training and evaluation. NeRF performs hierarchical sampling on each ray. It samples 64 points uniformly in the coarse stage and then samples another 64 points according to density distribution in the fine stage. It renders each pixel by evaluating RGB-$\alpha$ values of total 128 points.
\paragraph{NeX~\cite{Wizadwongsa2021NeX}} We used the \href{https://github.com/nex-mpi/nex-code/}{official implementation} in our experiments. 
To construct the multiplane image, the model selects one of the training views as reference, which is the closest to the 3D center of all training camera positions. 192 planes are parallel and placed in front of this reference view, rendering novel views with homography warping.
\paragraph{PlenOctree~\cite{yu2021plenoctrees}} 
We used the \href{https://github.com/sxyu/plenoctree}{official implementation} in our experiments. After training NeRF with spherical harmonics (i.e. NeRF-SH), it is then converted into octree structure for real-time rendering. However, with default settings, dense grids cannot be retained even on a 32G GPU, and the filtering strategy leads to several blank regions and low image quality of novel views. As a result, we extract with lower grid resolution $=256^3$ (default: $512^3$), and retain all grids for rendering. This removes blank regions and generates better image quality. Please see Table.~\ref{tab:replica_implementation} for qualitative comparison.
\paragraph{KiloNeRF~\cite{reiser2021kilonerf}}
We used the \href{https://github.com/creiser/kilonerf}{official implementation} in our experiments. It trains an ordinary NeRF model (teacher) then distills knowledge of teacher into multiple tiny MLPs. To avoid model querying in empty space, the occupancy grid is extracted from the trained teacher by a threshold $\tau$. However, blank holes appear in rendering images with threshold $\tau = 3$. Therefore, we set $\tau = 0$ to construct dense grids, which trades rendering efficiency for better image quality. Please see Table.~\ref{tab:replica_implementation} for qualitative comparison.

\begin{table*}[h]
    \centering
    \begin{tabular}{c|ccccc}
         Ground Truth & \makecell{KiloNeRF~\cite{reiser2021kilonerf} \\ (threshold $\tau = 3$)} &
         \makecell{KiloNeRF~\cite{reiser2021kilonerf} \\ (threshold $\tau = 0$)} &
         \makecell{PlenOctree~\cite{yu2021plenoctrees} \\ (grid res.$=512^3$)} & \makecell{PlenOctree~\cite{yu2021plenoctrees} \\ (grid res.$=256^3$)} & Ours\\
         
         \includegraphics[width=0.12\textwidth]{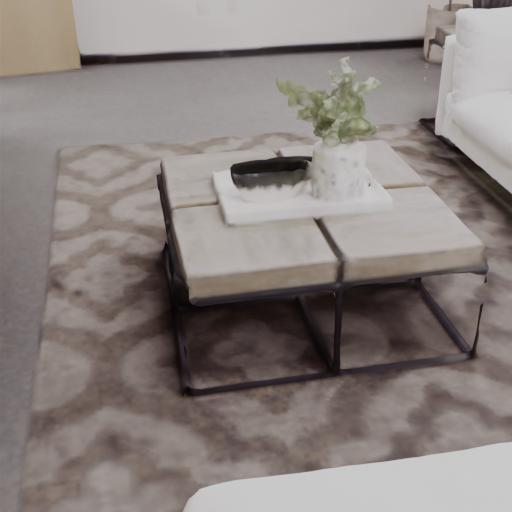} & \includegraphics[width=0.12\textwidth]{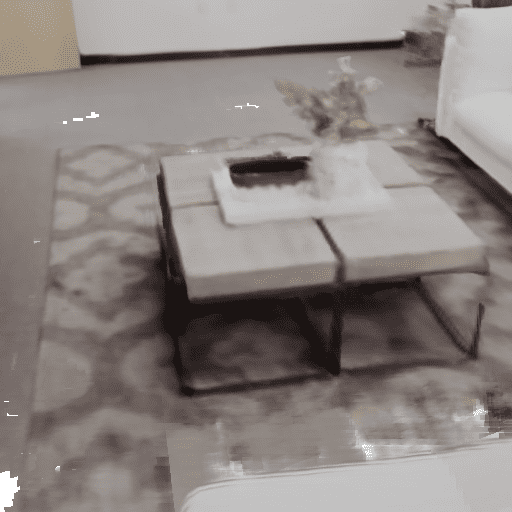} & \includegraphics[width=0.12\textwidth]{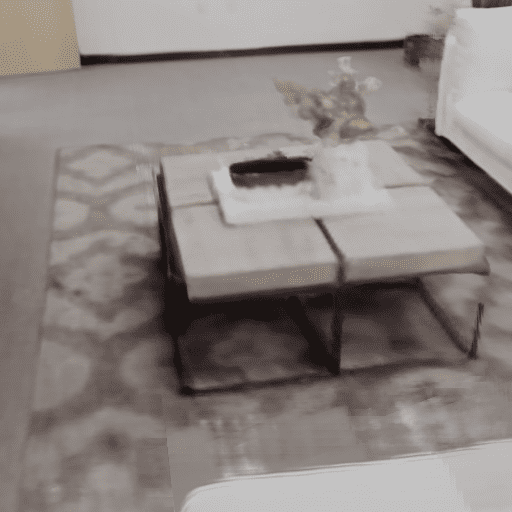} &\includegraphics[width=0.12\textwidth]{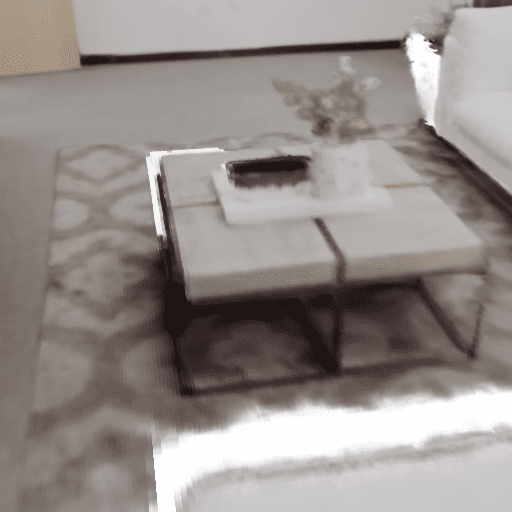} & \includegraphics[width=0.12\textwidth]{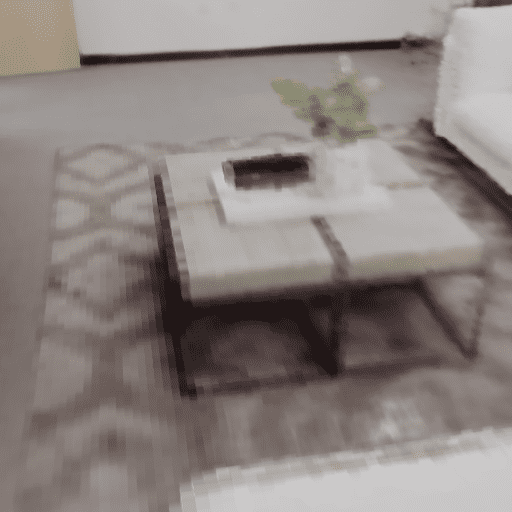} & \includegraphics[width=0.12\textwidth]{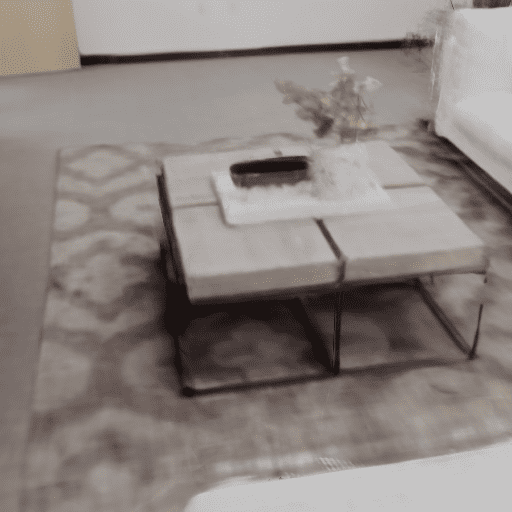} \\
         
         \includegraphics[width=0.12\textwidth]{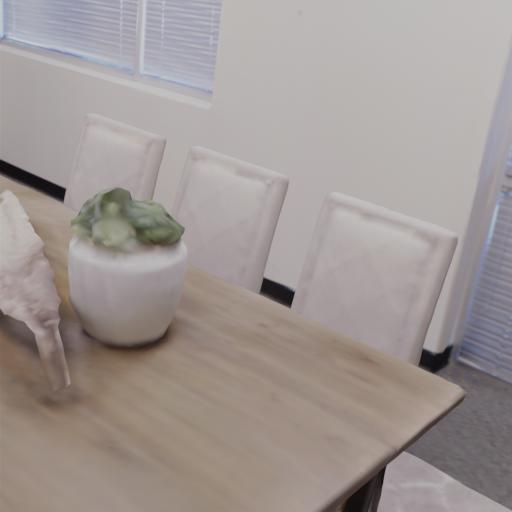} & \includegraphics[width=0.12\textwidth]{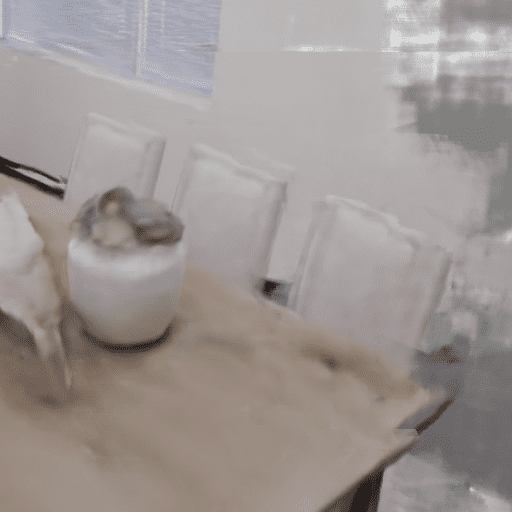} & \includegraphics[width=0.12\textwidth]{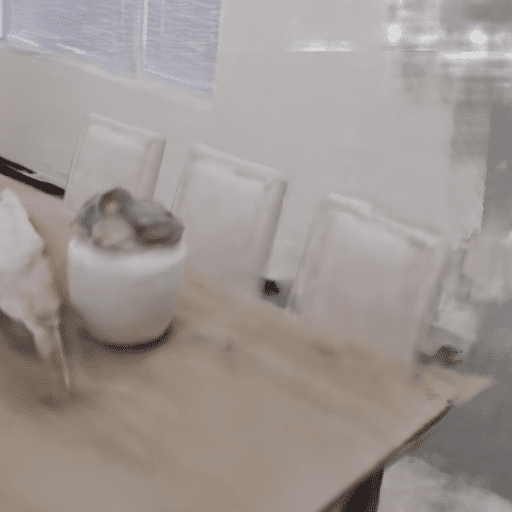} & \includegraphics[width=0.12\textwidth]{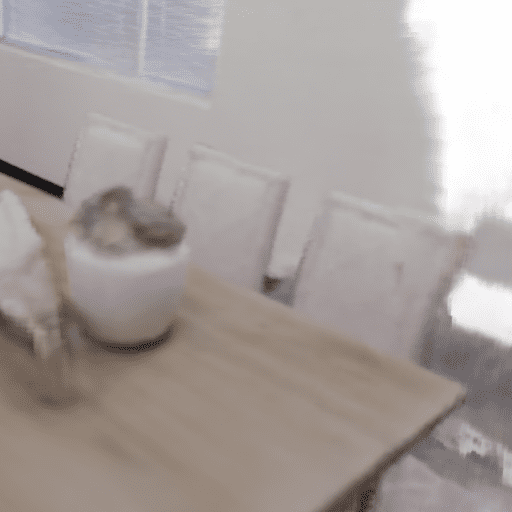} & \includegraphics[width=0.12\textwidth]{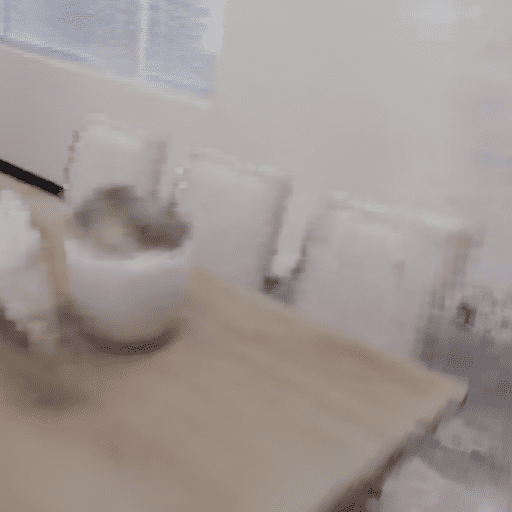} & \includegraphics[width=0.12\textwidth]{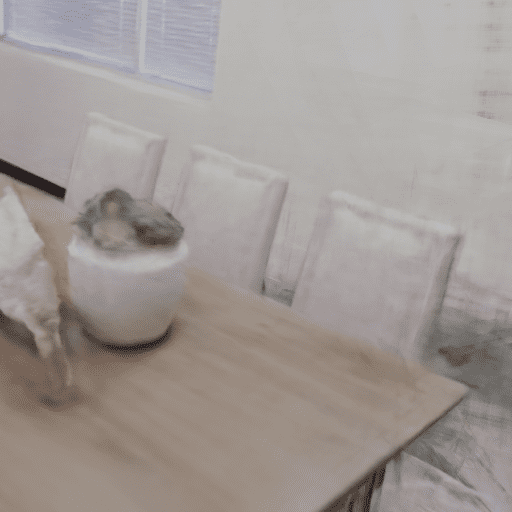} \\
         
         \includegraphics[width=0.12\textwidth]{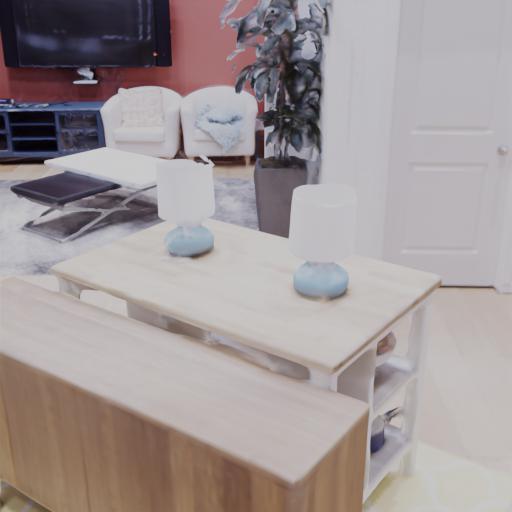} & \includegraphics[width=0.12\textwidth]{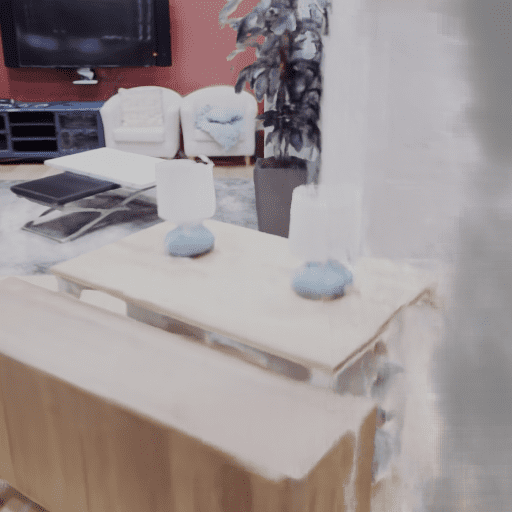} & \includegraphics[width=0.12\textwidth]{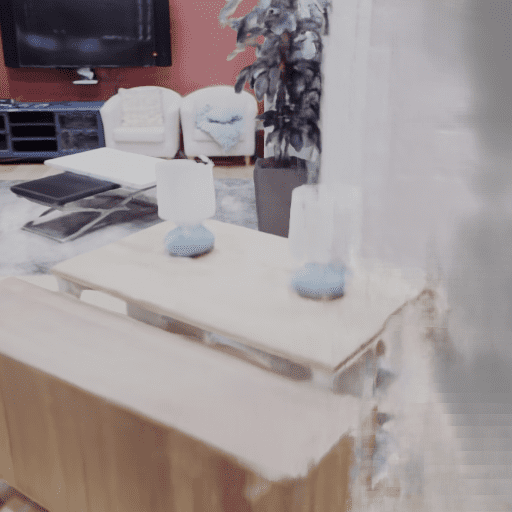} & \includegraphics[width=0.12\textwidth]{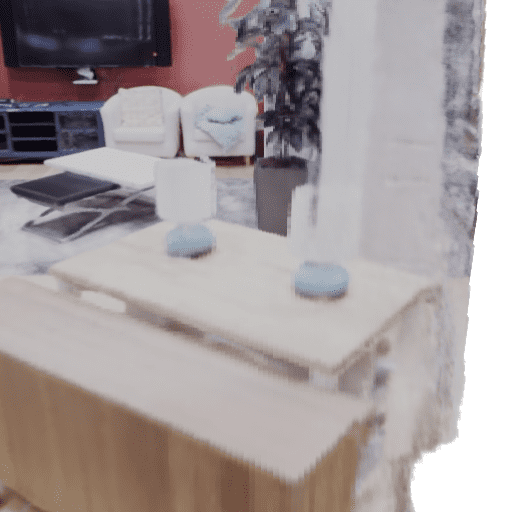} & \includegraphics[width=0.12\textwidth]{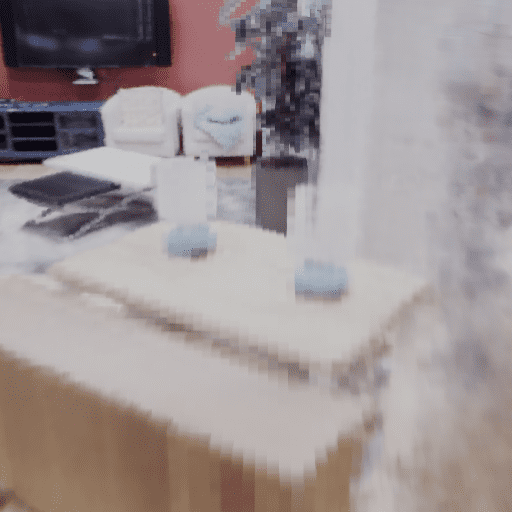} & \includegraphics[width=0.12\textwidth]{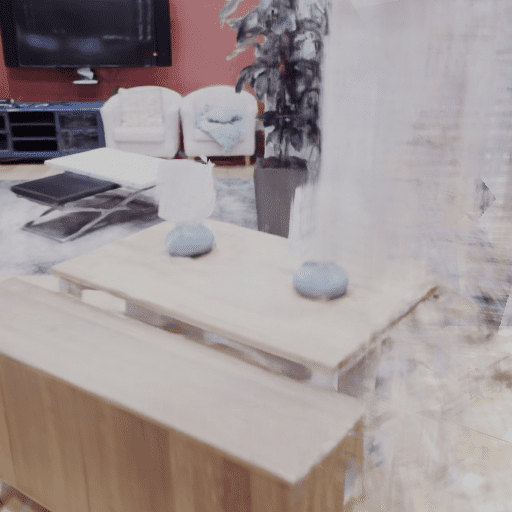}\\
    \end{tabular}
    \caption{\textbf{Qualitative results of different implementation.} Dataset: Replica}
    \label{tab:replica_implementation}
\end{table*}
\section{More qualitative comparison}\label{sect:qualitative}

We show additional qualitative results in Fig.~\ref{fig:tanks_quality_1},
Fig.~\ref{fig:tanks_quality_2},
Fig.~\ref{fig:tanks_quality_3},
Fig.~\ref{fig:replica_quality_1},
Fig.~\ref{fig:replica_quality_2}, and Fig.~\ref{fig:blendedmvs_quality}.
From Fig.~\ref{fig:tanks_quality_1}, we can see that \method~can capture planar structures well, including the grid patterns on the wall (1st row), words on the Truck (3rd row), and Caterpillar (5th row). The model can also capture thin structures in the scene, as shown in the 2nd and 4th rows of the figure. 
Moreover, Fig.~\ref{fig:tanks_quality_2} demonstrates that \method~is able to capture complex non-planar geometry, such as the surface of human statues, especially their facial expressions (1st, 4th rows). KiloNeRF~\cite{reiser2021kilonerf} produces grid-like artifacts due to its space partitioning (5th, 6th rows). In contrast, our method has a more smooth texture and fewer artifacts. 
Fig.~\ref{fig:tanks_quality_3}, Fig.~\ref{fig:replica_quality_1} and Fig.~\ref{fig:replica_quality_2} visualize the depth image of the scene, and compare the results of different methods. Our method is able to capture the boundaries between wood panels well, as seen in the last row of Fig.~\ref{fig:tanks_quality_3}. Note that dataset Tanks\&Temple does not have ground truth depth images, so in Fig.~\ref{fig:tanks_quality_3} we provide RGB images as reference. 
Fig.~\ref{fig:replica_quality_1} and Fig.~\ref{fig:replica_quality_2} shows that while NeX~\cite{Wizadwongsa2021NeX} have black artifacts when viewing from extreme views, \method~learns to generate good texture and depths in novel views. Please refer to the videos for more qualitative comparisons. 

We also evaluate and compare our method on BlendedMVS dataset~\cite{yao2020blendedmvs}, and provide qualitative results in Fig.~\ref{fig:blendedmvs_quality}. 

\begin{table*}[h]
\centering
\vspace{-5mm}
{
\setlength\tabcolsep{1.5pt} 
\begin{tabular}{cccc}

\multicolumn{1}{c}{Ground-Truth} &
\multicolumn{1}{c}{NeRF~\cite{mildenhall2020nerf}} &
\multicolumn{1}{c}{KiloNeRF~\cite{reiser2021kilonerf}} &
\multicolumn{1}{c}{Ours} \\


\includegraphics[width=0.2\textwidth]{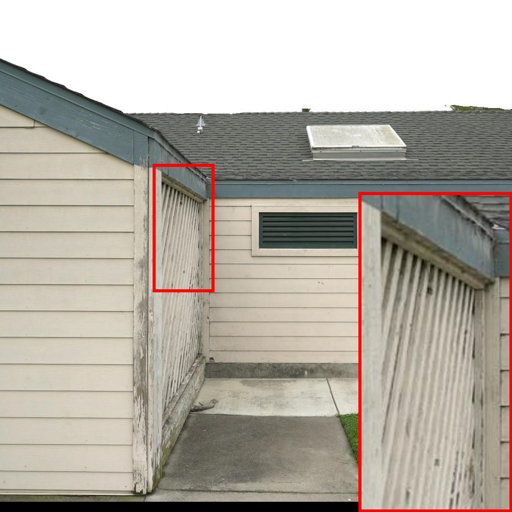} &
\includegraphics[width=0.2\textwidth]{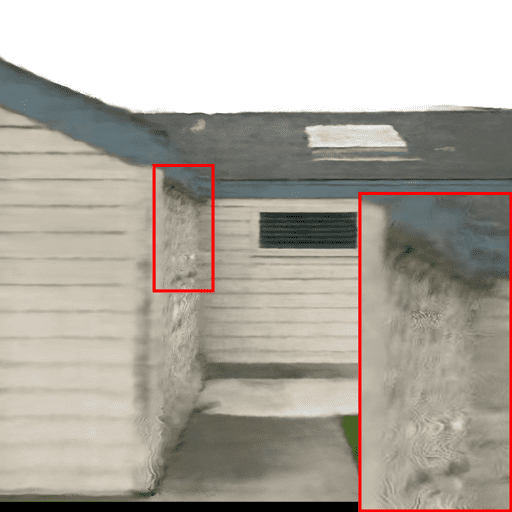} &
\includegraphics[width=0.2\textwidth]{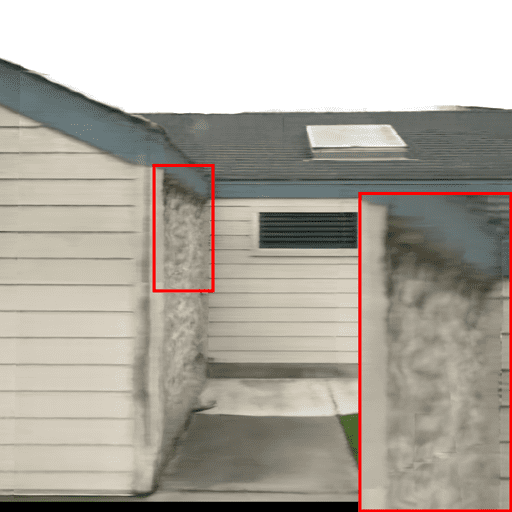} &
\includegraphics[width=0.2\textwidth]{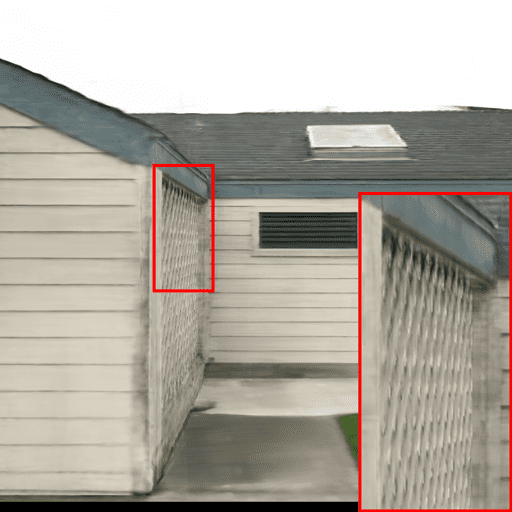} \\

\includegraphics[width=0.2\textwidth]{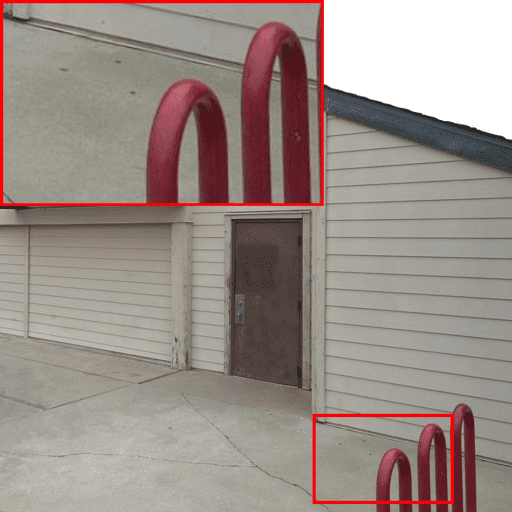} &
\includegraphics[width=0.2\textwidth]{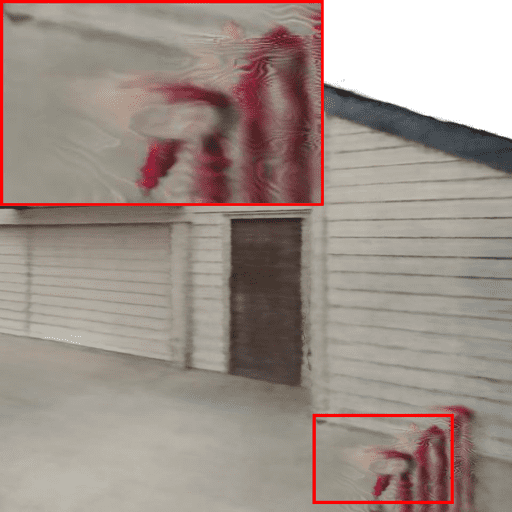} &
\includegraphics[width=0.2\textwidth]{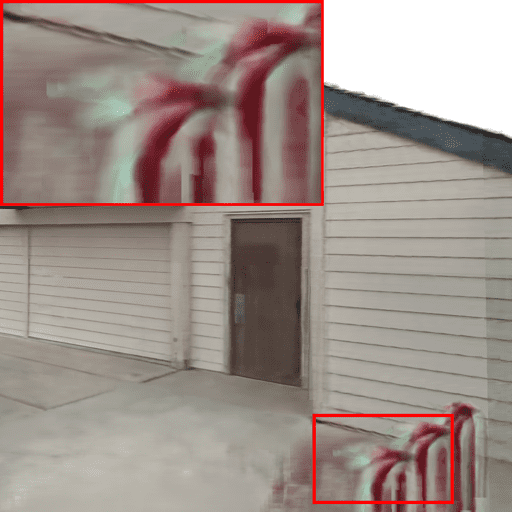} &
\includegraphics[width=0.2\textwidth]{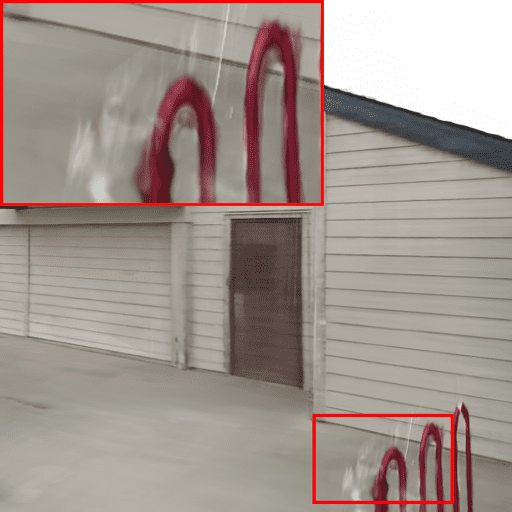} \\

\includegraphics[width=0.2\textwidth]{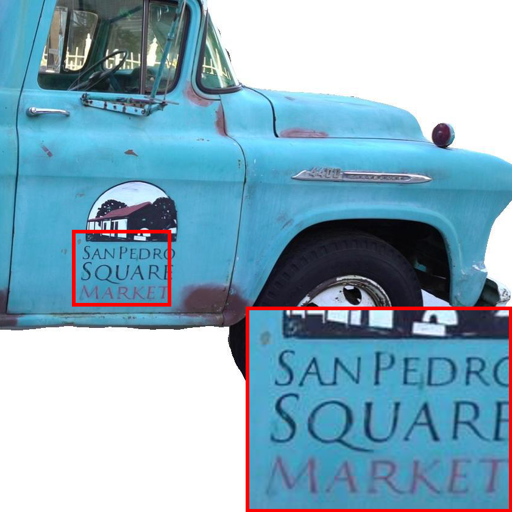} &
\includegraphics[width=0.2\textwidth]{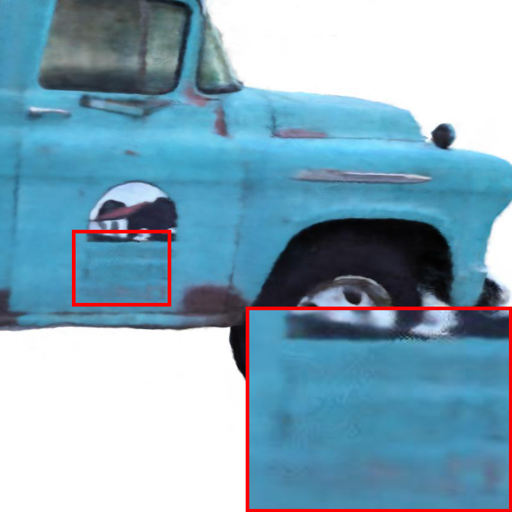} &
\includegraphics[width=0.2\textwidth]{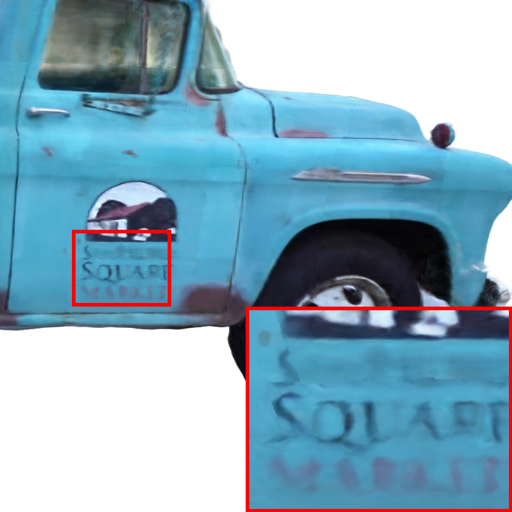} &
\includegraphics[width=0.2\textwidth]{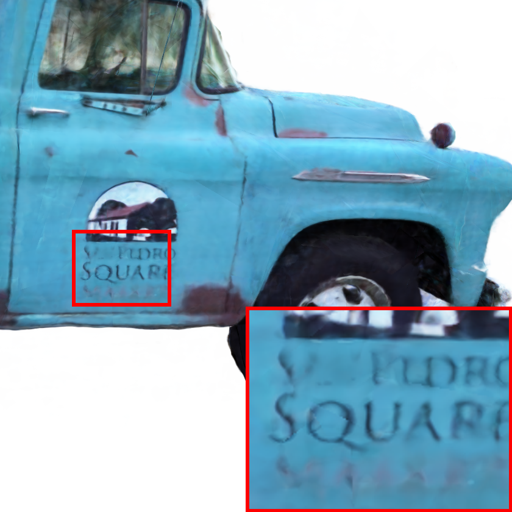} \\

\includegraphics[width=0.2\textwidth]{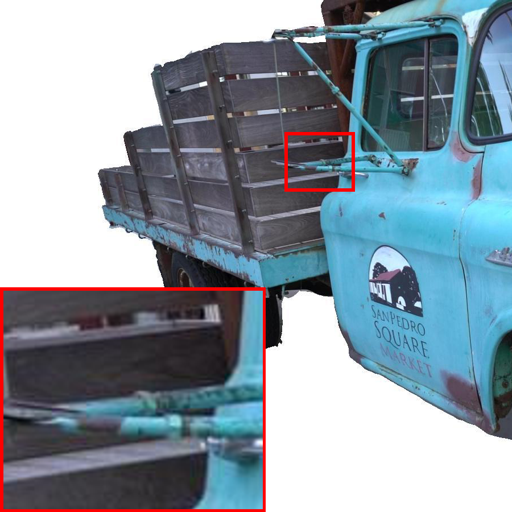} &
\includegraphics[width=0.2\textwidth]{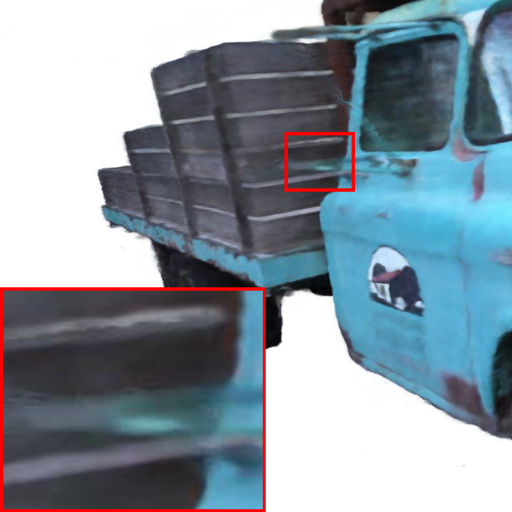} &
\includegraphics[width=0.2\textwidth]{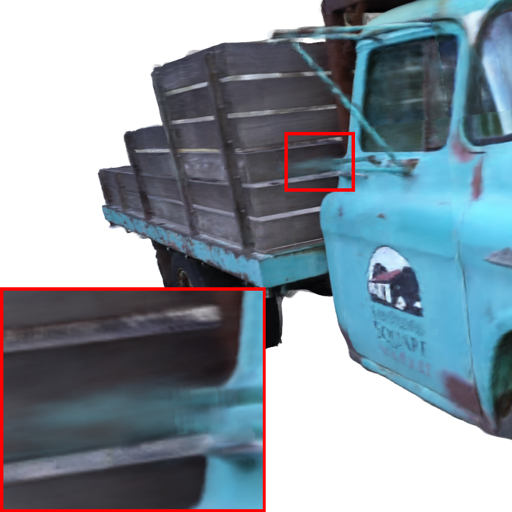} &
\includegraphics[width=0.2\textwidth]{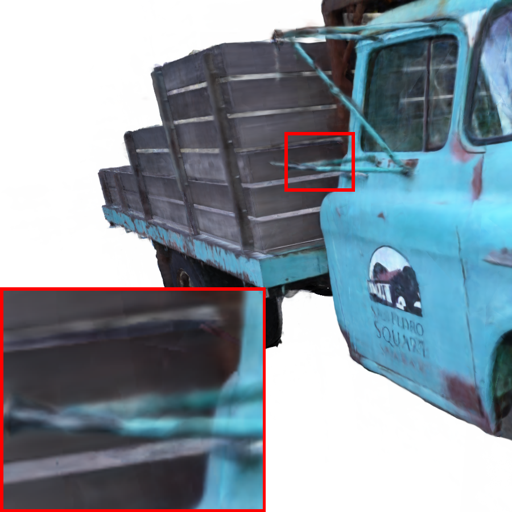} \\

\includegraphics[width=0.2\textwidth]{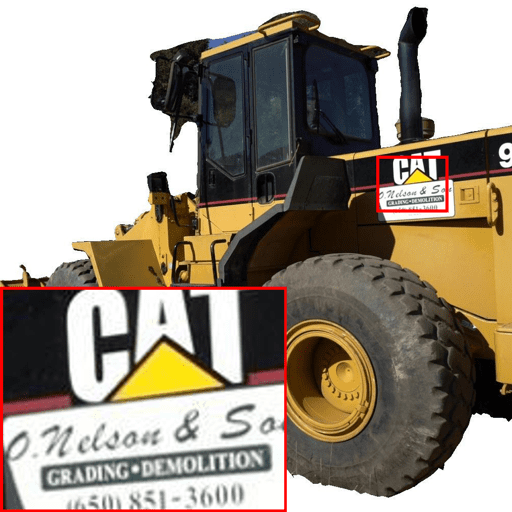} &
\includegraphics[width=0.2\textwidth]{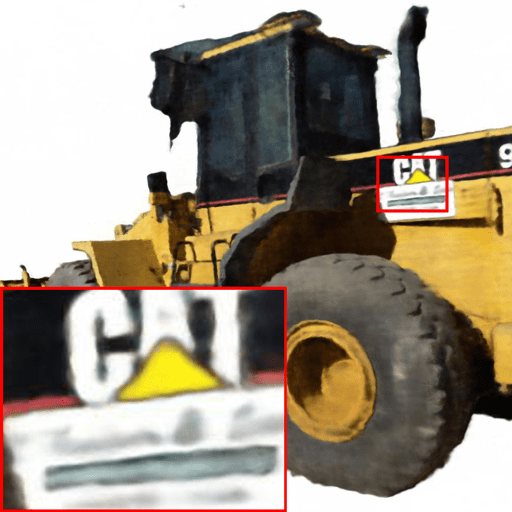} &
\includegraphics[width=0.2\textwidth]{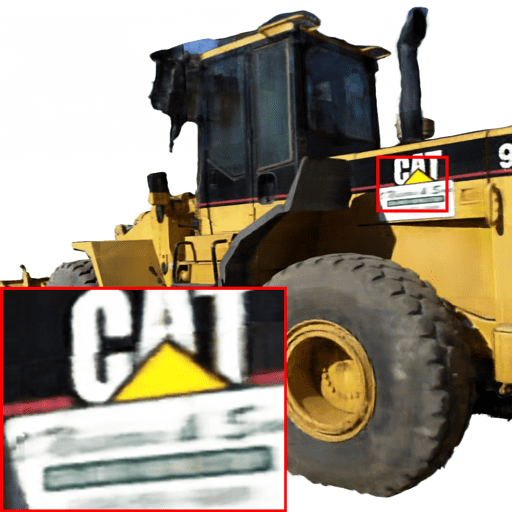} &
\includegraphics[width=0.2\textwidth]{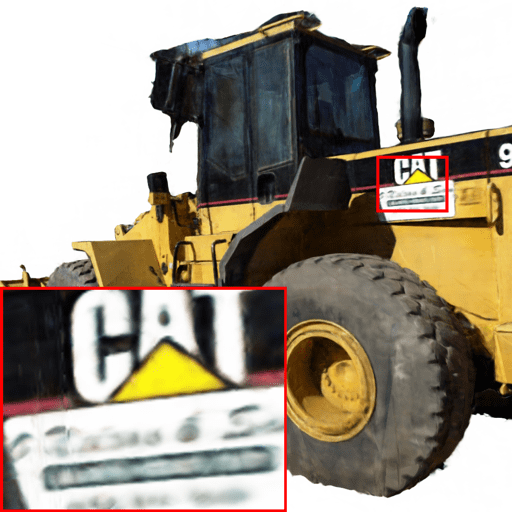} \\

\includegraphics[width=0.2\textwidth]{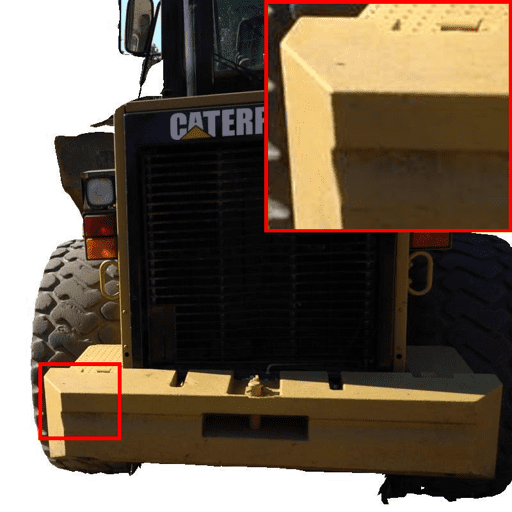} &
\includegraphics[width=0.2\textwidth]{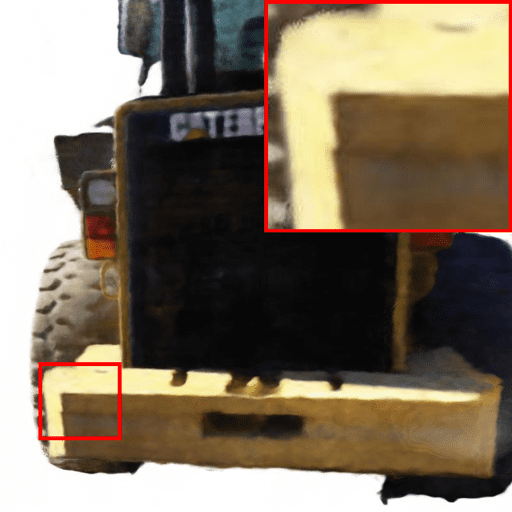} &
\includegraphics[width=0.2\textwidth]{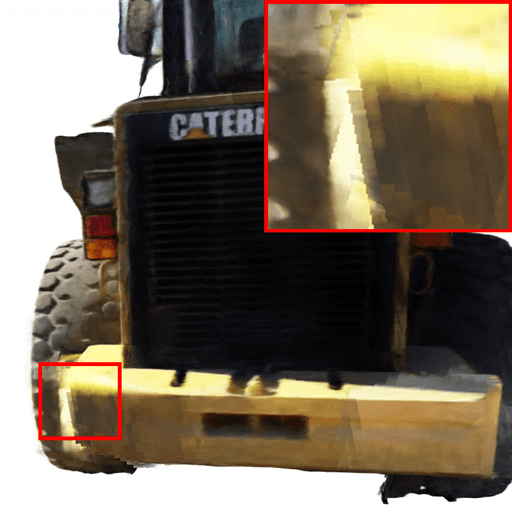} &
\includegraphics[width=0.2\textwidth]{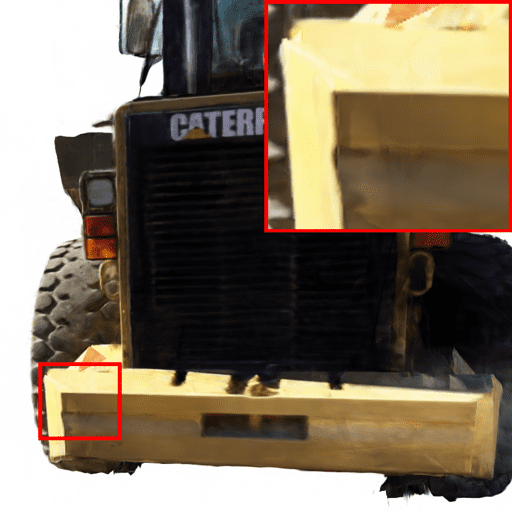} \\

\end{tabular}
}
\vspace{-3mm}
\captionof{figure}{Qualitative results for the scenes of Barn (Tanks \& Temples), Truck (Tanks \& Temples), and Caterpillar (Tanks \& Temples)}
\label{fig:tanks_quality_1}
\vspace{-5mm}
\end{table*}

\begin{table*}[h]
\centering
\vspace{-5mm}
{
\setlength\tabcolsep{1.5pt} 
\begin{tabular}{cccc}

\multicolumn{1}{c}{Ground-Truth} &
\multicolumn{1}{c}{NeRF~\cite{mildenhall2020nerf}} &
\multicolumn{1}{c}{KiloNeRF~\cite{reiser2021kilonerf}} &
\multicolumn{1}{c}{Ours} \\

\includegraphics[width=0.2\textwidth]{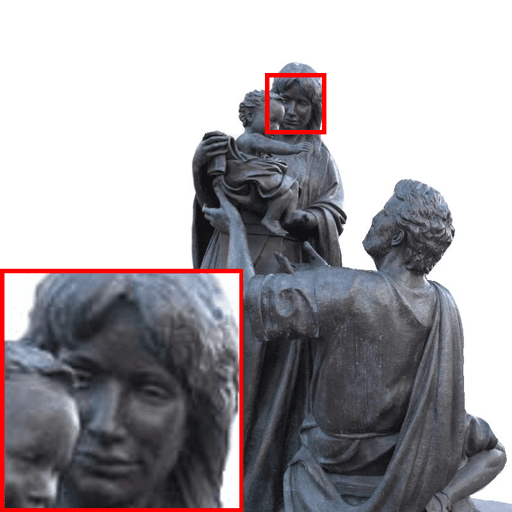} &
\includegraphics[width=0.2\textwidth]{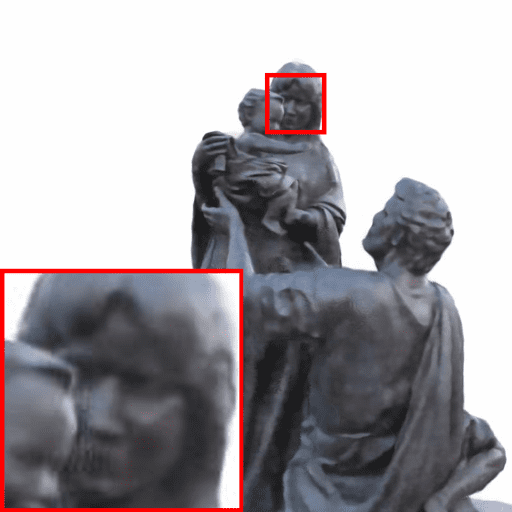} &
\includegraphics[width=0.2\textwidth]{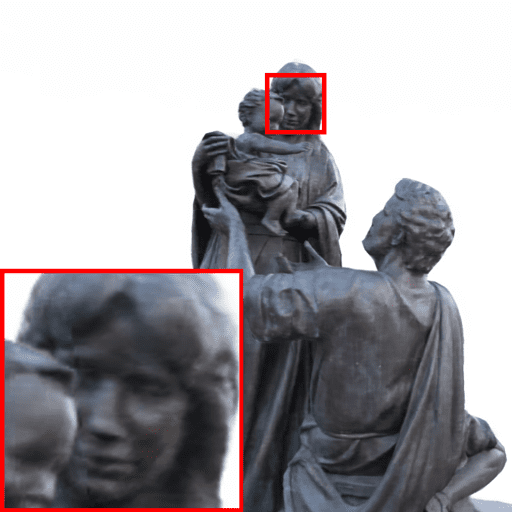} &
\includegraphics[width=0.2\textwidth]{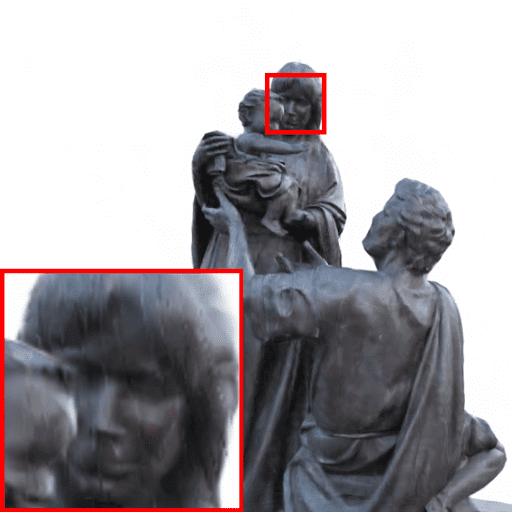} \\

\includegraphics[width=0.2\textwidth]{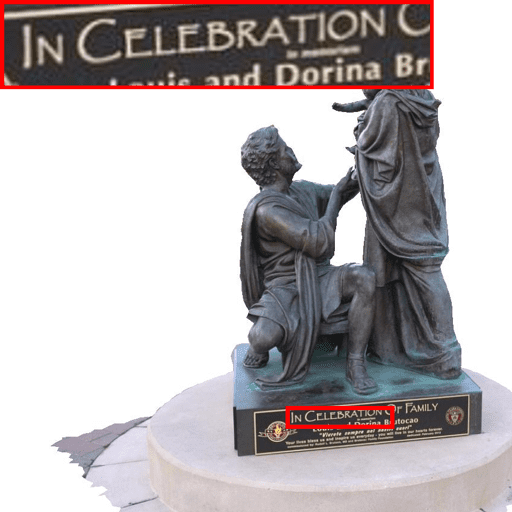} &
\includegraphics[width=0.2\textwidth]{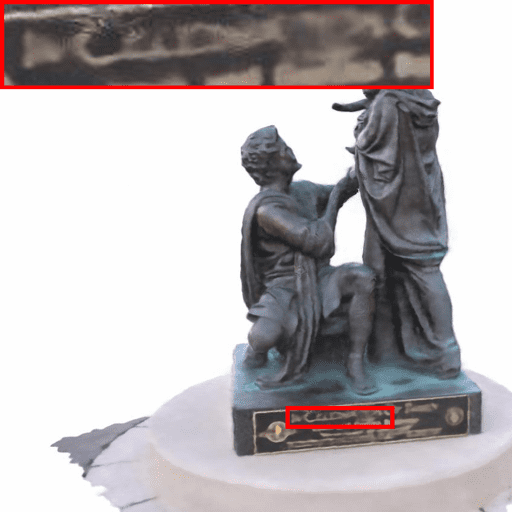} &
\includegraphics[width=0.2\textwidth]{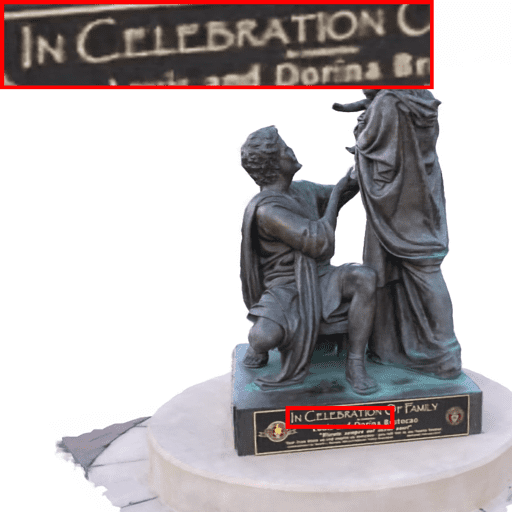} &
\includegraphics[width=0.2\textwidth]{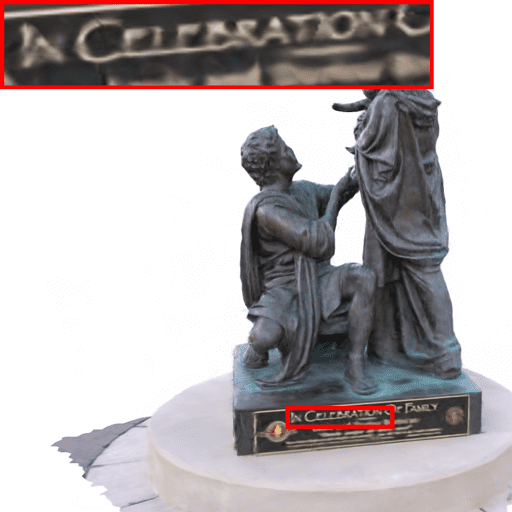} \\

\includegraphics[width=0.2\textwidth]{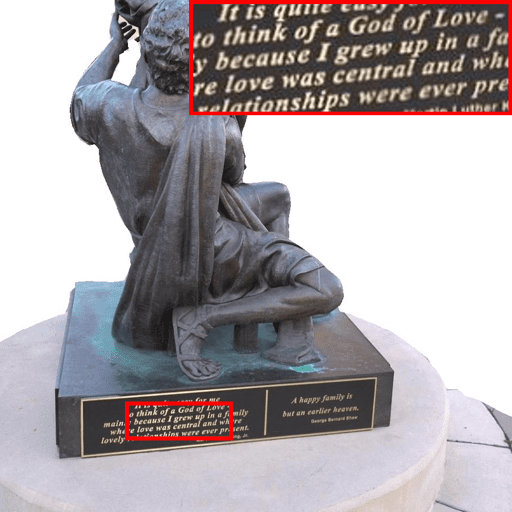} &
\includegraphics[width=0.2\textwidth]{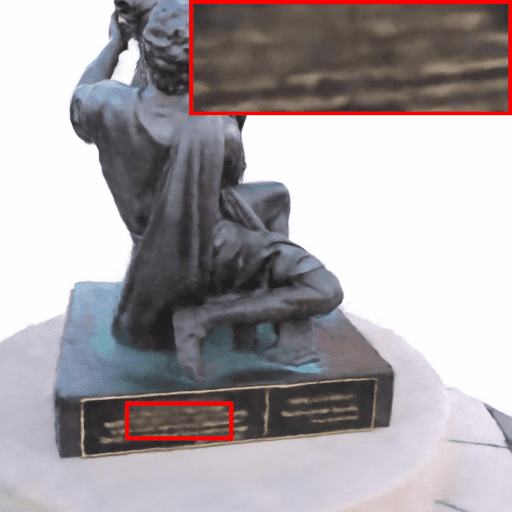} &
\includegraphics[width=0.2\textwidth]{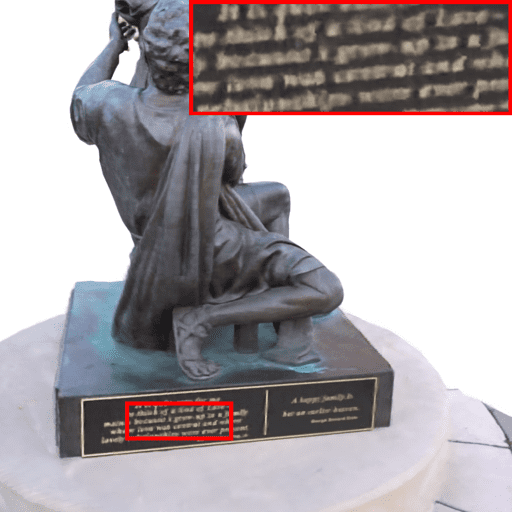} &
\includegraphics[width=0.2\textwidth]{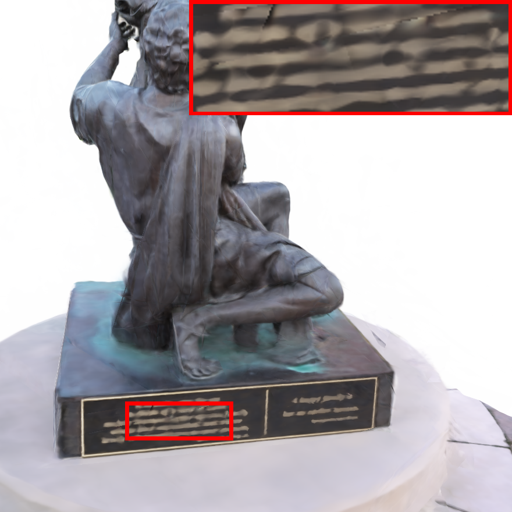} \\

\includegraphics[width=0.2\textwidth]{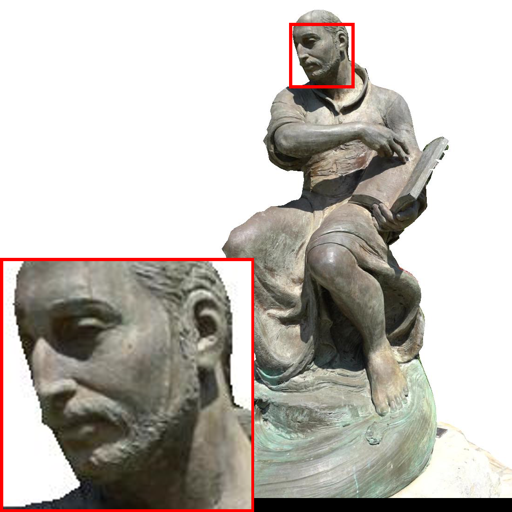} &
\includegraphics[width=0.2\textwidth]{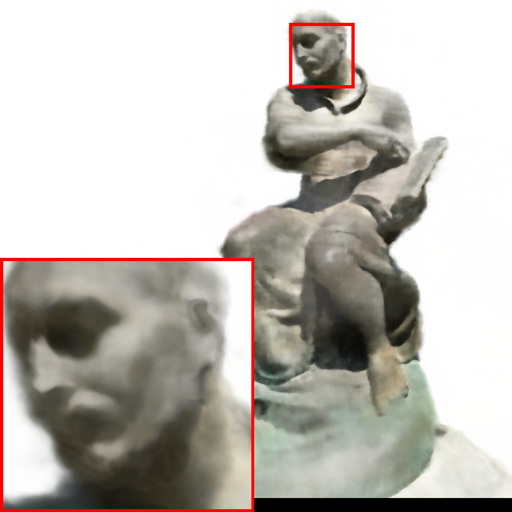} &
\includegraphics[width=0.2\textwidth]{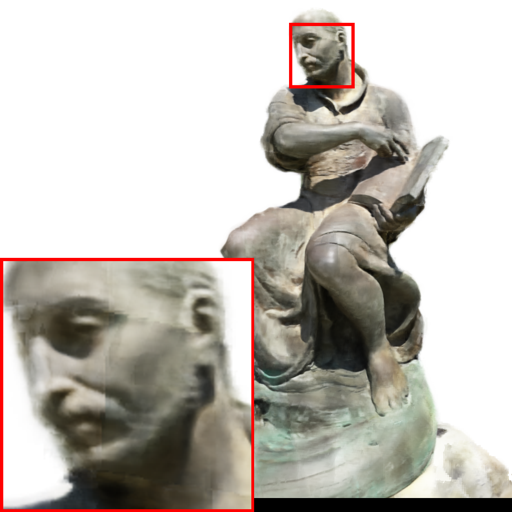} &
\includegraphics[width=0.2\textwidth]{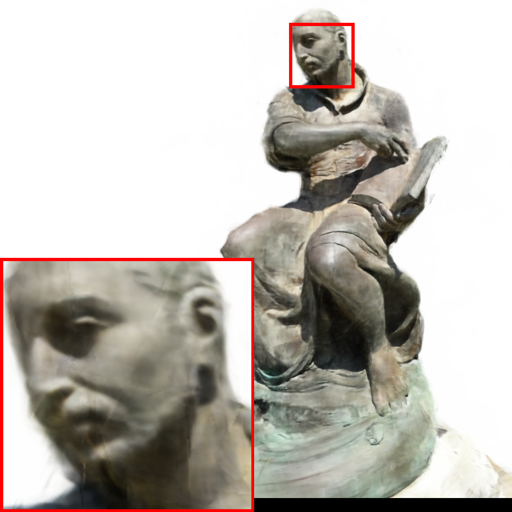} \\

\includegraphics[width=0.2\textwidth]{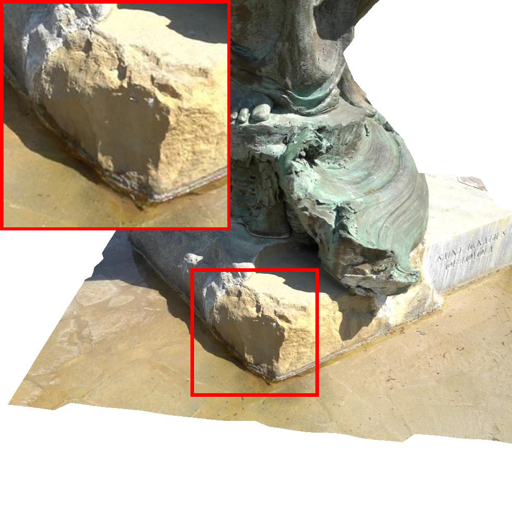} &
\includegraphics[width=0.2\textwidth]{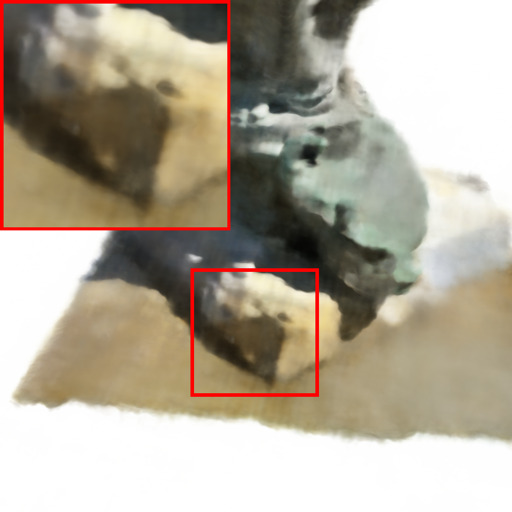} &
\includegraphics[width=0.2\textwidth]{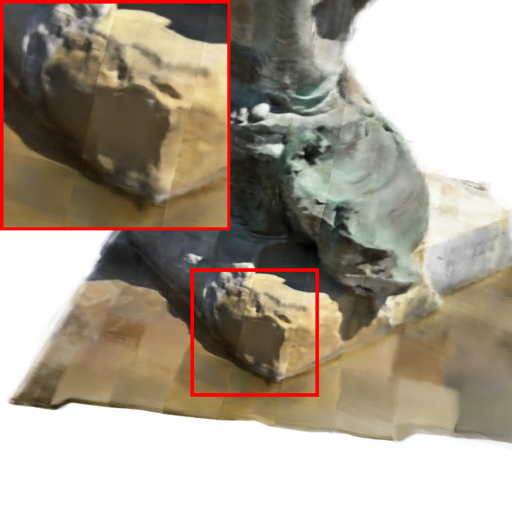} &
\includegraphics[width=0.2\textwidth]{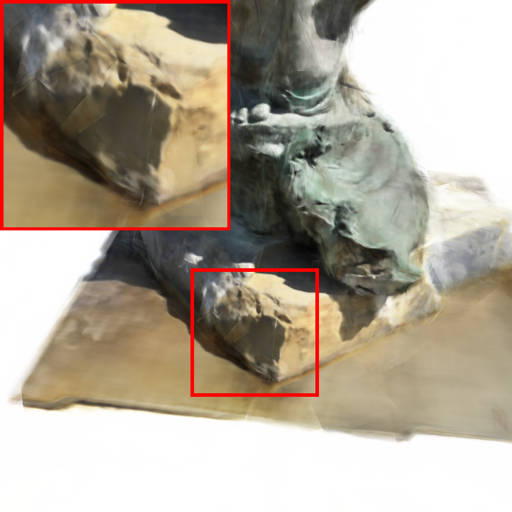} \\

\includegraphics[width=0.2\textwidth]{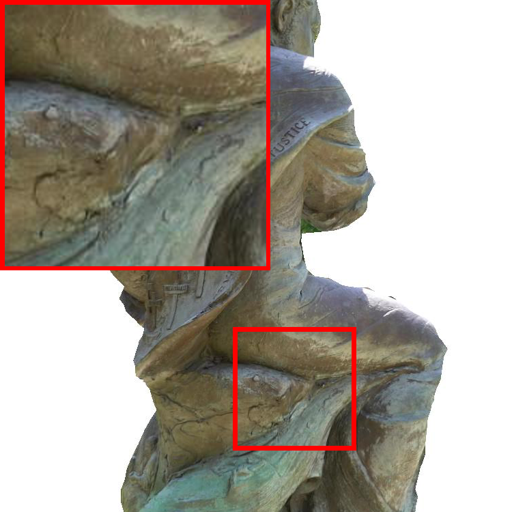} &
\includegraphics[width=0.2\textwidth]{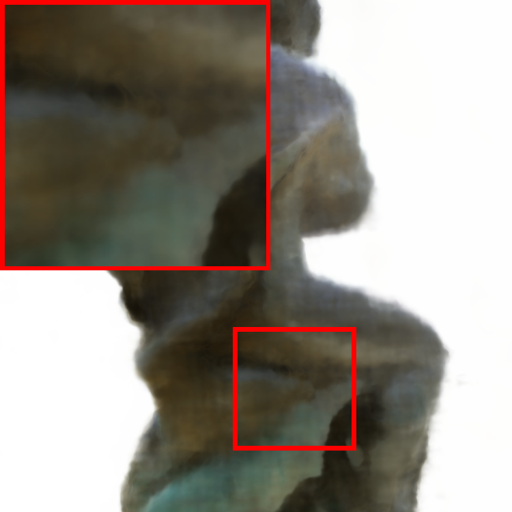} &
\includegraphics[width=0.2\textwidth]{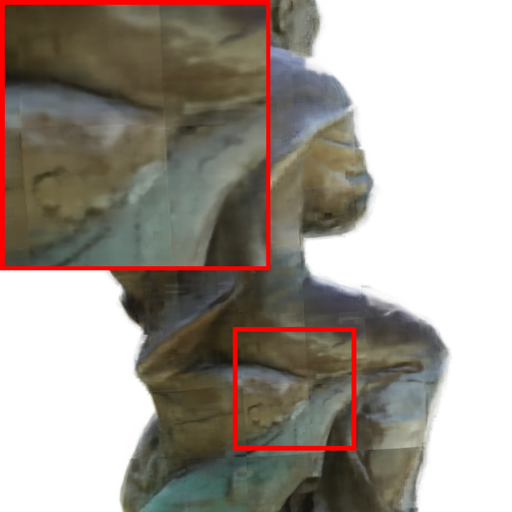} &
\includegraphics[width=0.2\textwidth]{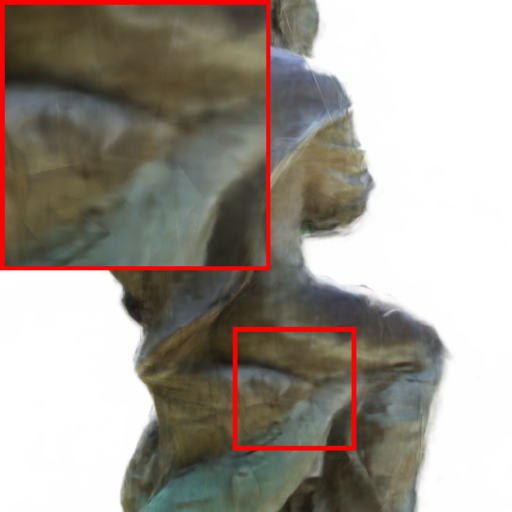} \\

\end{tabular}
}
\vspace{-3mm}
\captionof{figure}{Qualitative results for the scenes of Family (Tanks \& Temples)
and Ignatius (Tanks \& Temples)}
\label{fig:tanks_quality_2}
\vspace{-5mm}
\end{table*}

\begin{table*}[h]
\centering
\vspace{-5mm}
{
\setlength\tabcolsep{1.5pt} 
\begin{tabular}{cccc}

\multicolumn{1}{c}{Color Image} &
\multicolumn{1}{c}{NeRF~\cite{mildenhall2020nerf}} &
\multicolumn{1}{c}{KiloNeRF~\cite{reiser2021kilonerf}} &
\multicolumn{1}{c}{Ours} \\

\includegraphics[width=0.2\textwidth]{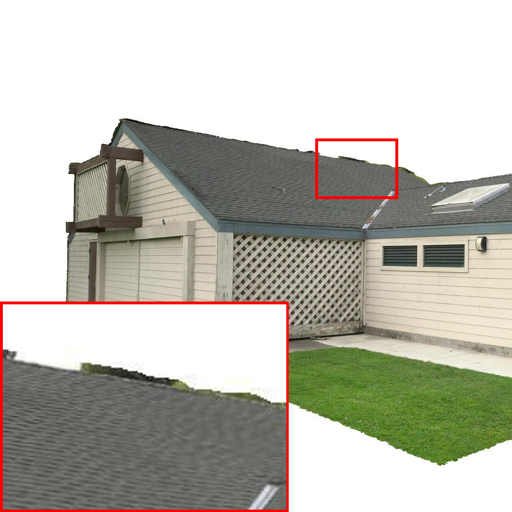} &
\includegraphics[width=0.2\textwidth]{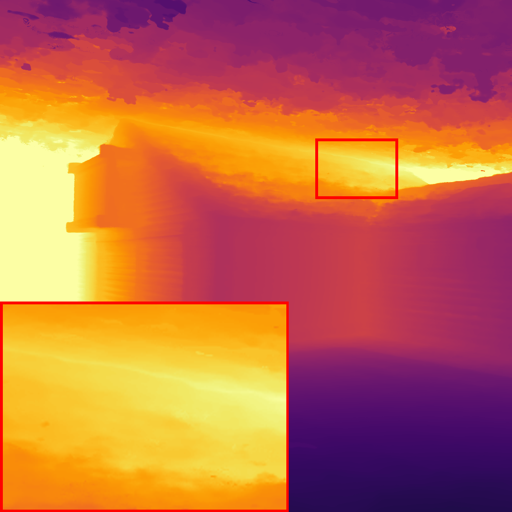} &
\includegraphics[width=0.2\textwidth]{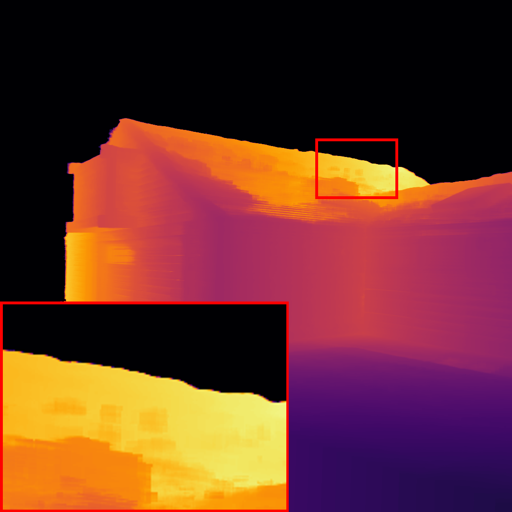} &
\includegraphics[width=0.2\textwidth]{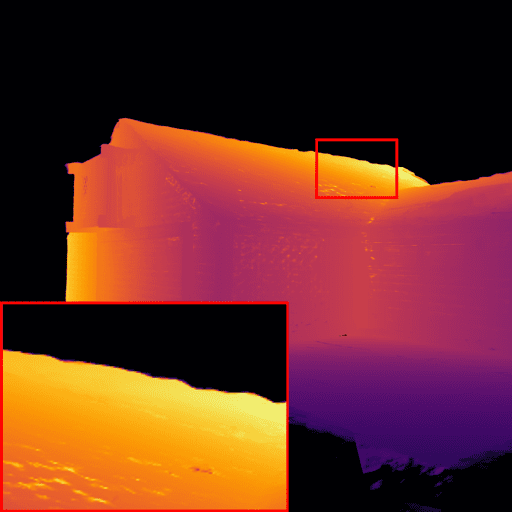} \\

\includegraphics[width=0.2\textwidth]{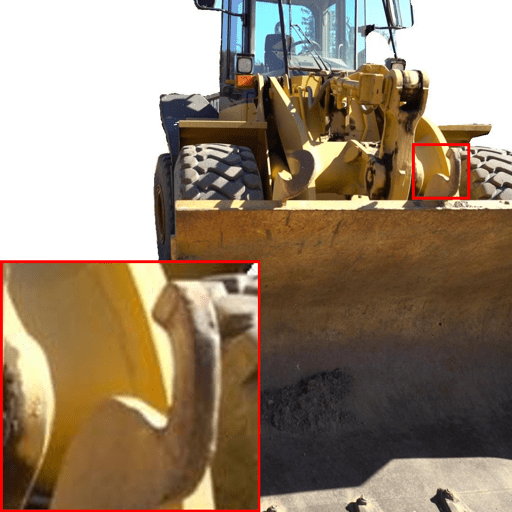} &
\includegraphics[width=0.2\textwidth]{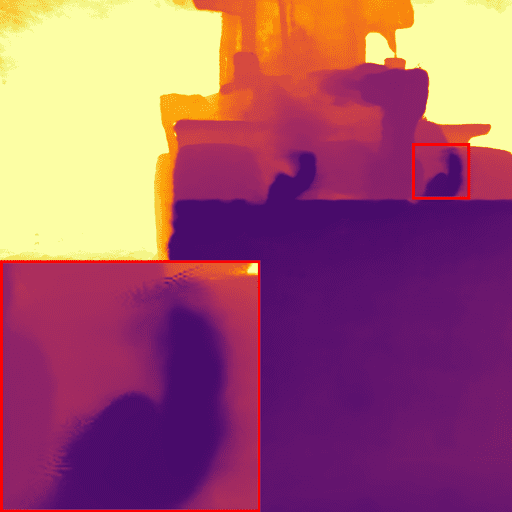} &
\includegraphics[width=0.2\textwidth]{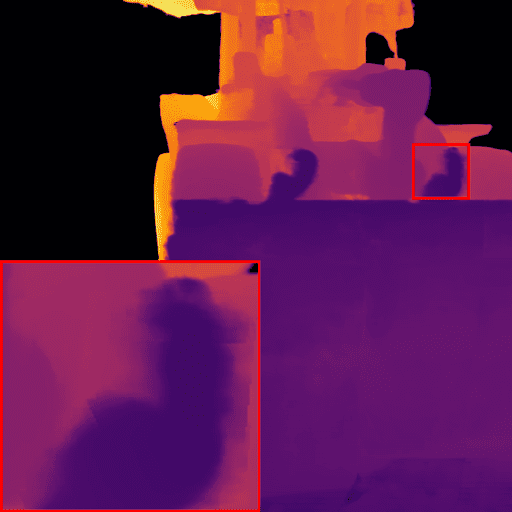} &
\includegraphics[width=0.2\textwidth]{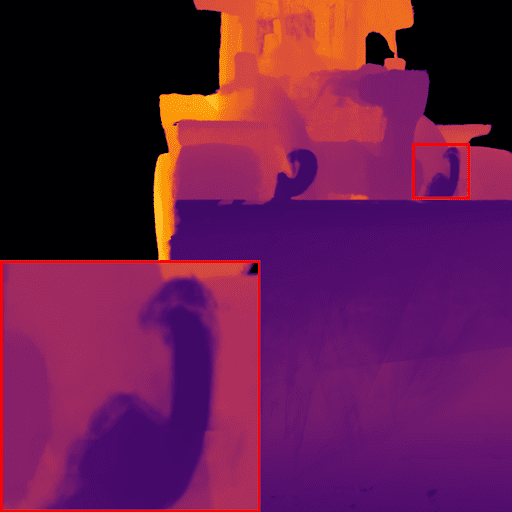} \\

\includegraphics[width=0.2\textwidth]{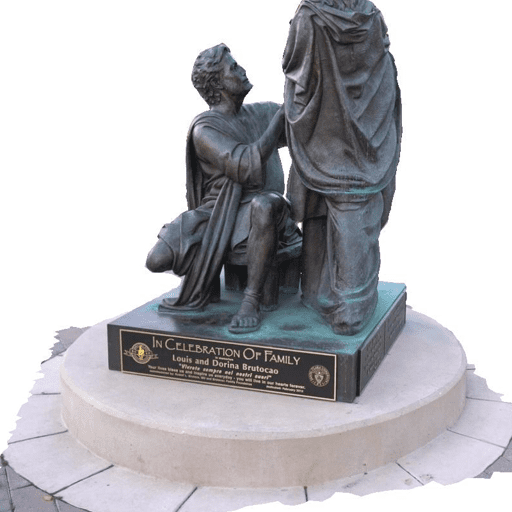} &
\includegraphics[width=0.2\textwidth]{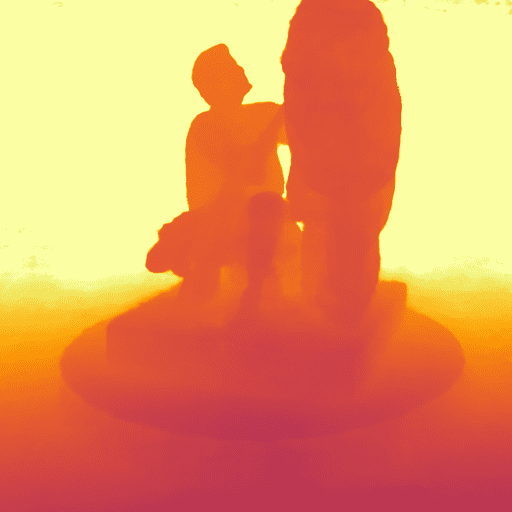} &
\includegraphics[width=0.2\textwidth]{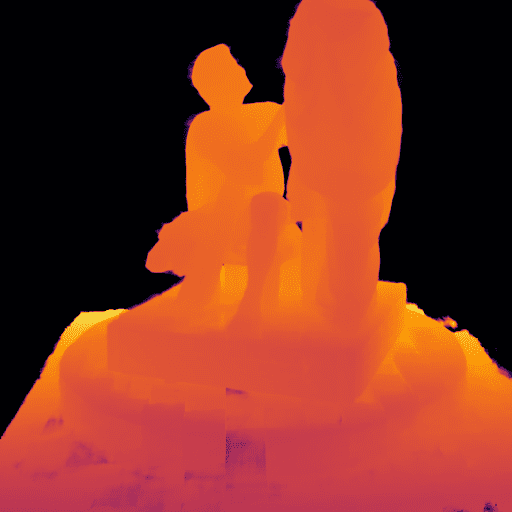} &
\includegraphics[width=0.2\textwidth]{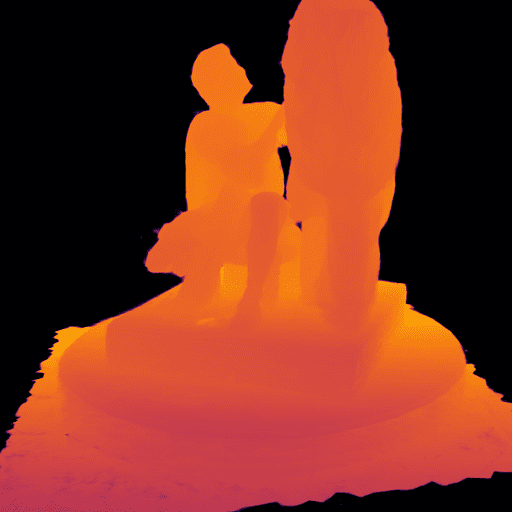} \\

\includegraphics[width=0.2\textwidth]{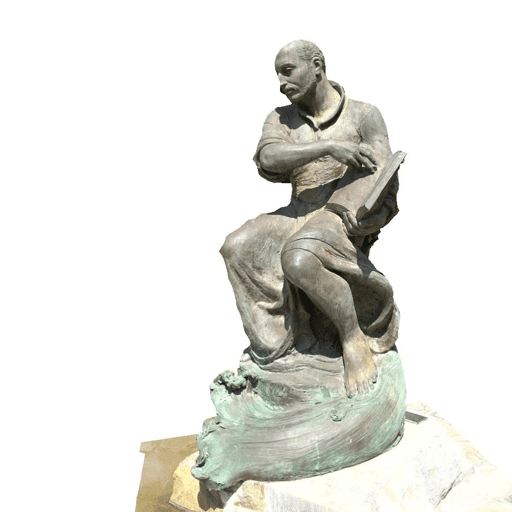} &
\includegraphics[width=0.2\textwidth]{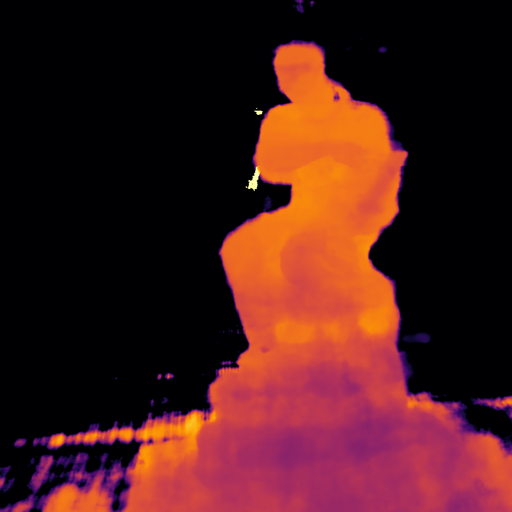} &
\includegraphics[width=0.2\textwidth]{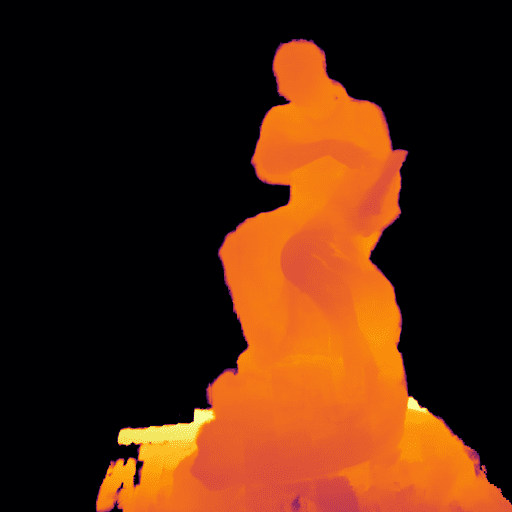} &
\includegraphics[width=0.2\textwidth]{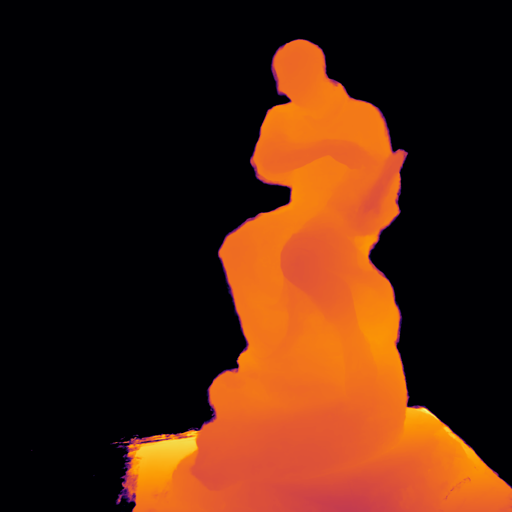} \\

\includegraphics[width=0.2\textwidth]{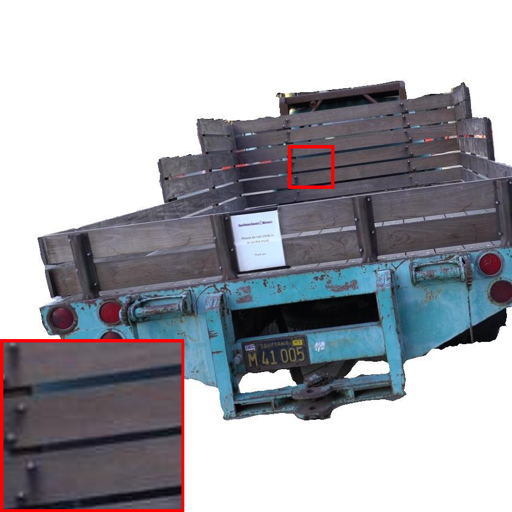} &
\includegraphics[width=0.2\textwidth]{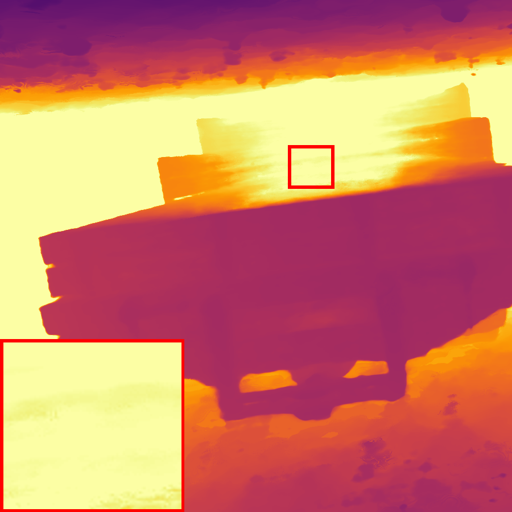} &
\includegraphics[width=0.2\textwidth]{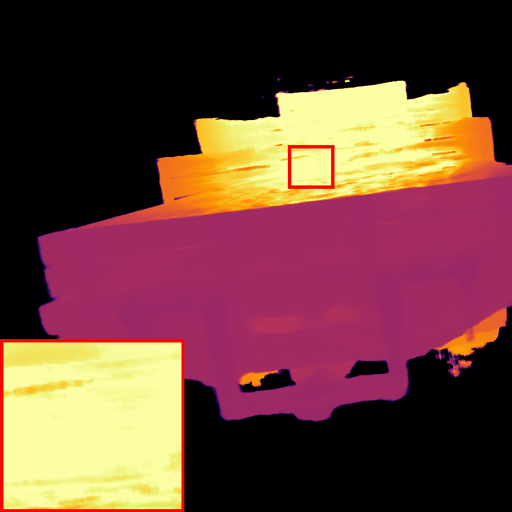} &
\includegraphics[width=0.2\textwidth]{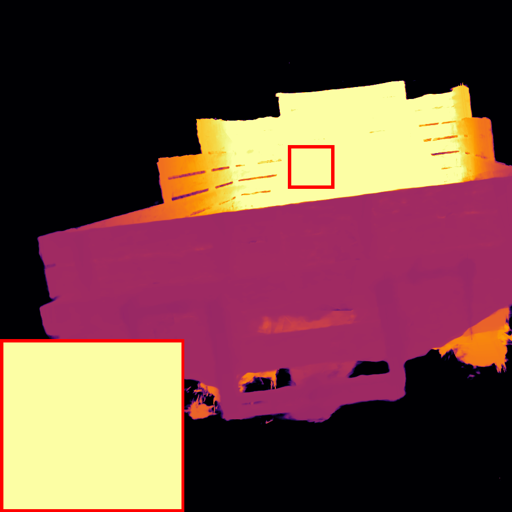} \\

\end{tabular}
}
\vspace{-3mm}
\captionof{figure}{Qualitative results of depth images for the scenes of Barn (Tanks \& Temples), Caterpillar (Tanks \& Temples), Family (Tanks \& Temples), Ignatius (Tanks \& Temples), and Truck (Tanks \& Temples)}
\label{fig:tanks_quality_3}
\vspace{-5mm}
\end{table*}

\begin{table*}[h]
\centering
\resizebox{.95\linewidth}{!}
{
\begin{tabular}{ ccccc } 

GT & NeRF~\cite{mildenhall2020nerf} & NeX~\cite{Wizadwongsa2021NeX} & KiloNeRF~\cite{reiser2021kilonerf} & Ours \\

\includegraphics[width=0.18\textwidth]{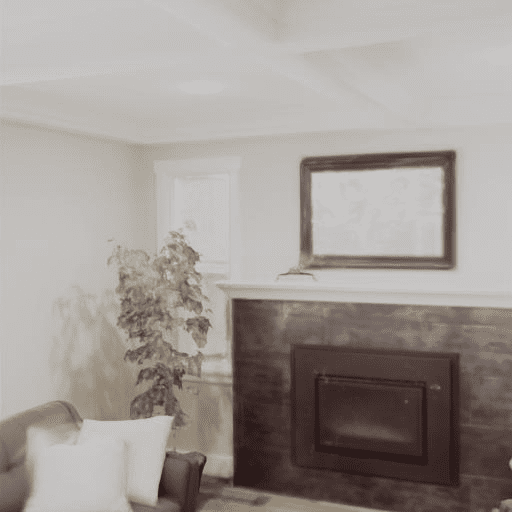} &
\includegraphics[width=0.18\textwidth]{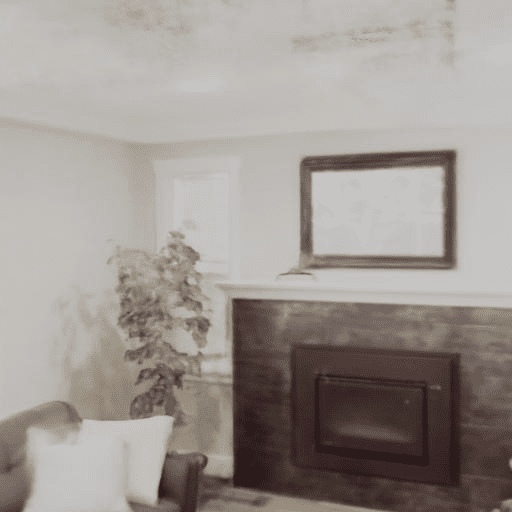} &
\includegraphics[width=0.18\textwidth]{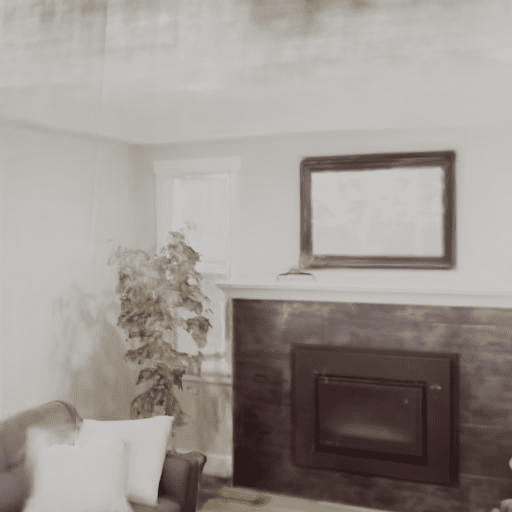} &
\includegraphics[width=0.18\textwidth]{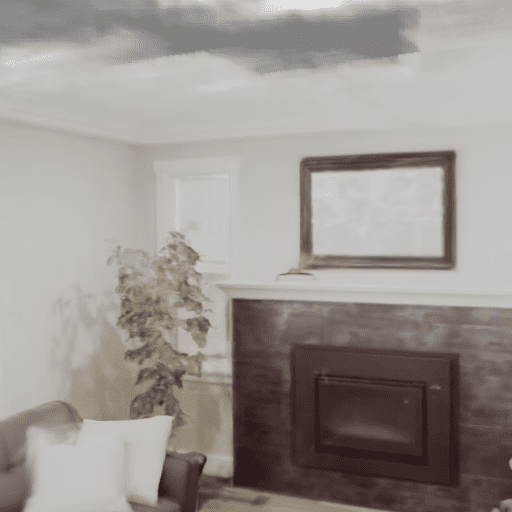} &
\includegraphics[width=0.18\textwidth]{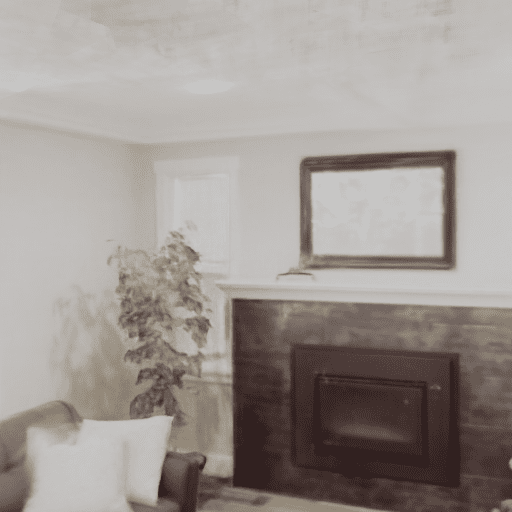} \\

\includegraphics[width=0.18\textwidth]{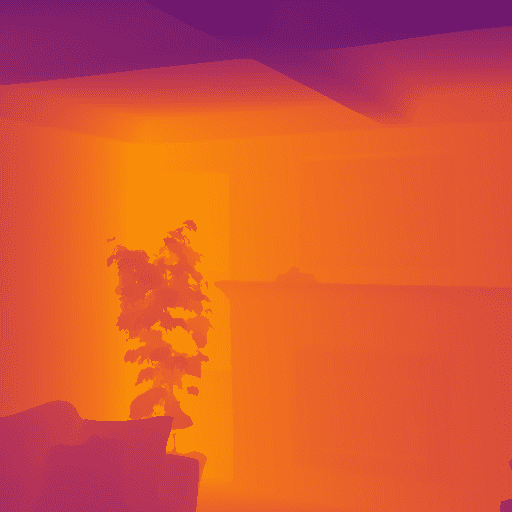} &
\includegraphics[width=0.18\textwidth]{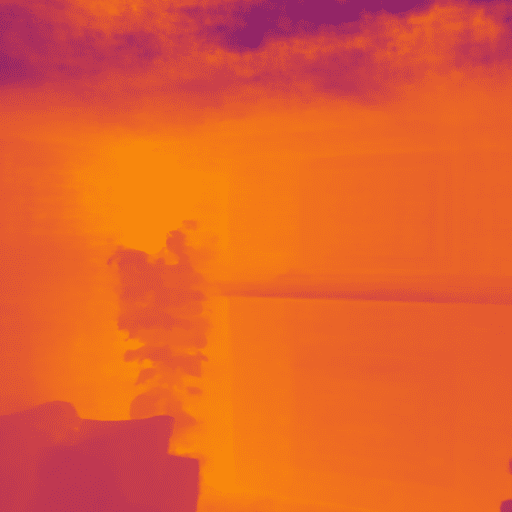} &
\includegraphics[width=0.18\textwidth]{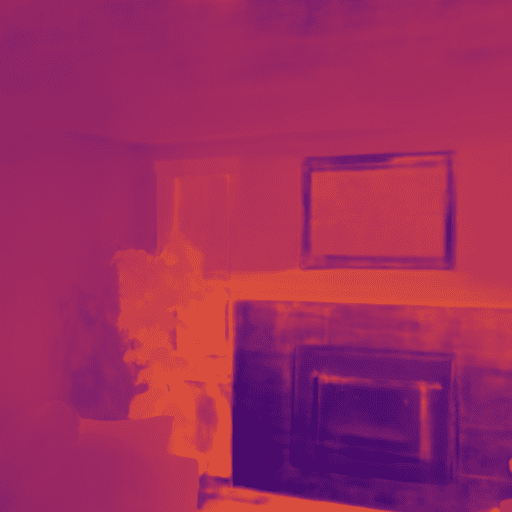} &
\includegraphics[width=0.18\textwidth]{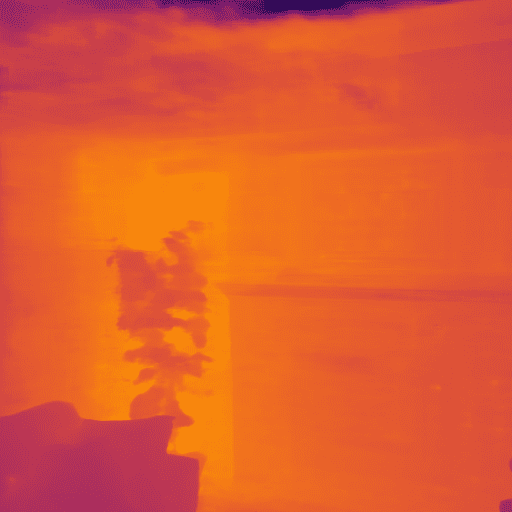} &
\includegraphics[width=0.18\textwidth]{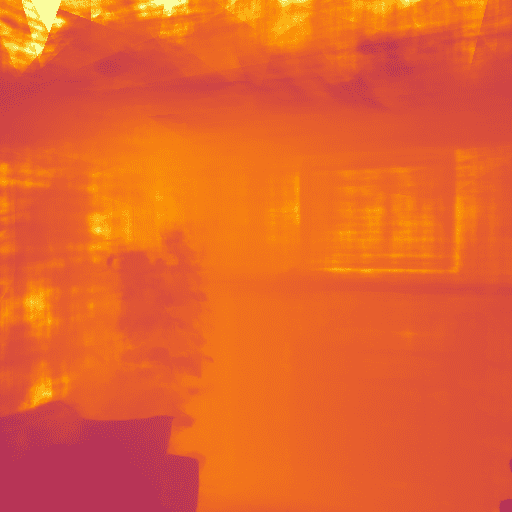} \\

\includegraphics[width=0.18\textwidth]{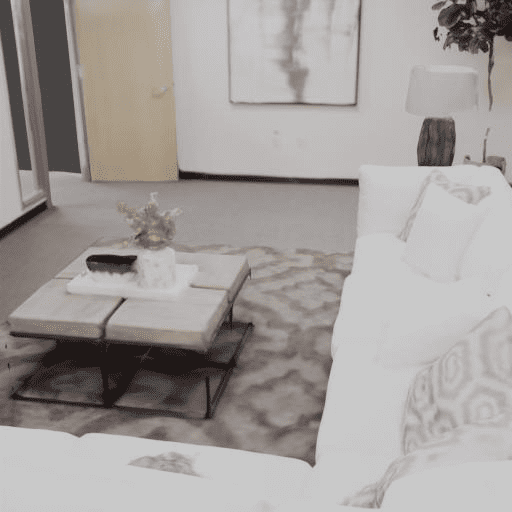} &
\includegraphics[width=0.18\textwidth]{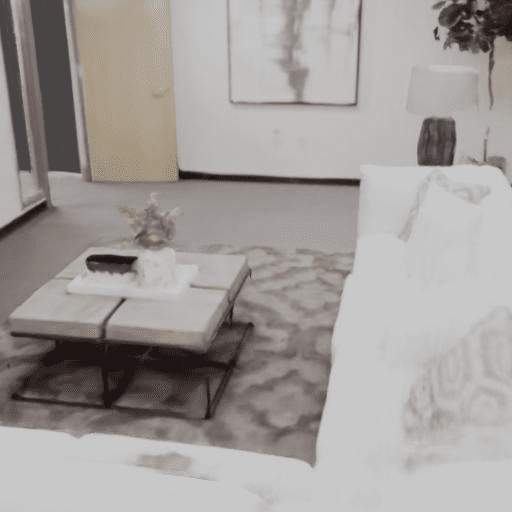} &
\includegraphics[width=0.18\textwidth]{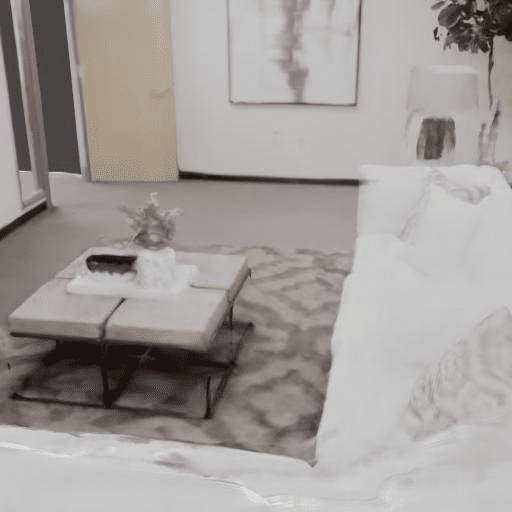} &
\includegraphics[width=0.18\textwidth]{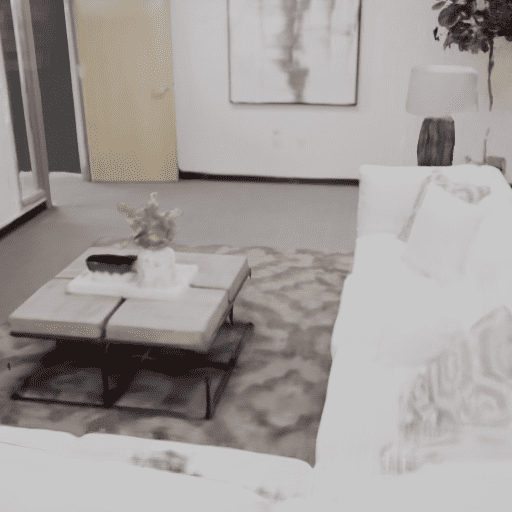} &
\includegraphics[width=0.18\textwidth]{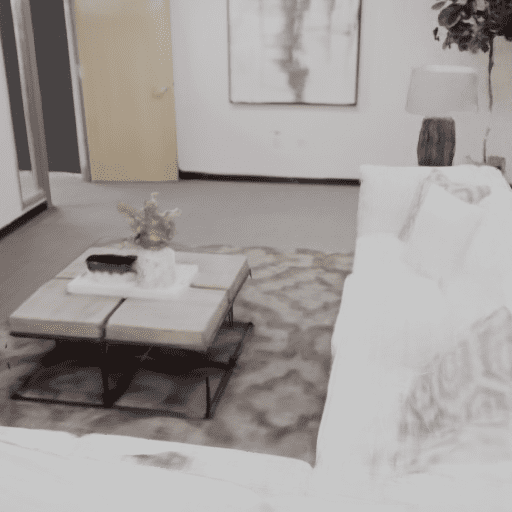} \\

\includegraphics[width=0.18\textwidth]{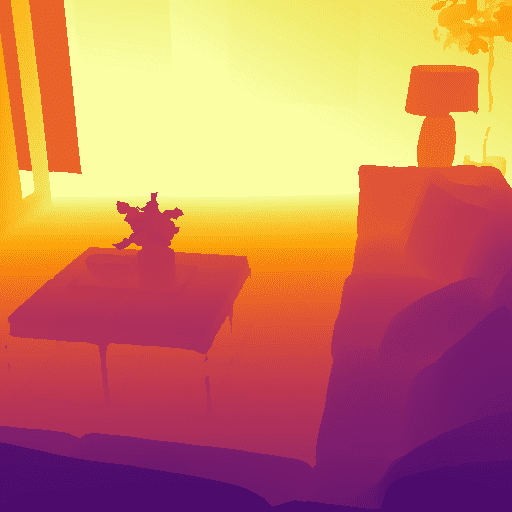} &
\includegraphics[width=0.18\textwidth]{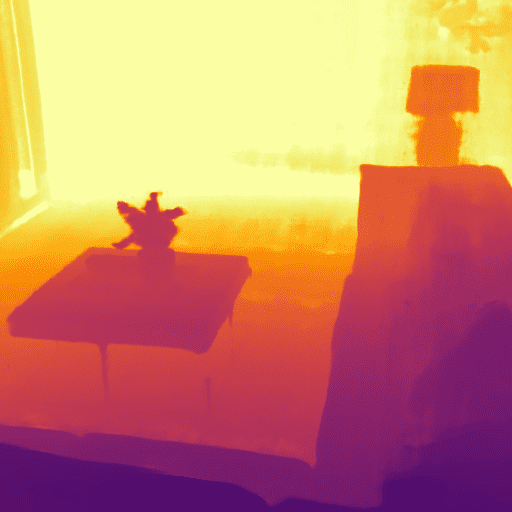} &
\includegraphics[width=0.18\textwidth]{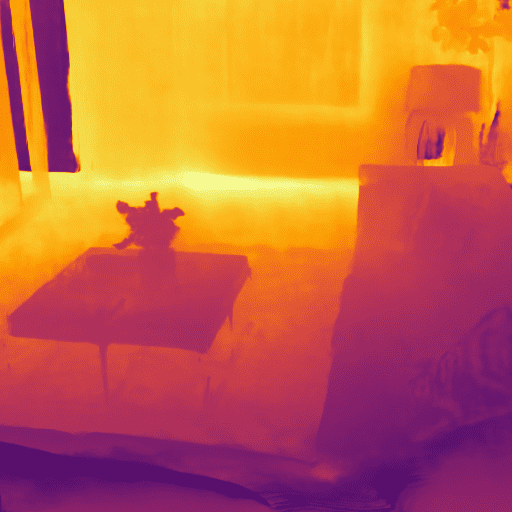} &
\includegraphics[width=0.18\textwidth]{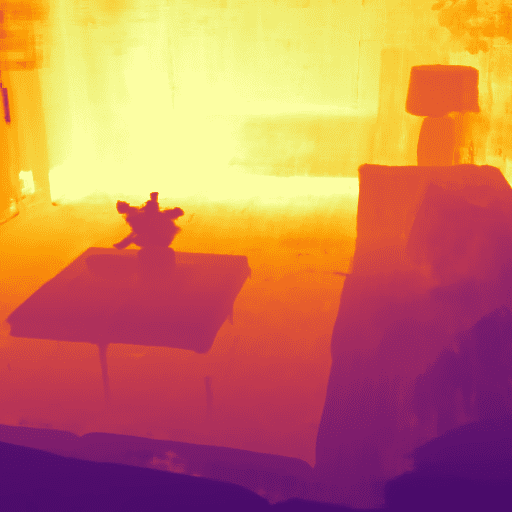} &
\includegraphics[width=0.18\textwidth]{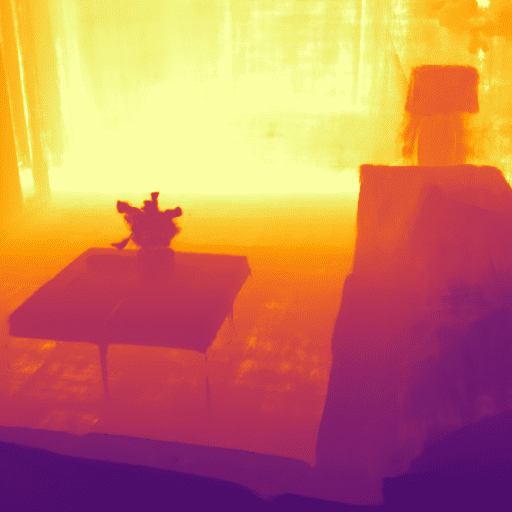} \\

\end{tabular}
}
\captionof{figure}{\textbf{Qualitative comparisons of rendered RGB and depth images of Replica.}}
\label{fig:replica_quality_1}
\end{table*}
\begin{table*}[h]
\centering
\resizebox{.95\linewidth}{!}
{
\begin{tabular}{ ccccc } 

GT & NeRF~\cite{mildenhall2020nerf} & NeX~\cite{Wizadwongsa2021NeX} & KiloNeRF~\cite{reiser2021kilonerf} & Ours \\

\includegraphics[width=0.18\textwidth]{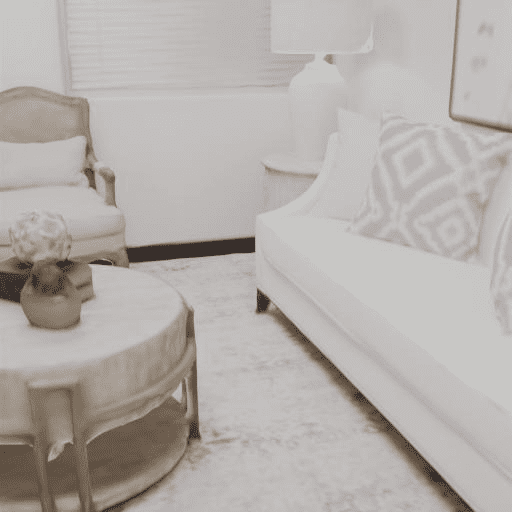} &
\includegraphics[width=0.18\textwidth]{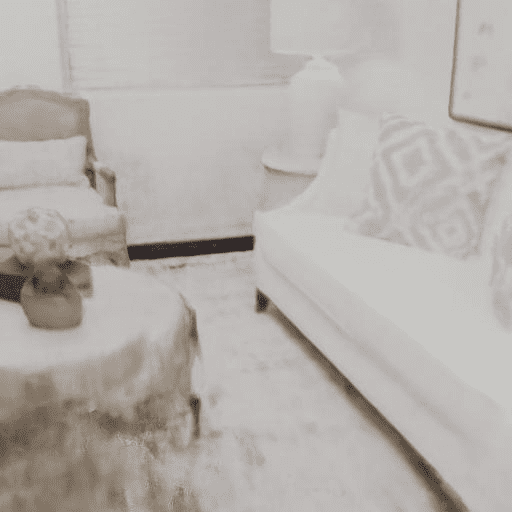} &
\includegraphics[width=0.18\textwidth]{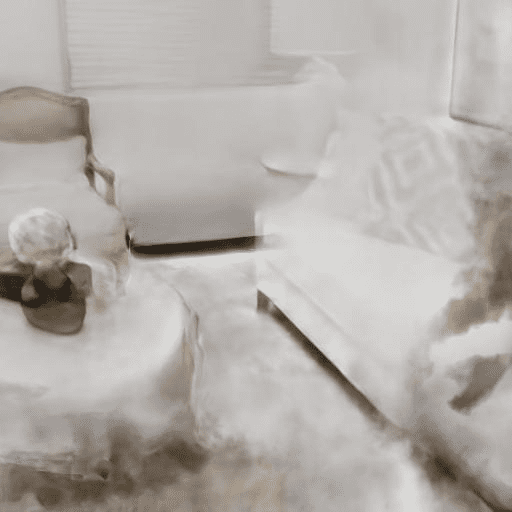} &
\includegraphics[width=0.18\textwidth]{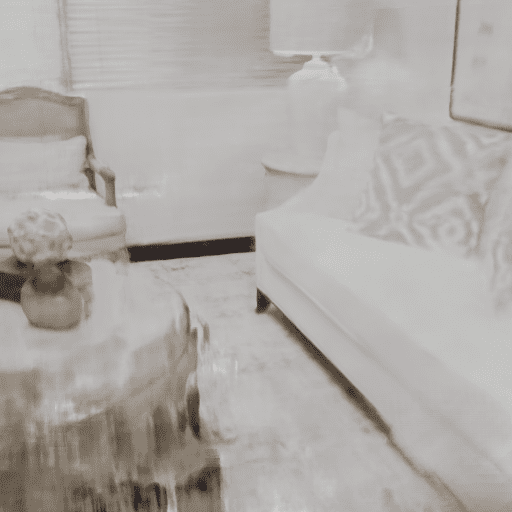} &
\includegraphics[width=0.18\textwidth]{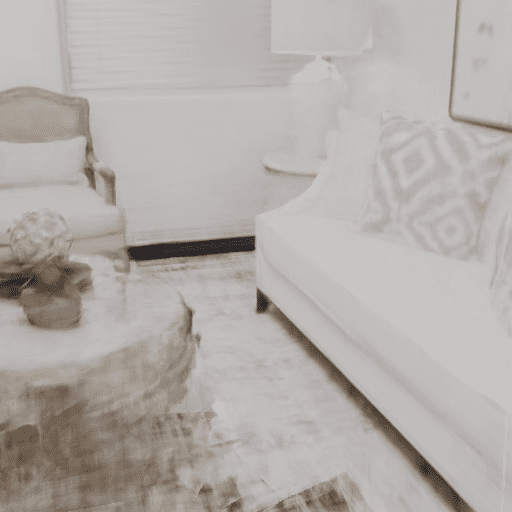} \\

\includegraphics[width=0.18\textwidth]{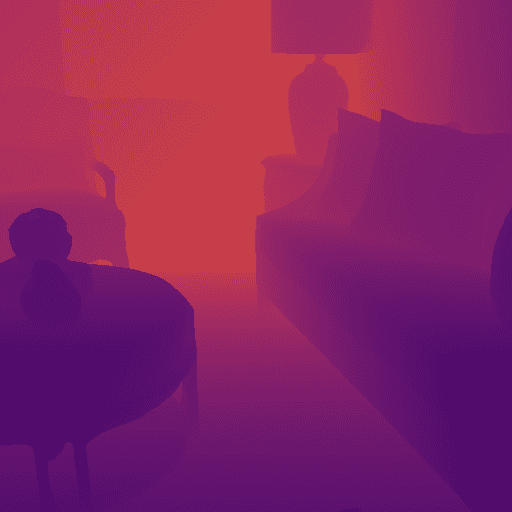} &
\includegraphics[width=0.18\textwidth]{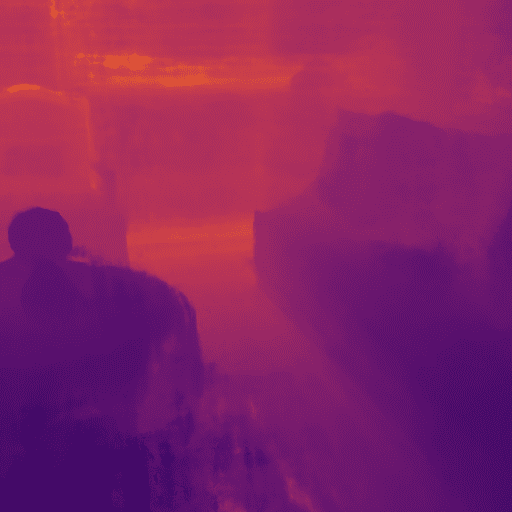} &
\includegraphics[width=0.18\textwidth]{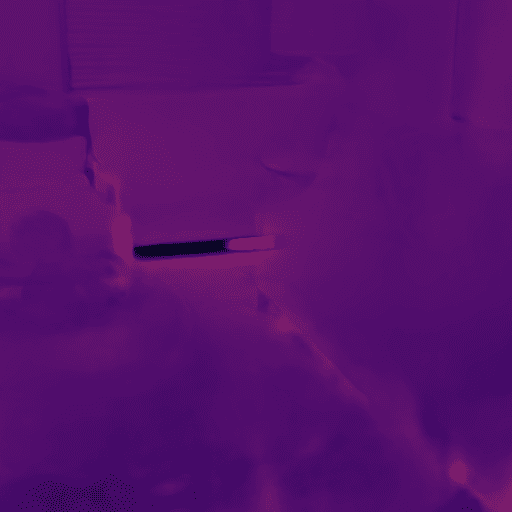} &
\includegraphics[width=0.18\textwidth]{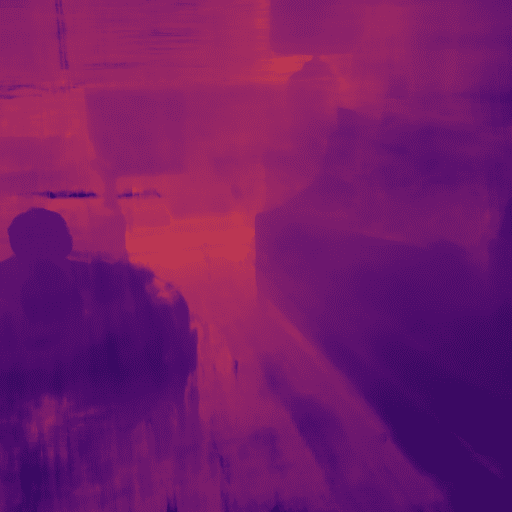} &
\includegraphics[width=0.18\textwidth]{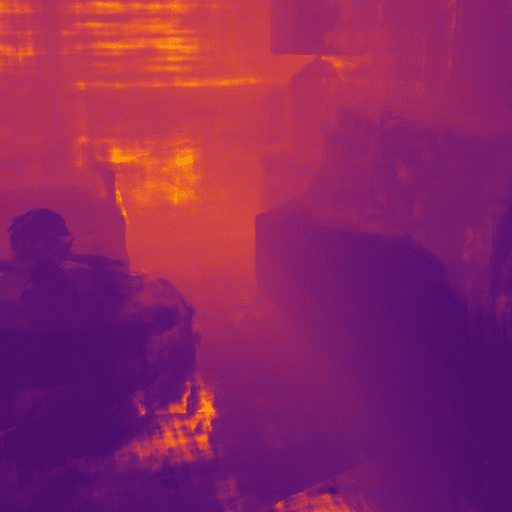} \\

\includegraphics[width=0.18\textwidth]{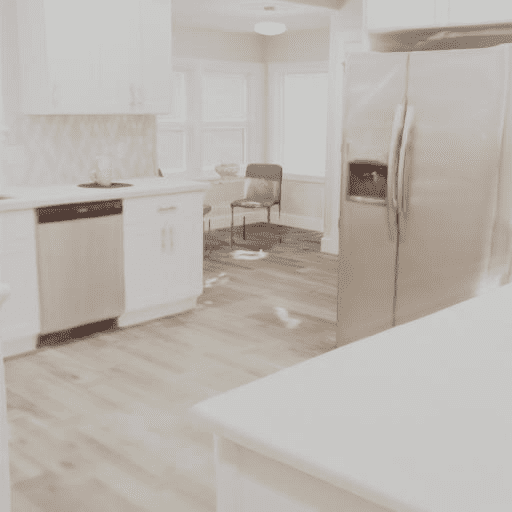} &
\includegraphics[width=0.18\textwidth]{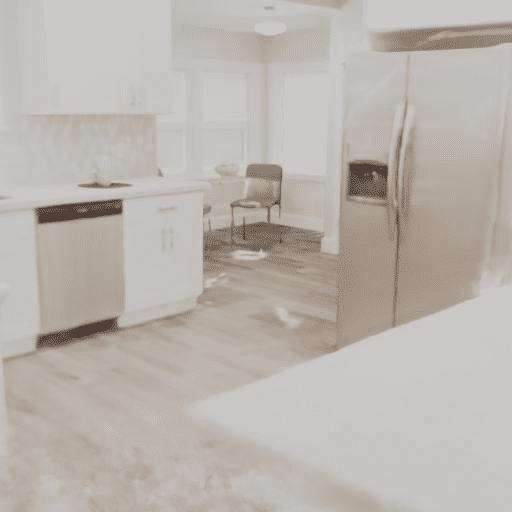} &
\includegraphics[width=0.18\textwidth]{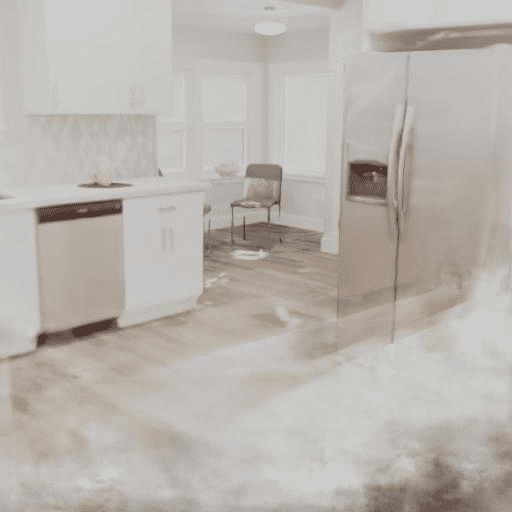} &
\includegraphics[width=0.18\textwidth]{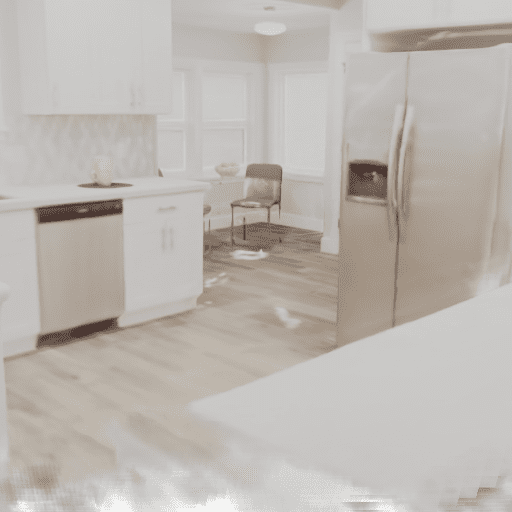} &
\includegraphics[width=0.18\textwidth]{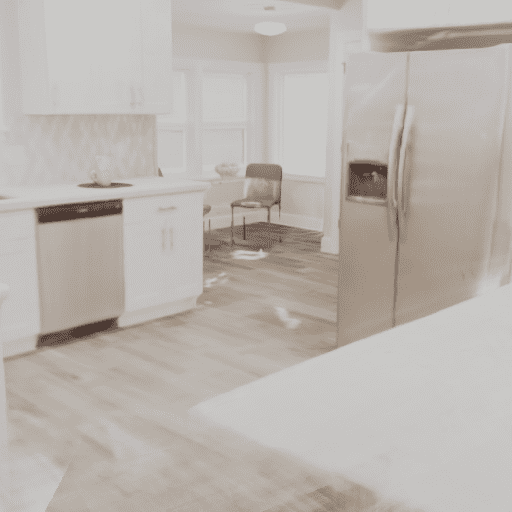} \\

\includegraphics[width=0.18\textwidth]{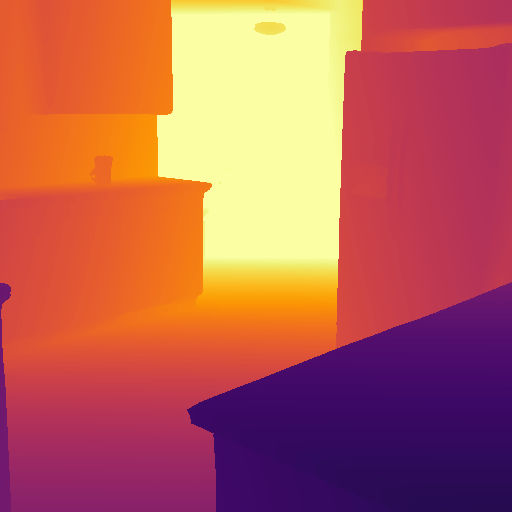} &
\includegraphics[width=0.18\textwidth]{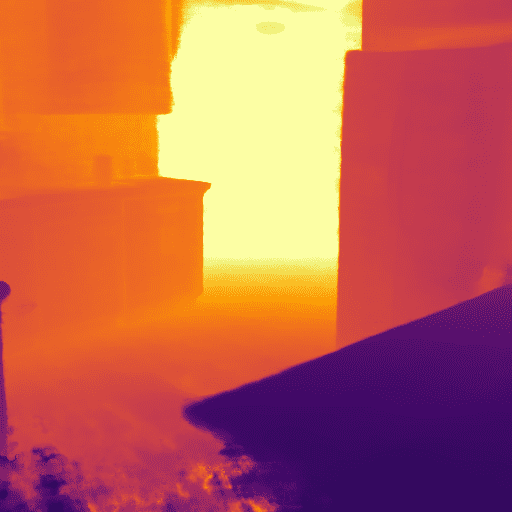} &
\includegraphics[width=0.18\textwidth]{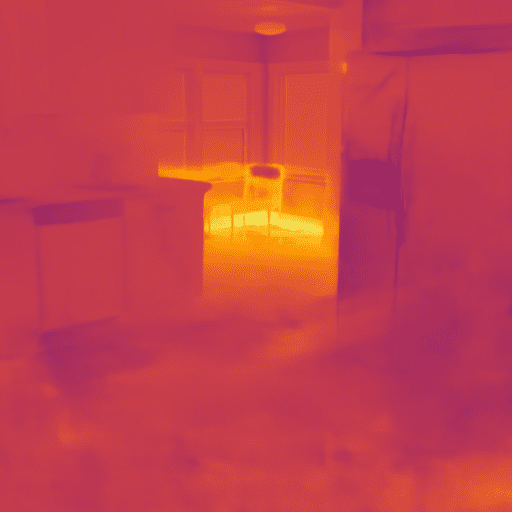} &
\includegraphics[width=0.18\textwidth]{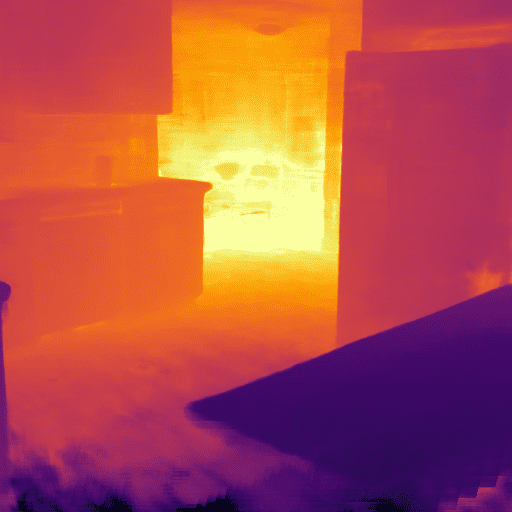} &
\includegraphics[width=0.18\textwidth]{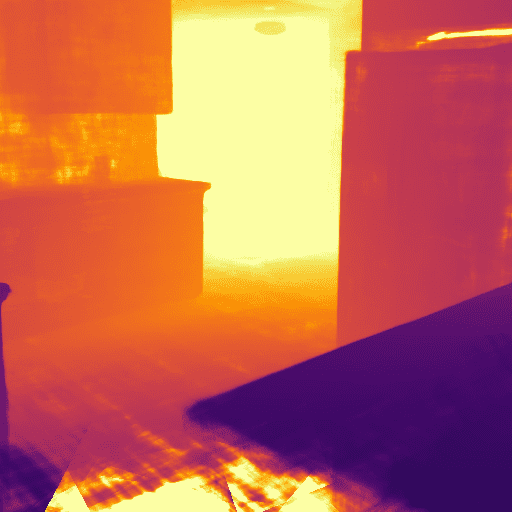} \\

\end{tabular}
}
\captionof{figure}{\textbf{Qualitative comparisons of rendered RGB and depth images of Replica.}}
\label{fig:replica_quality_2}
\end{table*}
\begin{table*}[h]
\centering
\vspace{-5mm}
{
\setlength\tabcolsep{1.5pt}
\begin{tabular}{cccc}

\multicolumn{1}{c}{Ground-Truth} &
\multicolumn{1}{c}{NeRF~\cite{mildenhall2020nerf}} &
\multicolumn{1}{c}{KiloNeRF~\cite{reiser2021kilonerf}} &
\multicolumn{1}{c}{Ours} \\

\includegraphics[width=0.2\textwidth]{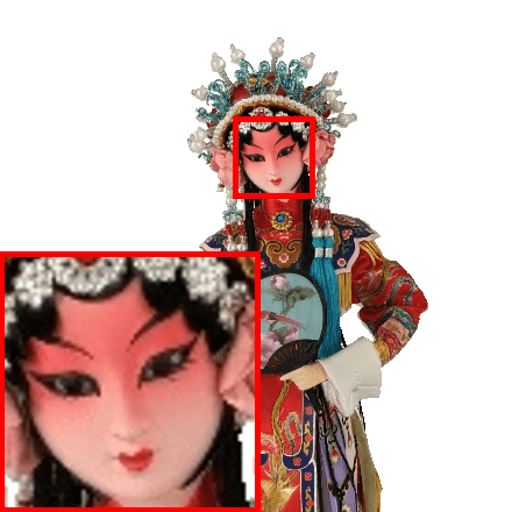} &
\includegraphics[width=0.2\textwidth]{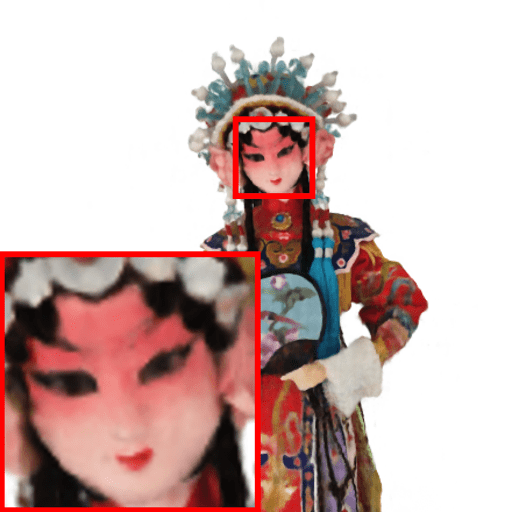} &
\includegraphics[width=0.2\textwidth]{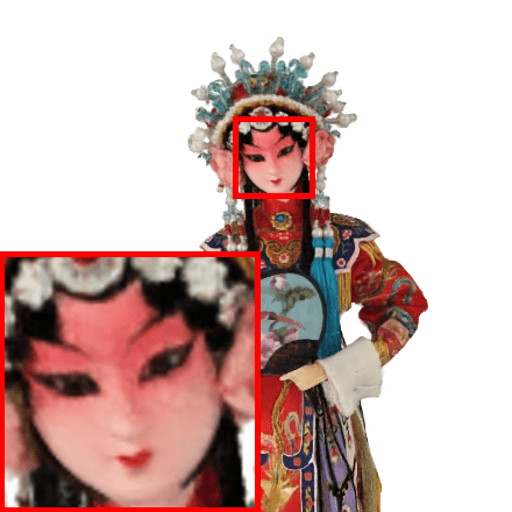} &
\includegraphics[width=0.2\textwidth]{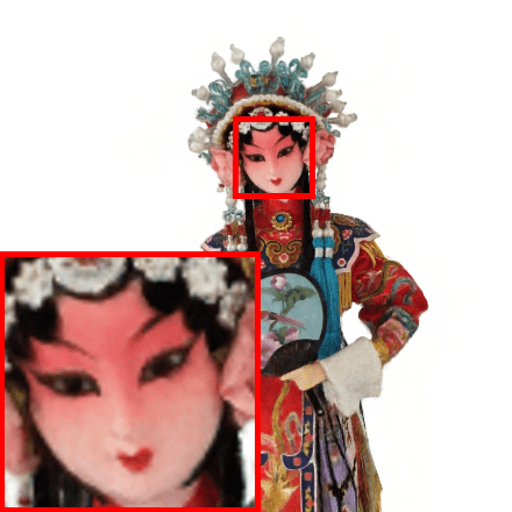} \\

\includegraphics[width=0.2\textwidth]{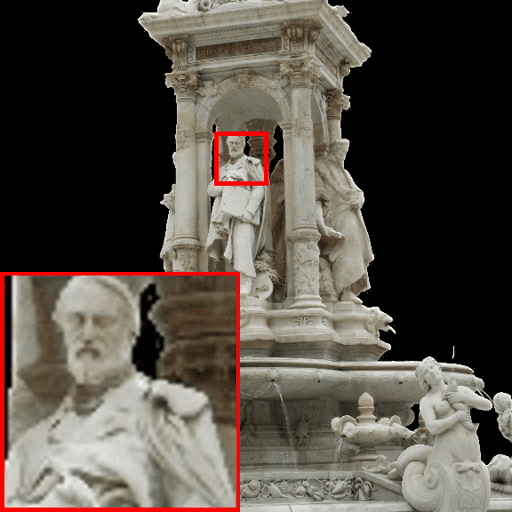} &
\includegraphics[width=0.2\textwidth]{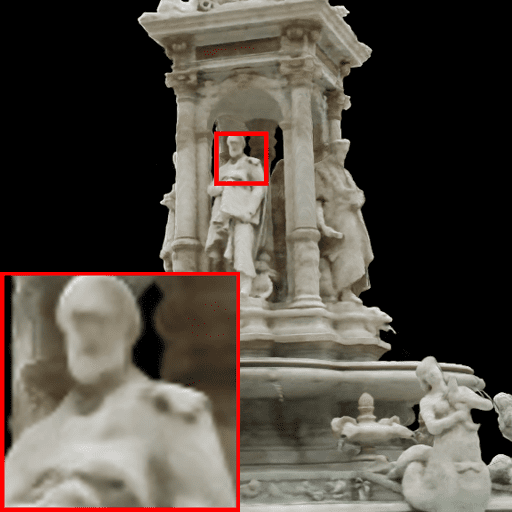} &
\includegraphics[width=0.2\textwidth]{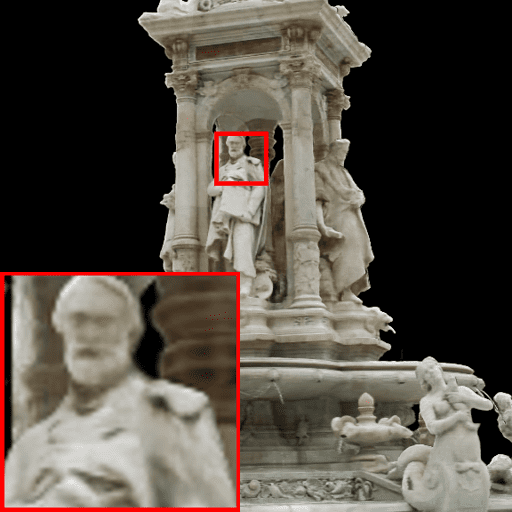} &
\includegraphics[width=0.2\textwidth]{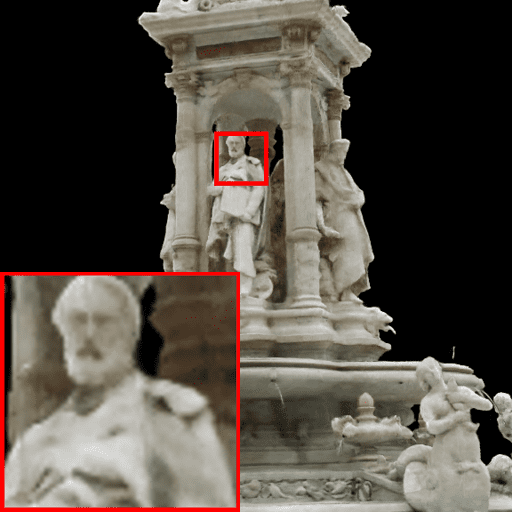} \\

\includegraphics[width=0.2\textwidth]{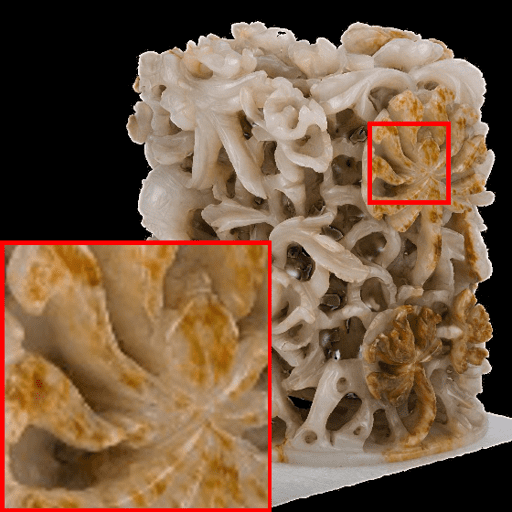} &
\includegraphics[width=0.2\textwidth]{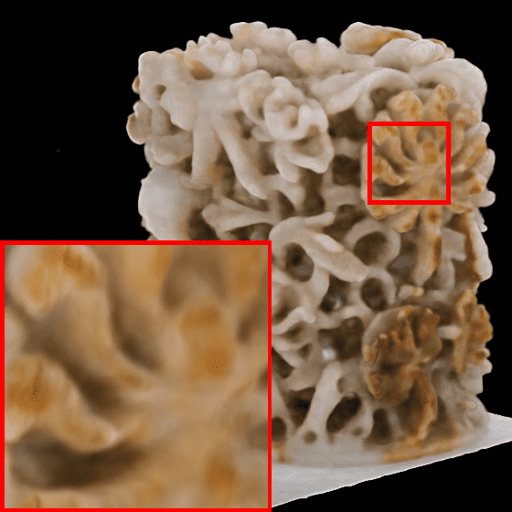} &
\includegraphics[width=0.2\textwidth]{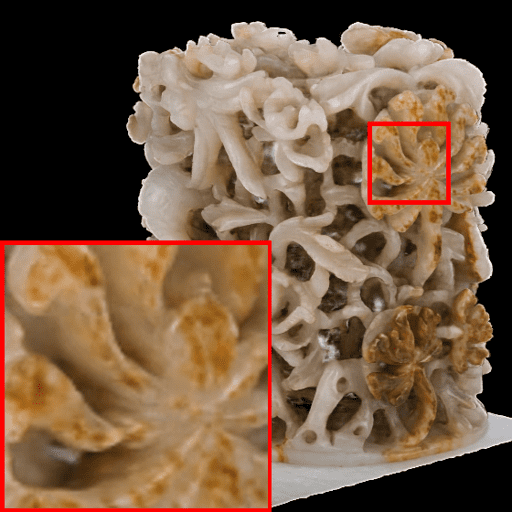} &
\includegraphics[width=0.2\textwidth]{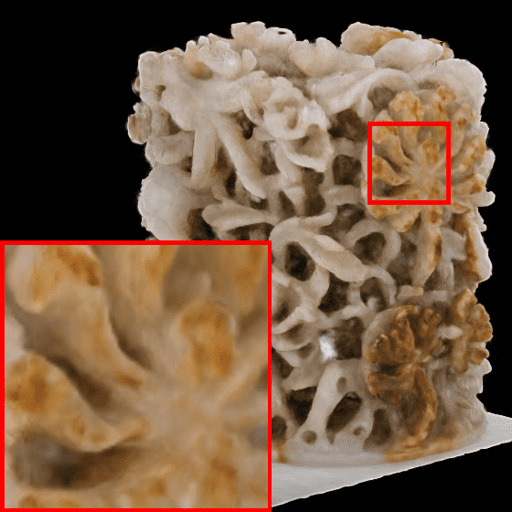} \\

\includegraphics[width=0.2\textwidth]{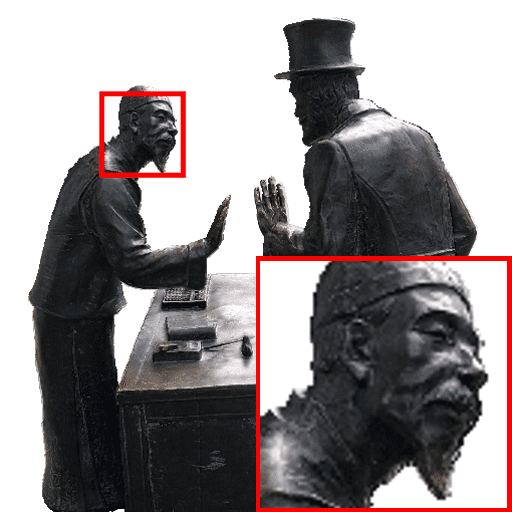} &
\includegraphics[width=0.2\textwidth]{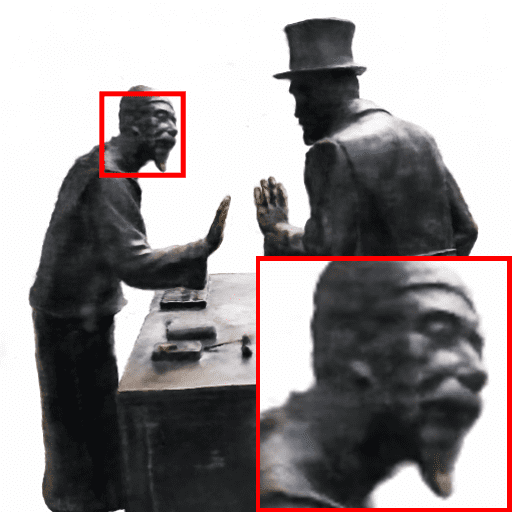} &
\includegraphics[width=0.2\textwidth]{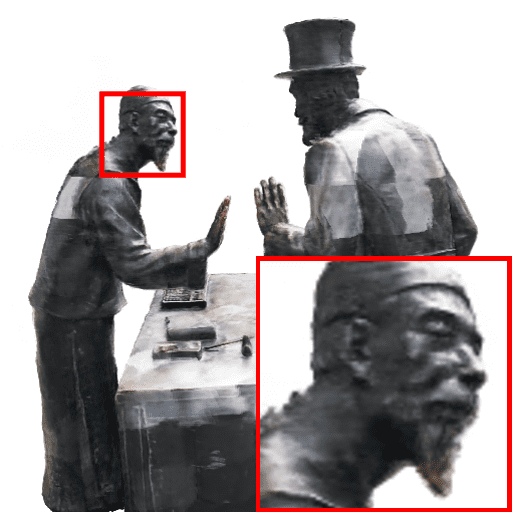} &
\includegraphics[width=0.2\textwidth]{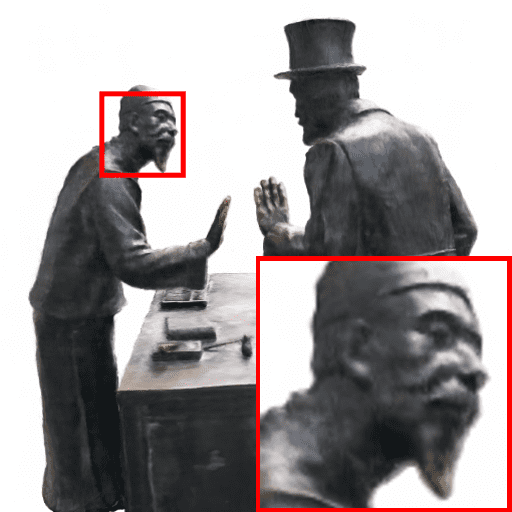} \\

\end{tabular}
}
\vspace{-3mm}
\captionof{figure}{Qualitative results for the scenes of Character (BlendedMVS), Fountain (BlendedMVS), Jade (BlendedMVS), Statues (BlendedMVS).}
\label{fig:blendedmvs_quality}
\vspace{-5mm}
\end{table*}

\section{More detailed quantitative results}\label{sect:quantitative}

The detailed quantitative results of per-scene breakdown are reported in Tab.~\ref{tab:quantity_tanks},
Tab.~\ref{tab:quantity_replica} and Tab.~\ref{tab:blendedmvs_quantity}. In Tab.~\ref{tab:quantity_tanks}, we want to highlight that while \method~achieve comparable performance on human statues (\eg Family, Ignatius), it outperforms others in the scenes which are rich in planar structures (\eg Barn, Caterpillar, Truck), and this demonstrates the strength of the effectiveness of our planar experts. Tab.~\ref{tab:quantity_replica} shows that our method can produce high-quality images in challenging novel views and outperform the current state-of-the-art MPI-based method~\cite{Wizadwongsa2021NeX} by a large margin. Tab.~\ref{tab:blendedmvs_quantity} shows that our model produces reasonable results in dataset BlendedMVS~\cite{yao2020blendedmvs}. However, it is challenging for our planes to fit the complex interior surface structures (\eg ``Jade'', ``Fountain''). We conjecture the reasons are two-fold: 1) limited expressiveness of our mixture of planar representations for such complicated structure 2) poor geometry initialization from COLMAP in these two scenes. We leave these challenging cases for future work. 

\begin{table*}[h]
\centering
{
\vspace{-5mm}

{

    \begin{tabular}{ lccccc}
    
        \multicolumn{6}{c}{PSNR$\uparrow$} \\
        \hline
        \multicolumn{1}{c}{} & \multicolumn{1}{c}{Barn} & \multicolumn{1}{c}{Caterpillar} & \multicolumn{1}{c}{Family} & \multicolumn{1}{c}{Ignatius} & \multicolumn{1}{c}{Truck} \\
        \hline
        NeRF (original) & 24.05 & 23.75 & 30.29 & 25.43 & 25.36 \\
        NeRF            & 27.39 & 25.24 & 32.47 & 27.95 & 26.66 \\
        SRN             & 22.44 & 21.14 & 27.57 & 26.70 & 22.62 \\
        Neural Volumes  & 20.82 & 20.71 & 28.72 & 26.54 & 21.71 \\
        NSVF            & 27.16 & \textbf{26.44} & 33.58 & 27.91 & 26.92 \\
        KiloNeRF        & 27.81 & 25.61 & \textbf{33.65} & 27.92 & 27.04 \\
        PlenOctree      & 26.80 & 25.29 & 32.85 & \textbf{28.19} & 26.83 \\
        \hline
        Ours            & \textbf{28.04} & 26.40 & 32.77 & 27.96 & \textbf{27.15} \\
        \hline
    
    \end{tabular}
}

{

    \begin{tabular}{ lccccc}
    
        \multicolumn{6}{c}{SSIM$\uparrow$} \\
        \hline
        \multicolumn{1}{c}{} & \multicolumn{1}{c}{Barn} & \multicolumn{1}{c}{Caterpillar} & \multicolumn{1}{c}{Family} & \multicolumn{1}{c}{Ignatius} & \multicolumn{1}{c}{Truck} \\
        \hline
        NeRF (original) & 0.750 & 0.860 & 0.932 & 0.920 & 0.860 \\
        NeRF            & 0.842 & 0.892 & 0.951 & 0.940 & 0.896 \\
        SRN             & 0.741 & 0.834 & 0.908 & 0.920 & 0.832 \\
        Neural Volumes  & 0.721 & 0.819 & 0.916 & 0.922 & 0.793 \\
        NSVF            & 0.823 & 0.900 & 0.954 & 0.930 & 0.895 \\
        KiloNeRF        & 0.850 & 0.900 & 0.960 & 0.940 & 0.900 \\
        PlenOctree      & 0.856 & \textbf{0.907} & \textbf{0.962} & \textbf{0.948} & \textbf{0.914} \\
        \hline
        Ours            & \textbf{0.858} & 0.901 & 0.948 & 0.936 & 0.896 \\
        \hline
    
    \end{tabular}
}

{
    \begin{tabular}{ lccccc} 
    
        \multicolumn{6}{c}{LPIPS$\downarrow$} \\
        \hline
        \multicolumn{1}{c}{} & \multicolumn{1}{c}{Barn} & \multicolumn{1}{c}{Caterpillar} & \multicolumn{1}{c}{Family} & \multicolumn{1}{c}{Ignatius} & \multicolumn{1}{c}{Truck} \\
        \hline
        NeRF (original) & 0.395 & 0.196 & 0.098 & 0.111 & 0.192 \\
        NeRF            & 0.286 & 0.189 & 0.092 & 0.102 & 0.173 \\
        SRN             & 0.448 & 0.278 & 0.134 & 0.128 & 0.266 \\
        Neural Volumes  & 0.479 & 0.280 & 0.111 & 0.117 & 0.312 \\
        NSVF            & 0.307 & 0.141 & 0.063 & 0.106 & 0.148 \\
        KiloNeRF        & 0.160 & 0.100 & \textbf{0.040} & \textbf{0.060} & \textbf{0.100} \\
        PlenOctree      & 0.226 & 0.148 & 0.069 & 0.080 & 0.130 \\
        \hline
        Ours            & \textbf{0.140} & \textbf{0.091} & 0.050 & 0.063 & 0.103 \\
        \hline
    
    \end{tabular}
}

\bigskip

\vspace{-3mm}
\caption{\textbf{Quantitative results on Tanks~\&~Temples.}}
\label{tab:quantity_tanks}
\vspace{-1mm}
}
\end{table*}
\begin{table*}[h]
\centering
{
{
    \begin{tabular}{ lccccccc}
    
        \multicolumn{8}{c}{PSNR$\uparrow$} \\
        \hline
        \multicolumn{1}{c}{} & \multicolumn{1}{c}{Apartment 0} & \multicolumn{1}{c}{Apartment 1} & \multicolumn{1}{c}{Apartment 2} & \multicolumn{1}{c}{FRL 0} & \multicolumn{1}{c}{Kitchen} & \multicolumn{1}{c}{Room 0} & \multicolumn{1}{c}{Room 2}\\
        \hline
        NeRF                & 29.75 & 32.08 & 31.78 & 30.98 & \textbf{33.17} & \textbf{26.92} & 26.17 \\
        NeX                 & 25.00 & 28.12 & 23.39 & 26.28 & 24.65 & 21.12 & 24.74 \\
        PlenOctree*          & 26.90 & 28.09 & 28.59 & 27.65 & 30.78 & 25.56 & 26.45 \\
        KiloNeRF* & 29.84 & 31.56 & 29.52 & 30.90 & 30.49 & 24.95 & 28.37 \\
        \hline
        Ours & \textbf{30.17} & \textbf{32.48} & \textbf{32.17} & \textbf{32.63} & 32.87 & 26.03 & \textbf{29.28}  \\
        \hline
    
    \end{tabular}
}
{

    \begin{tabular}{ lccccccc}
    
        \multicolumn{8}{c}{SSIM$\uparrow$} \\
        \hline
        \multicolumn{1}{c}{} & \multicolumn{1}{c}{Apartment 0} & \multicolumn{1}{c}{Apartment 1} & \multicolumn{1}{c}{Apartment 2} & \multicolumn{1}{c}{FRL 0} & \multicolumn{1}{c}{Kitchen} & \multicolumn{1}{c}{Room 0} & \multicolumn{1}{c}{Room 2}\\
        \hline
        NeRF                & 0.899 & 0.928 & \textbf{0.917} & 0.913 & \textbf{0.937} & \textbf{0.870} & 0.840 \\
        NeX                 & 0.826 & 0.899 & 0.760 & 0.880 & 0.834 & 0.794 & 0.834 \\
        PlenOctree*          & 0.856 & 0.889 & 0.890 & 0.866 & 0.916 & 0.864 & 0.822 \\
        KiloNeRF*            & \textbf{0.905} & 0.931 & 0.913 & \textbf{0.924} & 0.924 & 0.865 & 0.863 \\
        \hline
        Ours & 0.888 & \textbf{0.933} & 0.913 & 0.921 & 0.935 & 0.830 & \textbf{0.878} \\
        \hline
    
    \end{tabular}
}
{

    \begin{tabular}{ lccccccc}
    
        \multicolumn{8}{c}{LPIPS$\downarrow$} \\
        \hline
        \multicolumn{1}{c}{} & \multicolumn{1}{c}{Apartment 0} & \multicolumn{1}{c}{Apartment 1} & \multicolumn{1}{c}{Apartment 2} & \multicolumn{1}{c}{FRL 0} & \multicolumn{1}{c}{Kitchen} & \multicolumn{1}{c}{Room 0} & \multicolumn{1}{c}{Room 2}\\
        \hline
        NeRF                & 0.093 & 0.069 & 0.073 & 0.083 & 0.064 & \textbf{0.134} & 0.165 \\
        NeX                 & 0.152 & 0.088 & 0.212 & 0.100 & 0.160 & 0.200 & 0.154 \\
        PlenOctree*         & 0.181 & 0.148 & 0.155 & 0.180 & 0.124 & 0.168 & 0.265 \\
        KiloNeRF*           & 0.096 & 0.063 & 0.085 & 0.072 & 0.085 & 0.141 & 0.134 \\
        \hline
        Ours & \textbf{0.093} & \textbf{0.056} & \textbf{0.072} & \textbf{0.067} & \textbf{0.061} & 0.153 & \textbf{0.112}  \\
        \hline
    
    \end{tabular}
}

\bigskip

\vspace{-3mm}
\caption{\textbf{Quantitative results on Replica.}}
\label{tab:quantity_replica}
\vspace{-1mm}
}
\end{table*}
\begin{table*}[h]
\centering
{
\vspace{-5mm}
{

    \begin{tabular}{ lcccc}
    
        \multicolumn{5}{c}{PSNR$\uparrow$} \\
        \hline
        \multicolumn{1}{c}{} & \multicolumn{1}{c}{Character} & \multicolumn{1}{c}{Fountain} & \multicolumn{1}{c}{Jade} & \multicolumn{1}{c}{Statues} \\
        \hline
        SRN             & 21.98 & 21.04 & 18.57 & 20.46 \\
        Neural Volumes  & 24.10 & 22.71 & 22.08 & 23.22 \\
        NeRF            & 29.43 & 28.04 & 26.52 & 25.17 \\
        NSVF            & 27.95 & 27.73 & 26.96 & 24.97 \\
        KiloNeRF        & \textbf{29.44} & \textbf{28.50} & \textbf{27.14} & 24.49 \\
        \hline
        Ours            & 28.54 & 27.23 & 24.64 & \textbf{25.76} \\
        \hline
    
    \end{tabular}
}
{

    \begin{tabular}{ lcccc}
    
        \multicolumn{5}{c}{SSIM$\uparrow$} \\
        \hline
        \multicolumn{1}{c}{} & \multicolumn{1}{c}{Character} & \multicolumn{1}{c}{Fountain} & \multicolumn{1}{c}{Jade} & \multicolumn{1}{c}{Statues} \\
        \hline
        SRN             & 0.853 & 0.717 & 0.715 & 0.794 \\
        Neural Volumes  & 0.876 & 0.762 & 0.750 & 0.785 \\
        NeRF            & \textbf{0.950} & 0.910 & 0.890 & 0.870 \\
        NSVF            & 0.921 & 0.913 & 0.901 & 0.858 \\
        KiloNeRF        & \textbf{0.950} & \textbf{0.930} & \textbf{0.910} & \textbf{0.880} \\
        \hline
        Ours            & 0.945 & 0.901 & 0.846 & 0.866 \\
        \hline
    
    \end{tabular}
}

\bigskip

{

    \begin{tabular}{ lcccc} 
    
        \multicolumn{5}{c}{LPIPS$\downarrow$} \\
        \hline
        \multicolumn{1}{c}{} & \multicolumn{1}{c}{Character} & \multicolumn{1}{c}{Fountain} & \multicolumn{1}{c}{Jade} & \multicolumn{1}{c}{Statues} \\
        \hline
        SRN             & 0.208 & 0.291 & 0.323 & 0.354 \\
        Neural Volumes  & 0.140 & 0.263 & 0.292 & 0.277 \\
        NeRF            & \textbf{0.030} & 0.070 & 0.080 & 0.090 \\
        NSVF            & 0.074 & 0.113 & 0.094 & 0.171 \\
        KiloNeRF        & 0.040 & \textbf{0.060} & \textbf{0.060} & \textbf{0.080} \\
        \hline
        Ours            & 0.042 & 0.075 & 0.109 & 0.115 \\
        \hline
    
    \end{tabular}
}

\vspace{-3mm}
\caption{\textbf{Quantitative results on BlendedMVS.}}
\label{tab:blendedmvs_quantity}
\vspace{-1mm}
}
\end{table*}

\end{document}